\documentclass[a4paper,fleqn]{cas-sc}

\usepackage[square,numbers,compress,sort]{natbib}
\makeatletter
\renewcommand\@biblabel[1]{#1.}
\makeatother
\usepackage{csquotes}
\usepackage{wasysym}
\usepackage{xcolor}
\usepackage[colorinlistoftodos,prependcaption]{todonotes}
\usepackage[inline]{enumitem}
\usepackage{xspace}
\usepackage{multicol}
\usepackage{multirow}
\usepackage{color}
\usepackage{acronym} 
\usepackage{subcaption}
\usepackage{longtable}
\usepackage{afterpage}
\usepackage{csquotes}
\usepackage{amssymb}%
\usepackage{pifont}

\usepackage{adjustbox}

\usepackage[commandnameprefix=ifneeded]{changes}

\usepackage{tikz}
\usetikzlibrary{trees}

\def\tsc#1{\csdef{#1}{\textsc{\lowercase{#1}}\xspace}}
\tsc{WGM}
\tsc{QE}
\tsc{EP}
\tsc{PMS}
\tsc{BEC}
\tsc{DE}

\usepackage{float}
\usepackage[usenames,dvipsnames]{pstricks}
\usepackage{pstricks-add}
 \usepackage{epsfig}
 \usepackage{float}
 \usepackage{pst-grad} 
 \usepackage{pst-plot} 
 \usepackage[space]{grffile} 
 \usepackage{etoolbox} 
 \makeatletter 
 \patchcmd\Gread@eps{\@inputcheck#1 }{\@inputcheck"#1"\relax}{}{}
 \makeatother
\usepackage{verbatim}
\usepackage{tablefootnote}
\usepackage{multicol}
\usepackage[ruled,lined,noresetcount ]{algorithm2e}
\usepackage{algpseudocode}
\usepackage{mathtools}

\begin{document}
\let\WriteBookmarks\relax
\def\floatpagepagefraction{1}
\def\textpagefraction{.001}

\shorttitle{Recent Advances in Transformer Models for UAV Applications}

\shortauthors{Kheddar et~al.}
                      
\title [mode = title]{Recent Advances in Transformer and Large Language  Models for UAV Applications }

\vskip2mm

\author[1]{Hamza Kheddar}
[orcid=0000-0002-9532-2453]
\ead{kheddar.hamza@univ-medea.dz}
\credit{Conceptualization; Methodology; Data Curation; Resources; Investigation; Visualization;  Writing original draft; Writing, review, and editing}

\author[2]{Yassine Habchi}
[orcid=0000-0002-8764-9675]
\ead{habchi@cuniv-naama.dz}
\credit{Conceptualization; Methodology; Resources; Investigation; Writing original draft; Writing, review, and editing}

\author[3]{Mohamed Chahine Ghanem\corref{cor1}}
[orcid=0000-0002-7067-7848]
\ead{mohamed.chahine.ghanem@liverpool.ac.uk }
\cormark[1]
\credit{Conceptualization; Methodology; Resources; Investigation; Writing original draft; Writing, review, and editing}

\author[4]{Mustapha Hemis}
[orcid=0000-0002-6353-0215]
\ead{mhemis@usthb.dz}
\credit{Conceptualization; Methodology; Resources; Investigation; Writing original draft; Writing, review, and editing}

\author[5]{Dusit Niyato}
[orcid=0000-0002-7442-7416]
\ead{dniyato@ntu.edu.sg}
\credit{Conceptualization; Methodology; Resources; Validation; Supervision; Writing, review, and editing}

\address[1]{LSEA Laboratory, Department of Electrical Engineering, University of Medea, 26000, Algeria}

\address[2]{Institute of Technology, University Centre Salhi Ahmed, Naama, Algeria}

\address[3]{Cybersecurity Institute, Department of Computer Science, University of Liverpool, L69 3BX, Liverpool, UK}

\address[4]{LCPTS Laboratory, University of Sciences and Technology Houari Boumediene (USTHB), Algiers, 16111, Algeria.}

\address[5]{College of Computing and Data Science, Nanyang Technological University, Singapore.}

\tnotetext[1]{M. C. Ghanem is the corresponding author.}

\begin{abstract}
The rapid advancement of Transformer-based models has reshaped the landscape of uncrewed aerial vehicle (UAV) systems by enhancing perception, decision-making, and autonomy. This review paper systematically categorizes and evaluates recent developments in Transformer architectures applied to UAVs, including attention mechanisms, CNN-Transformer hybrids, reinforcement learning Transformers, and large language models (LLMs). Unlike previous surveys, this work presents a unified taxonomy of Transformer-based UAV models, highlights emerging applications such as precision agriculture and autonomous navigation, and provides comparative analyses through structured tables and performance benchmarks. The paper also reviews key datasets, simulators, and evaluation metrics used in the field. Furthermore, it identifies existing gaps in the literature, outlines critical challenges in computational efficiency and real-time deployment, and offers future research directions. This comprehensive synthesis aims to guide researchers and practitioners in understanding and advancing Transformer-driven UAV technologies.
\end{abstract}




\maketitle



\section{Introduction} \label{sec1}

The rapid advancement of automation and sensor technology has sparked a growing shift from crewed (or manned) to uncrewed (unmanned) vehicles across various sectors, including defence, transportation, and industry. Uncrewed vehicles—whether aerial, ground-based, or maritime—offer significant advantages over traditional crewed systems, making their adoption a necessity rather than a mere technological trend. One of the primary drivers of this transition is safety. In hazardous environments such as war zones, disaster response areas, or deep-sea explorations, uncrewed vehicles eliminate risks to human operators. By removing direct human involvement, these systems can perform dangerous tasks more efficiently while reducing casualties and operational risks. In industries such as mining and logistics, autonomous vehicles enhance workplace safety by handling high-risk operations with precision.  

\begin{figure}[ht!]
    \centering
    \includegraphics[width=0.9\linewidth]{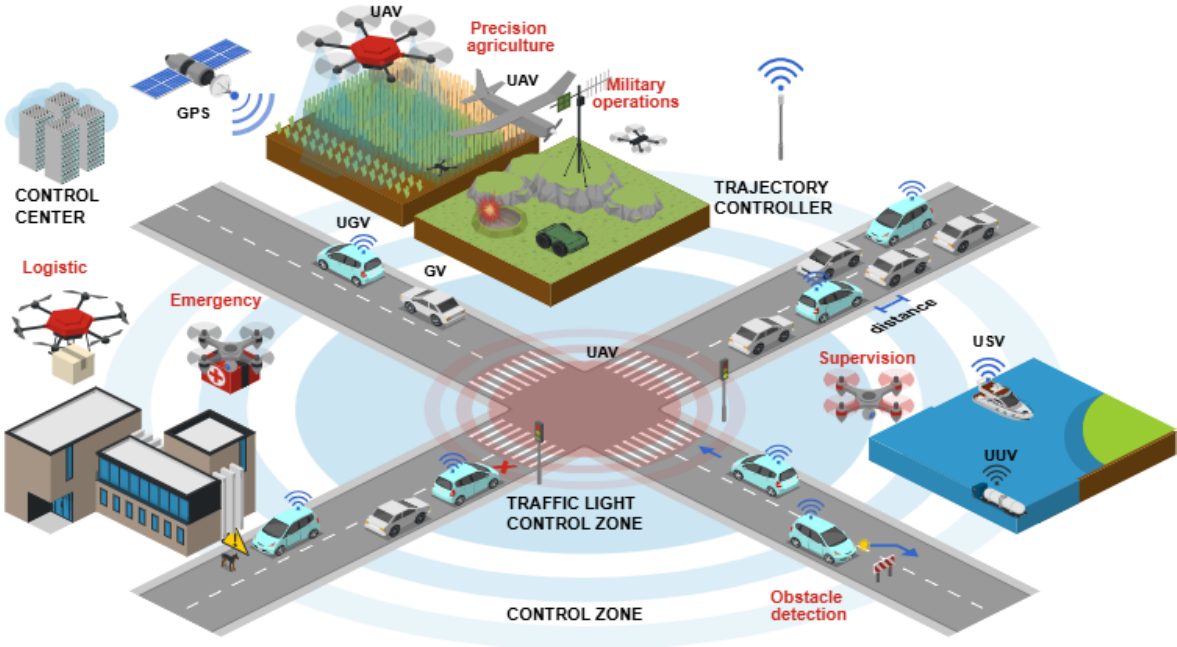}
    \caption{All sorts of uncrewed vehicles, namely \ac{UAV}, \ac{UGV}, \ac{USV}, and \ac{UUV}, are integrated into precision agriculture, military operations, emergency response, logistics, supervision, and obstacle detection within a traffic control zone, enhancing efficiency and safety through real-time monitoring and coordination.}
    \label{fig:ga}
\end{figure}

Efficiency and cost-effectiveness further justify the shift toward uncrewed systems. Autonomous vehicles optimise fuel consumption, minimise human errors, and require less downtime, leading to enhanced productivity. For instance, \acp{UGV} such as self-driving trucks improve logistics efficiency by operating continuously without human fatigue.  Additionally, \acp{USV}, such as the sea hunter, an autonomous naval vessel, enhance maritime surveillance and reconnaissance \cite{fernandes2024extended}, while \acp{UUV} are utilised for underwater exploration and mine detection \cite{waldner2024systematic}. Similarly, \acp{UAV} in agriculture enhance crop monitoring and precision farming, reducing resource wastage \cite{wang2025survey}. Furthermore, technological advancements have significantly improved the capabilities of uncrewed vehicles, making them more reliable and adaptable. Figure \ref{fig:ga} illustrates key applications of \acp{UAV} and autonomous vehicles across various domains, including precision agriculture, military operations, logistics, emergency response, supervision, obstacle detection, and traffic control. The figure also highlights the role of trajectory controllers, \ac{GPS}, and control centres in managing these systems. \acp{UAV} face significant challenges in non-homogeneous environments, such as roads shared with \acp{CV}. These challenges include maintaining safe distances between aerial and terrestrial \acp{CV} and \acp{UAV} to prevent collisions, detecting and avoiding obstacles, responding to potential hazards to protect animals and pedestrians, and complying with traffic signals. Recent advancements in \ac{DL} have significantly improved these capabilities by enhancing object detection, supervision, localisation, and overall system coordination.

\begin{table}[ht!]
\begin{flushleft}
\textbf{List of abbreviations}    
\end{flushleft}
\begin{multicols}{3}
{\scriptsize
\begin{acronym}[SoftNMS] 
\acro{AC}{actor-critic}
\acro{ADE}{average displacement error}
\acro{AE}{autoencoder}
\acro{AI}{artificial intelligence}
\acro{AoI}{age of information}
\acro{AP}{average precision}
\acro{AR}{action recognition}
\acro{BERT}{bidirectional encoder representations from Transformers}
\acro{CA}{coordinate attention}
\acro{CAM}{channel Attention mechanism}
\acro{CBAM}{convolutional block attention module}
\acro{CBAM}{convolutional block attention module}
\acro{CBAM}{convolutional block Attention module}
\acro{CLE}{center location error}
\acro{CNN}{convolutional neural network}
\acro{CV}{crewed vehicle}
\acro{DA}{domain adaptation}
\acro{DL}{deep learning}
\acro{DL}{deep learning}
\acro{DQN}{deep Q-network}
\acro{DRL}{deep reinforcement learning}
\acro{DTL}{deep transfer learning}
\acro{ECA}{efficient channel attention}
\acro{FDE}{final displacement error}
\acro{FL}{federated learning}
\acro{FLOPs}{floating point operations}
\acro{FOD}{foreign object debris}
\acro{FPN}{feature pyramid network}
\acro{FPS}{frames per second}
\acro{GAN}{generative adversarial networks}
\acro{GL}{geo-localization}
\acro{GNN}{graph neural network}
\acro{GPS}{global positioning system}
\acro{GPS}{global positioning system}
\acro{GPU}{graphics processing unit}
\acro{IMU}{inertial measurement unit}
\acro{IoT}{internet of things}
\acro{IoU}{intersection over union}
\acro{IRS}{intelligent reflecting surface}
\acro{LDA}{latent Dirichlet allocation}
\acro{LiDAR}{light detection and ranging}
\acro{LLM}{large language models}
\acro{LSTM}{long short-term memory}
\acro{MADDPG}{multi-agent deep deterministic policy gradient}
\acro{MAE}{mean absolute error}
\acro{mAP}{mean average precision}
\acro{MI}{mutual information}
\acro{mIoU}{mean intersection over union}
\acro{ML}{machine learning}
\acro{MLP}{multilayer perceptron}
\acro{MOT}{multi-object tracking}
\acro{MOTA}{multiple object tracking accuracy}
\acro{MSA}{multihead self-Attention}
\acro{MSFF}{multi-scale feature fusion}
\acro{MTL}{multitask learning}
\acro{PG}{policy gradient}
\acro{PSNR}{peak signal-to-noise ratio}
\acro{RAG}{retrieval-augmented generation}
\acro{RMSE}{root mean square error}
\acro{RNN}{recurrent neural network}
\acro{SAC}{soft actor-critic}
\acro{SCT}{spatial-channel Transformer}
\acro{semi-SL}{semi-supervised learning}
\acro{SLM}{small language models}
\acro{SPP}{spatial pyramid pooling}
\acro{SSIM}{structural similarity index measure}
\acro{SSL}{self-supervised learning}
\acro{STT}{spatio--temporal Transformer}
\acro{TST}{time-series Transformer}
\acro{UAV}{uncrewed aerial vehicle}
\acro{UCV}{uncrewed vehicle}
\acro{UGV}{uncrewed ground vehicle}
\acro{USV}{uncrewed surface vehicle}
\acro{UUV}{uncrewed underwater vehicle}
\acro{ViT}{vision Transformers}
\acro{VLM}{vision-language model}
\acro{YOLO}{you only look once}
\end{acronym}

}
\end{multicols}
\end{table}

\Ac{AI}-powered navigation, \ac{ML} algorithms, and real-time data processing enable these systems to function autonomously with minimal human intervention. This has led to increased military reliance on uncrewed combat systems, reducing personnel exposure to hostile environments while enhancing strategic capabilities. As industries and governments increasingly recognize these benefits, transitioning from \acp{CV} to \acp{UCV} is no longer an option but a necessity. With continuous innovation, uncrewed systems will revolutionise mobility, safety, and efficiency, shaping the future of transportation and operational logistics. While \ac{AI} provides a broad framework for autonomous systems, \ac{ML} and \ac{DL} are more effective in enhancing \acp{UAV}. \ac{ML} enables \acp{UAV} to learn from vast datasets, improving navigation, obstacle avoidance, and target recognition without relying on pre-programmed rules.  \Ac{DL}, particularly \acp{CNN} and it related pre-trained models \cite{himeur2023video} enhances real-time decision-making in \acp{UAV} by enabling object detection, \acp{RNN} provide power consumption prediction \cite{saadi2025power}, and other \ac{DL} techniques that help in autonomous flight path planning, optimization, and obstacle avoidance \cite{debnath2024review}. Unlike general \ac{AI}, which aims at human-like reasoning, \ac{DL} excels in pattern recognition, reducing computational overhead while improving precision.  Moreover, \ac{ML} and \ac{DL} algorithms allow \acp{UAV} to adapt dynamically to environmental changes, improving flight efficiency and mission success \cite{sharma2025path}. Their ability to process complex sensor data in real time makes them indispensable in surveillance, disaster response, and logistics, surpassing traditional \ac{AI} approaches in reliability and performance.

Recently, advanced \ac{DL} techniques—such as \ac{DRL} \cite{gueriani2023deep}, \ac{DTL} \cite{kheddar2023deepASR,lachenani2025advancing}, and \ac{FL} have demonstrated superior adaptability, efficiency, and autonomy \cite{kheddar2024automatic}. \ac{DRL} enables \acp{UAV} to make real-time decisions in dynamic environments by learning from trial and error, which is crucial for applications such as autonomous flight control and multi-agent coordination. Additionally, \ac{DRL} enhances \ac{UAV} security by improving resilience against attacks and intrusions \cite{kheddar2024reinforcement}.  Similarly, \ac{DTL} allows \acp{UAV} to leverage pre-trained models, significantly reducing computational costs and enabling faster adaptation to new tasks \cite{wu2020transfer}, such as recognising novel terrains or objects. Furthermore, \ac{DTL} strengthens security by mitigating vulnerabilities to \ac{DL}-based attacks \cite{kheddar2023deep}. By leveraging \ac{FL} \cite{himeur2023federated}, \acp{UAV} can learn from distributed data sources while reducing communication overhead and computational costs. This approach is particularly useful in dynamic and resource-constrained environments, such as disaster response, surveillance, and military operations. \ac{FL} improves adaptability by allowing \acp{UAV} to update models locally before aggregating insights with a central server, ensuring robustness against adversarial attacks and data breaches. Additionally, \ac{FL} mitigates latency issues, making real-time decision-making more efficient.

Despite the benefits of traditional \ac{DL} and advanced \ac{DL} techniques, their limitations—such as poor long-range dependencies, inefficiency in handling sequential data, and high computational demands—necessitate integration with Transformer-based architectures \cite{djeffal2023automatic}, to enhance \ac{UAV} performance in real-world scenarios.  For example, CNN-Transformers improve \ac{UAV} vision systems by combining CNN’s spatial feature extraction with Transformers’ global context awareness, enhancing object detection, terrain mapping, and low-light imaging. \ac{DRL}-Transformers refine autonomous decision-making by incorporating attention mechanisms, allowing \acp{UAV} to optimise flight strategies in complex environments through improved sequential decision modelling.  Moreover, \ac{ViT} enhance \ac{UAV}-based re-identification and tracking, crucial for security surveillance and disaster management. By leveraging self-attention mechanisms, they improve feature discrimination, reducing false positives in long-term tracking. Similarly, \acp{STT} strengthen \ac{UAV} trajectory prediction by capturing both spatial and temporal dependencies, essential for motion planning and swarm coordination.  Furthermore, \acp{LLM} can be integrated with \acp{UAV} for intelligent mission planning, natural language-based drone control, and enhanced situational awareness through multimodal data fusion. By processing text, image, and sensor data together, \ac{LLM}-augmented \acp{UAV} can perform more autonomous, context-aware operations.  Merging Transformers with existing \ac{DL} techniques offers \acp{UAV} greater adaptability, improved computational efficiency, and enhanced perception, paving the way for more autonomous, resilient, and intelligent \ac{UAV} systems across various applications, including surveillance, logistics, and disaster response.

\subsection{Motivation and related work}

The increasing adoption of \acp{UAV} across various domains, including surveillance, agriculture, logistics, and security, has driven the need for more intelligent and autonomous decision-making capabilities. Consequently, there has been a significant rise in research interest in leveraging advanced AI-driven techniques, particularly Transformer-based architectures, to enhance \ac{UAV} capabilities. Researchers are increasingly exploring novel approaches to improve \ac{UAV} perception, adaptability, and operational efficiency, leading to a surge in publications and innovative applications in this domain.  Transformer-based models, which have revolutionised natural language processing and computer vision, offer a promising solution by enhancing \ac{UAV} perception, adaptability, and operational efficiency. Recent developments, including \acp{ViT}, Swin Transformers, and \acp{LLM}, have shown potential in improving \ac{UAV}-based tracking, anomaly detection, and autonomous navigation. However, despite their growing adoption, a comprehensive review systematically categorising and analysing Transformer-based \ac{UAV} applications remains lacking. 

On the one hand, several studies have surveyed Transformer applications, but most overlook advancements in specialised variants, such as \acp{ViT}, \ac{STT}s, and hybrid approaches integrating Transformers with \ac{DL} techniques (e.g., CNN-Transformer, \ac{DRL}-Transformer, YOLO-Transformer). Their role in \ac{UAV} applications also remains underexplored. Table~\ref{tab:01} presents a comparative summary of key contributions, highlighting existing gaps in current \ac{UAV} surveys. Most existing surveys and reviews exhibit significant omissions across several core domains. Attention mechanisms, \ac{DRL}-based Transformers, Siamese architectures, and \acp{STT} have never been discussed in the surveyed literature. In addition, variants such as YOLO-Transformer, ViT, Swin, and \acp{LLM} are frequently overlooked. Moreover, works by \cite{teixeira2023deep,andrade2022went,vo2024aerial} provide only limited or no coverage of these architectures, focusing instead on narrow subsets or omitting technical analysis altogether. Similarly, application-oriented discussions remain sparse in \cite{teixeira2023deep,andrade2022went,debas2024forensic}, or are superficial in \cite{kang2022survey,tian2025uavs,vo2024aerial}, restricting their practical relevance. Likewise, analyses of challenges and future directions are absent in \cite{teixeira2023deep,andrade2022went,vo2024aerial,tian2025uavs}, or are treated only superficially in \cite{kang2022survey}. This limited or shallow treatment restricts the ability of these reviews to guide future research priorities or address open technical and practical issues in the \ac{UAV} field.

On the other hand, a comprehensive overview of CNN-Transformer models is provided by \cite{kang2022survey}. The paper \cite{debas2024forensic} addresses only challenges and future directions, without technical depth, while another work, \cite{javaid2024large}, focuses exclusively on \acp{LLM} and applications, omitting key model architectures.

Notably, this review extends prior work by comprehensively covering both Transformer-based and hybrid Transformer-\ac{DL} \ac{UAV} models, thereby providing a more holistic and up-to-date synthesis. In contrast to previous works, our review systematically addresses the full spectrum of Transformer models, \acp{LLM}, application domains, and open challenges, bridging the fragmented landscape of existing literature.

\begin{table}[ht!]
\centering
\small
\caption{Comparison of the proposed review with existing \ac{UAV}-related surveys and reviews. A  mark (\CIRCLE{}) signifies that a specific area has been covered, whereas a  mark (\Circle{}) denotes that it has not been addressed. The symbol (\LEFTcircle{}) indicates that key aspects of the field remain largely unaddressed.}
\label{tab:01}
\begin{tabular}[ht!]{m{0.4cm}m{0.7cm}cccccccccccc}
\hline
Ref. & Paper  &\multirow{2}{*}{Attention } & 
\multirow{2}{*}{CNN-T} & \multirow{2}{*}{RL-T}  & \multirow{2}{*}{Siamese}  & \multirow{2}{*}{Swin}  & \multirow{2}{*}{STT} & \multirow{2}{*}{YOLO-T} & \multirow{2}{*}{ViT} & \multirow{2}{*}{LLM} & \multirow{2}{*}{Apps} & Challenges and  \\ 
& type &  &  &   &    &   &   &    &  & & & 
 future directions  \\ \hline

\cite{teixeira2023deep} & Review &  \Circle{} & \LEFTcircle{} & \Circle{} & \Circle{} &  \Circle{} & \Circle{} &  \Circle{} & \LEFTcircle{} & \Circle{} & \Circle{} & \Circle{} \\ \hline
  
\cite{andrade2022went} & Survey &  \Circle{}  & \Circle{}  & \Circle{}  & \Circle{}  &  \Circle{}  & \Circle{}  &  \Circle{}  & \Circle{}  & \LEFTcircle{} & \Circle{}   & \Circle{} \\ \hline

 \cite{vo2024aerial} &  Survey &  \Circle{} & \Circle{} & \Circle{} & \Circle{} &  \LEFTcircle{} & \Circle{} &  \LEFTcircle{} & \Circle{} & \Circle{} & \LEFTcircle{}  & \Circle{} \\ \hline
 
 \cite{kang2022survey} & Survey &  \Circle{} & \CIRCLE{} & \Circle{} & \Circle{} &  \Circle{}&  \Circle{} & \Circle{} &  \Circle{} & \Circle{} & \LEFTcircle{} & \LEFTcircle{}\\ \hline

 \cite{debas2024forensic} & Review &  \Circle{} & \Circle{} & \Circle{} & \Circle{} &  \Circle{} & \Circle{} &  \Circle{} & \Circle{} & \Circle{} & \Circle{}  & \CIRCLE{} \\ \hline
 
 \cite{javaid2024large} &  Survey &  \Circle{} & \Circle{} & \Circle{}  & \Circle{} &  \Circle{} & \Circle{} &  \Circle{} & \Circle{} & \CIRCLE{} & \CIRCLE{}  & \CIRCLE{} \\  \hline

\cite{tian2025uavs} & Review  & \Circle{} & \Circle{} & \Circle{} & \Circle{} & \Circle{} & \Circle{} & \Circle{} & \Circle{} & \CIRCLE{} & \LEFTcircle{}& \Circle{} \\\hline

\textbf{Our} &  Review &  \CIRCLE{} & \CIRCLE{} & \CIRCLE{} & \CIRCLE{} &  \CIRCLE{} & \CIRCLE{} &  \CIRCLE{} & \CIRCLE{} & \CIRCLE{} & \CIRCLE{}& \CIRCLE{}\\ 
\hline
\end{tabular}
\begin{flushleft}
\scriptsize{Abbreviations: Spatio-temporal Transformer (STT);  Transformer (T); Applications (Apps).}     
\end{flushleft}
\end{table}


\subsection{Our contribution and review organisation}

This paper bridges the gap in existing literature by providing a comprehensive and structured evaluation of Transformer-driven UAV methodologies. Unlike previous surveys, our work offers a holistic and up-to-date synthesis that addresses the rapidly evolving landscape of UAV autonomy. The primary contributions of this paper are as follows:
\begin{itemize}
    \item Introduces a comprehensive taxonomy of Transformer-based \ac{UAV} architectures, covering attention mechanisms, \ac{CNN}, \ac{DRL}, \acp{STT}, Swin, \ac{ViT}, YOLO, Siamese, and \ac{LLM}-based approaches to map recent advancements.

    \item Analyses the role of Transformers in a diverse set of \ac{UAV} applications, such as real-time tracking, object detection, anomaly detection, localisation, autonomous navigation, precision agriculture, security, and multimodal sensor fusion.
    \item Reviews existing \ac{UAV} simulators used for dataset generation and scenario-based virtual simulations. Summarizes key evaluation metrics across many application categories. Also reviews major datasets covering diverse modalities and tasks for Transformer-based \ac{UAV} research.

    \item  The manuscript includes Tables~\ref{tab:transUAVcompar}, \ref{tab:attention_comparison}, and \ref{tab:rl_comparison}, which facilitate model selection by comparing Transformer architectures, attention mechanisms, and \ac{DRL} techniques based on application fit, complexity, modality, and real-time suitability in \ac{UAV} systems.

    \item 
    The manuscript includes Tables \ref{table:6}, \ref{table:8}, and \ref{table:9}, which offer performance comparisons of Transformer-based \ac{UAV} applications, enabling quick assessment of accuracy, efficiency, and limitations across benchmarks and tasks.

    \item  Presents two case studies: one on implementing a Transformer-based \ac{UAV} solution, and another on using LLM-based Transformers for intelligent control and mission planning, offering practical insights into design and deployment.

    \item  Beyond technical analysis, the manuscript discusses key challenges in integrating Transformers into \ac{UAV} systems, such as scalability, real-time constraints, and data limitations, and outlines future research directions to address these issues and advance the field.

\end{itemize}


The remainder of this paper is organized as follows:  Section  Section \ref{sec4} presents an in-depth analysis of Transformer-based \ac{UAV} models, categorized by Transformer architecture combined with \ac{DL} model. \ref{sec3} reviews essential background knowledge, including \ac{UAV} simulators, evaluation metrics, and benchmark datasets. Section \ref{sec5} reviews and presents state-of-the-art applications of Transformer-based \ac{UAV} systems. Section \ref{sec6} presents two case studies: one on Transformer-based \ac{UAV} implementation and one on LLM-based Transformer solutions for UAVs. Section \ref{sec7} discusses key research challenges, open problems and future directions. Finally, Section \ref{sec8} concludes the review with a summary of findings and potential directions for future work.

\section{Transformer-based models for UAV}
\label{sec4}

This section provides a comprehensive discussion of the diverse Transformer-based architectures currently applied in \ac{UAV} systems. Given the heterogeneity of \ac{UAV} tasks—ranging from object tracking and scene segmentation to trajectory forecasting and decision-making—different Transformer variants have been tailored to specific operational demands. To guide the reader through this evolving landscape, Table~\ref{tab:transUAVcompar} offers a refined comparison across major Transformer-based \ac{UAV} models. It summarizes their suitable problem scenarios, advantages, limitations, typical applications, and real-time capabilities. 

\renewcommand{\arraystretch}{1.30}
\begin{table}[ht!]
\centering
\scriptsize
\caption{Refined comparison of Transformers-based \ac{UAV} models covering Key applications} \label{tab:transUAVcompar}
\begin{tabular}{m{2.2cm}m{3cm}m{3cm}m{2.5cm}m{3cm}m{1.5cm}}
\hline
\textbf{Model} & \textbf{Problem scenario \newline suitable for} & \textbf{Advantages} & \textbf{Limitations} & \textbf{Typical applications} & \textbf{Real-time capability} \\
\hline
Attention & General tracking and navigation & Strong focus, global context & Heavy computation & Tracking and detection & Low \\
\hline
Siamese & One-shot matching for tracking & Robust similarity learning & Sensitive to occlusions & Tracking and detection & Medium \\
\hline
YOLO-Transformers & Real-time object detection & Fast and accurate & Heavier than pure YOLO & Tracking and detection, Security & High \\
\hline
CNN-Transformer & Complex scenes needing local and global features & Best of CNN and Transformer & Training complexity & Segmentation, Localization, Precision Agriculture & Medium \\
\hline
Swin & High-resolution input handling & Efficient with shifted windows & May lose global links & Localization, Segmentation & Medium \\
\hline
ViT & Large dataset classification & Strong performance & Needs big data & Precision Agriculture, Segmentation & Low \\
\hline
RL-Transformer & Sequential decision tasks & Learns adaptive strategies & Large training times & Security, Localization & Low \\
\hline
LLM & Multimodal reasoning and planning & Understands language, integrates multimodal data & Very high latency & Human-UAV collaboration & Low \\
\hline
Spatio-Temporal Transformer & Video/image sequence analysis & Spatial and temporal understanding & High memory use & AR, Security & Medium \\
\hline
Unsupervised Transformer & Prediction in unlabeled settings & Works without labels & Less robust than supervised & Security and anomalies detection, Trajectory Forecasting & Medium \\
\hline
Other Transformers & Customized hybrid tasks & Task-specific optimization & Poor generalization & Precision Agriculture, Cooperative Missions & Medium \\
\hline
\end{tabular}
\end{table}

\subsection{Attention-based UAV}

\noindent\textbf{-- Attention-only:} is the fundamental building block of the Transformer architecture. In \ac{UAV}-based vision tasks, attention-based architectures operate directly on image patches.  The input image \(X \in \mathbb{R}^{H \times W \times C}\) is divided into \(N = \frac{HW}{P^2}\) non-overlapping patches, each embedded into \(d\)-dimensional tokens to form \(Z_0 \in \mathbb{R}^{N \times d}\). These tokens, enriched with positional encodings, are passed through stacked Transformer blocks to model global dependencies, yielding \(Z_L\). A task-specific output head maps \(Z_L\) to prediction \(y\). This design enables strong contextual modelling and global reasoning, making it effective for \ac{UAV} imagery where small, variably-scaled targets require fine-grained and long-range attention. Table~\ref{tab:attention_comparison} provides a comparative overview of attention mechanisms used in Transformer-based \ac{UAV} applications.
It highlights the type, purpose, complexity, and best-fit scenarios for each model across different sensing modalities.

\[
X \xrightarrow{\text{Patchify}} Z_0 \xrightarrow{\text{Transformer}} Z_L \xrightarrow{\text{Head}} y
\]

\begin{table*}[ht!]
\centering
\caption{Comparison of Attention mechanisms in Transformer-based UAV studies.}
\label{tab:attention_comparison}
\scriptsize
\begin{tabular}{m{3cm}m{1.5cm}m{4cm}m{1.5cm}m{4cm}m{1cm}}
\hline
 \textbf{Attention Type}  & \textbf{Hierarchy} & \textbf{Modality} & \textbf{Complexity} & \textbf{Best Scenario Fit} & \textbf{Used in} \\
\hline
 Global spatial attention  & Yes & Occupancy map (vision-guided) & Moderate  & Autonomous 3D exploration in unknown indoor/outdoor environments & \cite{chen2023stexplorer,li2023boosting}\\
\hline
 Spatio-temporal attention   & No & Network traffic data & Low  & Real-time UAV deployment in dynamic wireless networks & \cite{hu2023stdformer} \\
\hline
Multi-head self-attention & No & Visual scenes / image patches & Moderate & Generic UAV perception and global feature encoding (e.g., ViT) & \cite{huang2023easy} \\ 
\hline
 Hierarchical self-attention  &  Yes \tiny{(Local+Global)} & 3D voxelized scenes & High  & Robust object tracking under motion and occlusion & \cite{li2023boosting} \\
\hline
 Pixel-wise spatial attention and cross-attention  & Yes \newline \tiny{(Pixel-wise+Cross)} & Infrared video & High  & IR-based surveillance with frequent occlusion and reappearance & \cite{fang2024online} \\
\hline
 Multi-scale feature attention   & Yes \newline \tiny{(Multi-scale)} & Visual surveillance data & Moderate to High  & Cross-platform vehicle tracking between UAV and fixed cameras & \cite{holla2024msfft} \\
\hline
 Channel attention (CAM/CBAM) & No & Multi-spectral or multi-modal UAV images & Low to Moderate & Enhancing feature map discrimination and fusion for better perception &  \cite{zhao2023ms} \\
\hline
 Window-based attention (Swin) & Yes \newline \tiny{(Hierar. windows)} & High-resolution aerial images & Moderate to High & Large-scale UAV image segmentation and cross-view localization tasks & \cite{ye2022tracker} \\
\hline
\end{tabular}
\end{table*}

\begin{figure}[ht!]
\centering
\includegraphics[width=0.55\linewidth]{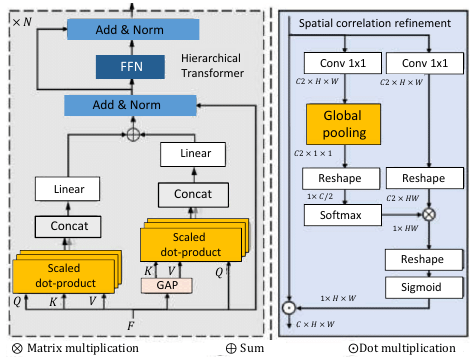}
\caption{An example of the hierarchical self-attention Transformer, incorporating both local and global attention mechanisms. The local attention module is shown on the left, and the global attention module is on the right, as described in \cite{li2023boosting}.}
\label{fig:localGlobal}
\end{figure}

For example, in \cite{chen2023stexplorer}, a hierarchical exploration framework integrates a Transformer-based occupancy predictor to infer spatial information beyond the \ac{UAV}'s current field of view. The Attention mechanism enhances information gain estimation by modelling occluded areas and fragmented regions, thereby supporting efficient spatio-temporal exploration. The Transformer component is strategically embedded within a two-layer planning structure to optimize path selection while minimizing unnecessary backtracking and redundant navigation. In the context of autonomous wireless network planning, \cite{wang2023autonomous} introduces a dual Transformer network (DTN) that jointly captures temporal and spatial dependencies in dynamic traffic patterns. The Attention mechanism plays a central role in modelling user demand fluctuations, enabling \acp{UAV} to be deployed proactively. This design balances accuracy and efficiency by reducing Transformer complexity from quadratic to log-linear time, thus supporting real-time \ac{UAV} network orchestration. Wang et al. \cite{wang2023autonomous} aim to solve routing problems for multiple cooperative \acp{UAV} by leveraging Attention mechanisms to address the multi-agent orienteering problem with shared regions. Attention layers are employed to model inter-region dependencies and facilitate adaptive route generation. The architecture follows a two-stage strategy: initial clustering-based planning, followed by Transformer-based inference for efficient routing. The work \cite{li2023boosting} proposed a hierarchical self-attention Transformer to track \ac{UAV} targets in complex environments. The Attention module is divided into local and global branches, allowing simultaneous modelling of fine-grained appearance details and global 3D structure. The Transformer is pre-trained using a novel trajectory-aware strategy, which disentangles object shape from \ac{UAV} motion, leading to more accurate and robust tracking under viewpoint changes. Figure \ref{fig:localGlobal} provides a reproducible example of a hierarchical self-attention Transformer for \ac{UAV} applications, demonstrating both local and global attention mechanisms. \cite{fang2024online} presents a Transformer-driven tracking framework that incorporates full spatial resolution channel Attention that modulates pixel-wise importance without spatial compression, improving the ability to detect small targets in cluttered infrared scenes. Additionally, a cross-attention mechanism facilitates template updating during online tracking, enhancing robustness to occlusion, target disappearance, and reappearance in dynamic surveillance scenarios. Lastly, \cite{holla2024msfft} employs a Transformer to fuse vehicle features observed from \ac{UAV}'s cameras. Attention scores across multiple inception layers help learn scale-invariant, viewpoint-agnostic representations. The model is trained in two stages: a self-supervised phase for learning cross-platform transformations and a supervised phase for re-identification. The Transformer’s role is crucial in harmonizing features from diverse visual domains using semantic Attention.

\begin{table}[ht]
\centering
    \caption{Overview of UAV-based proposed algorithms using CNN-T, Siamese-T, STT, Swin, Attention, and \ac{DRL} Transformer models.}
{\fontsize{6.5}{7.2}\selectfont
\begin{tabular}{m{0.5cm} m{1.2cm} m{5cm} m{2cm} m{2cm} m{4cm} m{0.8cm}}
\label{table:6} \\
\hline
\textbf{Ref} & \textbf{AI model}  & \textbf{Contributions}  & \textbf{Used data}  & \textbf{Best result} & \textbf{Limitations} & Code \\
\hline

\cite{lu2024lightweight} & CNN-T & Enhancing real-time semantic segmentation of UAV imagery, particularly for boundaries and small objects.  & UDD & mIoU= 76.6  & Scalability, real-time tradeoffs, and dataset size constraints. & --- \\\hline

\cite{hafeez2023abnormal} & CNN-ViT & Abnormal image detection in smart agriculture to prevent decision errors & UAV images  & Acc= 95.12 & Lack of interpretability and anomaly visualization  & ---\\\hline

\cite{ye2022ct}  & CNN-T & Combines convolution and Transformers for accurate, lightweight, and efficient low-altitude object detection  & Low-Altitude  & mAP= 96.6  \newline 	FPS=  136 \newline GFLOPs=  67.3 & CT-Net-Large has high complexity, large size, and slower inference for real-time applications. & --- \\\hline

\cite{he2024dcd} & CNN-T & Employing Transformer and deformable convolutions to improve UAV localization accuracy.  & UL14 &  RDS= 77.15 \newline GFLOPs= 11.54  & Limited generalization due to evaluation on only one dataset. & --- \\\hline

\cite{zhao2023dstnet} & CNN-T & Enhances cross-resolution vehicle ReID using consistent feature extraction.  & Ultra\_UAV & mAP= 78 & Super-resolution weakening, identity inconsistency,  alignment difficulty. & --- \\\hline

\cite{huang2023easy} & CNN-T & Lightweight building extraction network with feature fusion strategy based on observed building features & WHU Aerial \newline UAV  & IoU=89.16 \newline IoU=79.64  & Sensitive to resolution; discard deepest features to reduce complexity. &  \href{github.com/teddy132/EasyNet_for_building_extraction}{GitHub}\\\hline

\cite{cao2022tctrack} & CNN-T & Temporal contexts are exploited for aerial tracking via adaptive feature extraction and similarity refinement.  & DTB70 & Pre = 81.30\newline IoU = 62.20 \newline CLE < 20 & Modeling long-term dependencies and missing target scenarios.  & \href{https://github.com/vision4robotics/TCTrack}{GitHub}\\\hline

\cite{hu2024tfitrack} & CNN-T & Enhance UAV tracking using spatio-temporal integration and dual-attention mechanisms. & DTB70 & Pre= 81.3 \newline IoU= 60.5 &  High dependency on labeled data. &  ---\\\hline

\cite{zhou2023hybrid} & CNN-T & Hybrid Swin-CNN network improves UAV segmentation via multiscale and fusion  & UAVid & mIoU: 75.42 \newline mF1: 85.48 & High computational cost limits deployment on lightweight embedded UAV devices. & ---\\\hline

\cite{li2023boosting} & Hierarchical Self-Attention  & Introduces voxel-based trajectory-aware pre-training for UAV tracking, enhancing motion and viewpoint variation robustness. & UAV123 & Pre= 79.70 \newline Sr= 60.10 & High computational cost due to voxel processing and dual-branch attention. & \\\hline

\cite{fang2024online} & Cross-Attention & Proposes FSRCA module with full-resolution pixel-wise attention and cross-attention, improving tracking under occlusion and reappearance. &  Anti-UAV & Pre= 91.10 \newline Sr= 72.60 & Requires careful tuning of templates and hyperparameters for stable performance. &  \\\hline

\cite{xing2022siamese} & Siamese-T  & Adaptive network selection and efficient spatial attention for tracking  & UAV20L & Pre= 86.5 & Limited robustness under extreme motion and noisy predictions. & ---\\\hline

\cite{xing2022siamese2} & Siamese-T & Fused pyramid features with lightweight Transformers for robust, real-time tracking. & UAV123 & Pre= 85.83 \newline AUC= 66.04 \newline FPS= 30  & Performance drops on harder datasets (e.g., LaSOT)  & \href{https://github.com/RISC-NYUAD/SiamTPNTracker}{GitHub} \\\hline

\cite{zhai2024cas}  & Siamese-T & Enhanced change detection accuracy, reduced noise, faster inference, high-resolution UAV images. & UAV-SD & Acc = 93.1 \newline
Pre = 94.3 & Limited pixel-level resolution, patch size sensitivity, difficulty with minimal changes. & \href{https://github.com/tulingLab/CAS-Net}{GitHub} \\\hline

\cite{chen2024spatial} & Siamese-T  & Fused spatial multiscale features and temporal context to enhance UAV tracking. & DTB70 & Pre= 81.3 \newline AUC= 63.1& Performance degraded under heavy occlusion; lacked adaptive trajectory modeling.  & ---\\\hline

\cite{wu2023developing} & TST & Predict lane-level speed using UAV and TST model & Aerial traffic videos & Acc = 98.35 \newline MAPE= 1.65 & Limited generalization to other traffic scenarios & --- \\\hline

\cite{ahmad2024transformer} & TST & Predict and classify UAV sensor failures using Attention Transformer & BASiC  & Acc= 94 (classif.)\newline Acc= 86 (predic.) & Misclassification and post-failure detection delays & --- \\\hline

\cite{hu2023stdformer} & STT & Pure motion-based tracking using STT and contrastive learning integration. & VisDrone2019 & Pre = 77.9 \newline F1= 57.1 & Depends on detection quality; struggles under extreme occlusions and noise.  & \href{https://github.com/Xiaotong-Zhu/STDFormer}{GitHub}\\\hline

\cite{zhu2023spatio} & STT & Fused multi-level spatio-temporal features for robust and efficient UAV tracking. & UAV123  & Pre= 84.6 \newline Sr= 64.2 & Challenges handling low-resolution targets and efficiency concerns.  & ---\\\hline

\cite{siyuan2024learning} & STT & Capturing short-term spatial–temporal UAV dependencies & 3rd Anti-UAV & ADE= 0.0005 \newline FDE= 0.0002 & Requires high-precision sensors and struggles with multiple targets.  & ---\\\hline

\cite{shen2023adaptive}  & STT & Dual-template tracking using STT feature fusion. & VisDrone2019 & MOTA= 57.5 \newline IDF1= 52.2 & Limited global context integration affects long-term tracking robustness. & ---\\\hline

\cite{chen2023cross} & SwinV2-B & Dense partitioning strategy enhances cross-view localization accuracy & University-1652 & Acc= 91.57 & May struggle with extreme viewpoint variations & --- \\\hline

\cite{wang2024forest} & GCST & Enhances UAV fire detection via ghost Swin Transformer and EAGIoU loss function & Synthetic & mAP$_{50}$= 88.7\newline Speed= 320 FPS & Limited real data, occlusion challenges, and complex background hinder generalization.  & \href{https://github.com/luckylil/forest-fire-detection}{Github}\\\hline

\cite{li2024soybeannet}& SoybeanNet & Swin Transformer processes UAV data for accurate soybean disease leaf counting. & UAV-real images & Acc= 84.51 & Limited generalization across environments; UAV variability affects detection robustness. & \href{https://github.com/JiajiaLi04/Soybean-Pod-Counting-from-UAV-Images}{Github}\\\hline

\cite{li2024lswinsr} & LSwinSR & Develops lightweight Swin Transformer for UAV-based image super-resolution tasks. & AID & SSIM= 94.01\newline PSNR= 35.52\newline Speed= 14.2 FPS & Does not assess performance under noisy or compressed UAV conditions. No UAV deployment or field evaluation. & \href{https://github.com/lironui/GeoSR}{Github}\\\hline

\cite{gao2023sequential} & DeLighT-T & Sequential decision-making for composite UAV actions using DeLighT Transformer in multi-agent DRL & Simulation & 47\%$\uparrow$ Recon. \newline 9\%$\uparrow$  vs MAT & Limited by subtask coordination and reward design & --- \\\hline

\cite{jiang2023short}& MAT-V & Introduces virtual agents in multi-agent Transformer for improved UAV dogfight coordination &Simulator & WR= 82.8 & Scalability issues in dense multi-agent air combat scenarios & --- \\\hline

\cite{neves2024multimodal} & DRL + ViT & ViT-inspired Transformer with CNN-MLP fusion and \Ac{DQN} for UAV landing. &  Sensory data (visual, thermal, LiDAR) & AP$_{50}$ = 90\newline Time= 4.97 ms & Relies on multimodal markers; limited generalization to unseen marker types & --- \\

\hline

\end{tabular}}
\begin{flushleft}
    \scriptsize{\textbf{Abbreviations:} CNN-Transformer (CNN-T); Siamese-Transformer (Siamese-T); Accuracy (Acc); Re-identification (ReID); Success rate (Sr)\\
    \textbf{Note:} All AI metrics are in (\%).}
\end{flushleft}
\end{table}

\vspace{0.5cm}

\noindent\textbf{-- CNN–Transformer:} hybrids leverage \acp{CNN} for efficient local feature extraction and Transformers for capturing long-range dependencies, resulting in a balanced architecture that enhances accuracy, contextual understanding, and generalization in vision tasks. The working principle of CNN–Transformer models can be summarized as follows:

\[
X \xrightarrow{\text{CNN}} \text{Feature Maps} \xrightarrow{\text{Flatten}} \text{Tokens } Z_0 \xrightarrow{\text{Transformer}} Z_L \xrightarrow{\text{Head}} y
\]

\noindent where, \(X\) represents the input data, typically a 2D tensor of shape \(\mathbb{R}^{H \times W \times C}\), where \(H\), \(W\), and \(C\) are the height, width, and number of channels, respectively. The \ac{CNN} block extracts local spatial features and may reduce the resolution, producing feature maps of shape \(\mathbb{R}^{H' \times W' \times K}\). These feature maps are then flattened or reshaped into a sequence of \(N = H' \times W'\) tokens, each of dimension \(d = K\), resulting in \(Z_0 \in \mathbb{R}^{N \times d}\). This token sequence, possibly enriched with positional encodings, is passed through a Transformer, which applies self-attention and feed-forward layers to model global dependencies across the input, yielding the output sequence \(Z_L\). Finally, a task-specific output head processes \(Z_L\) to produce the final prediction \(y\), which could represent class probabilities (for classification), spatially dense predictions (for segmentation), or regression targets.

\begin{figure}
    \centering
    \includegraphics[width=0.9\linewidth]{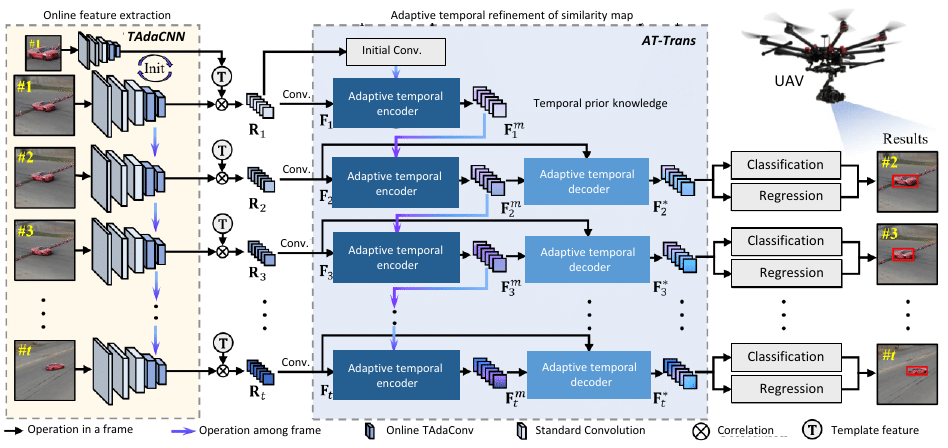}
    \caption{An example of CNN-Transformer architecture for \ac{UAV} tracking. The figure illustrates the workflow of TCTrack, highlighting the integration of temporal contexts during both feature extraction and similarity map refinement across a sequence of $t$ frames \cite{cao2022tctrack}.}
    \label{fig:exCNN-T}
\end{figure}

Several techniques have been proposed using CNN–Transformer for \ac{UAV}. For example as illustrated in Figure \ref{fig:exCNN-T}, \cite{cao2022tctrack} proposes TCTrack, a novel CNN-Transformer framework for \ac{UAV} object tracking, integrating temporal knowledge for enhanced performance. The framework consists of three main modules: TAdaCNN for online temporal-aware feature extraction, AT-Trans for refining the similarity map using adaptive temporal knowledge, and classification-regression heads for final prediction. Temporal contexts are modeled both before and after the correlation step, enabling robust tracking in dynamic aerial scenarios. TCTrack runs at 125.6 FPS on PC and over 27 FPS on NVIDIA Jetson AGX Xavier, demonstrating high accuracy and efficiency suitable for real-world \ac{UAV} applications. The framework was evaluated on four \ac{UAV} tracking benchmarks, using several metrics including precision, success rate, and FPS. The model achieved the best results with improvements of up to 5\% AUC over state-of-the-art methods. Similarly,  \cite{lu2024lightweight}  proposes LAPNet, a lightweight CNN–Transformer network with Laplacian loss for semantic segmentation of low-altitude \ac{UAV} imagery. It addresses the challenge of achieving accurate, real-time segmentation on devices with limited computing power, such as those used in \acp{UAV}. LAPNet combines a \ac{CNN}  encoder that captures local details using a multiscale kernel-sharing module, with a Transformer decoder that models global context through an efficient attention mechanism, which simplifies computation by removing the \textit{key} component. The Laplacian loss helps the network focus on boundaries and small objects. Three \ac{UAV} datasets were used, with \ac{mIoU}  as the main evaluation metric. LAPNet Large, the full version of the proposed model, achieved the best performance: over 15 \ac{FPS} on a mobile \ac{GPU} and 5.4\% higher accuracy than heavier models on the UAVid dataset. Similarly, \cite{ye2022ct} adopt the same approach, modifying the attention mechanism to develop a lightweight \ac{DL} model. The proposed scheme, named CT-Net, addresses key challenges in \ac{UAV} detection: improving accuracy, reducing computational complexity, and enabling real-time performance-—quantified through metrics like \ac{mAP} , GFLOPs, and \ac{FPS}. CT-Net integrates CNN-Transformer components to detect small, low-altitude objects such as \ac{UAV}s. Its main contribution lies in replacing the standard \ac{MSA} with a feature-enhanced \ac{MSA} (FEMSA), reducing computation while enhancing feature extraction. The lightweight bottleneck module minimizes parameters and computational load, and the directional feature fusion structure  enhances multiscale feature fusion, particularly improving small-object detection. CT-Net achieved an \ac{mAP}  1.6\% higher than
YOLOv5l on low-altitude dataset. Moving on, \cite{he2024dcd}  tackles challenges in \ac{UAV} localization caused by non-rigid deformations  ---perspective distortions, scale variations, shape inconsistencies, visual misalignment --- between \ac{UAV} and satellite images. To overcome this, \cite{he2024dcd}  introduces DCD-FPI, a fusion network that combines a CNN-Transformer backbone with deformable convolutions. The Transformer captures global context, while deformable convolutions adapt to geometric distortions. An adaptive pyramid fusion module (APFM) enhances \ac{MSFF}. However, 
\cite{zhao2023dstnet} proposes DSTNet, a CNN-Transformer hybrid network for cross-resolution vehicle re-identification. A dual-regression autoencoder first reconstructs low-resolution inputs. Then, DSTNet extracts identity features by fusing static local context from \acp{CNN}  with dynamic global context from Transformers, forming a static-to-dynamic pipeline. 

Huang et al. \cite{huang2023easy} proposes Easy-Net, a lightweight building extraction network tailored for remote sensing imagery to support \ac{UAV} navigation and urban analysis. Easy-Net introduces a CNN-Transformer fusion (CTF) block that integrates a double 3$\times$3 convolution for local detail enhancement and a mix-\ac{ViT} (MiT) block with overlapping patch merging for efficient global representation. Unlike the conventional \ac{ViT}, the MiT block is specifically designed for dense prediction tasks, employing convolution-based patch embedding and sequence reduction before applying self-attention, thereby preserving spatial structures while significantly reducing computational cost.  The work \cite{ghali2022deep} presents a \ac{DL} framework for wildfire detection and segmentation using \ac{UAV} imagery. For classification, it employs an ensemble model combining EfficientNet-B5 and DenseNet-201, enhancing feature diversity. For segmentation, the authors use two \acp{ViT}, TransUNet (a hybrid CNN-Transformer with ResNet50-ViT encoder) and TransFire (based on MedT with gated axial Attention), along with EfficientSeg model. Performance is assessed using accuracy, F1-score, and inference time. The ensemble classifier achieved 85.12\% accuracy, while TransUNet-R50-ViT reached the best segmentation result with an F1-score of 99.9\%, outperforming state-of-the-art models. However, the models rely on large datasets and pretrained Transformers for performance.


\vspace{0.5cm}

\noindent\textbf{-- RL-Transformer-based models:} enable adaptive decision-making in \ac{UAV} tracking tasks. These models combine the sequential modeling capability of Transformers with \ac{DRL}-driven policy optimization, allowing \acp{UAV} to learn optimal tracking strategies through interaction with dynamic environments (Trial-and-error learning). The working principle of \ac{DRL}-Transformer models can be summarized as follows:

\begin{equation*}
S_t \xrightarrow{\text{Feature Extractor}} Z_0 \xrightarrow{\text{Transformer}} Z_L \xrightarrow{\text{Policy Head}} \pi(a_t|S_t)
\end{equation*}

\noindent where $S_t$ represents the observed state at time step $t$, which may include visual features, motion cues, or environmental information. A feature extractor (e.g., \ac{CNN} or shallow encoder) processes $S_t$ into an initial token sequence $Z_0$. The Transformer applies \ac{MSA} and feed-forward operations to capture temporal dependencies and context across sequential states, producing $Z_L$. The policy head outputs a probability distribution $\pi(a_t|S_t)$ over possible actions $a_t$, guiding the \ac{UAV} to select optimal tracking decisions.


\begin{table*}[ht!]
\centering
\caption{Comparison of \ac{DRL} techniques used in Transformer-based \ac{UAV} systems.}
\label{tab:rl_comparison}
\scriptsize
\begin{tabular}{m{3.5cm}m{2cm}m{3.5cm}m{1.5cm}m{5cm}}
\hline
\textbf{DRL Technique} & \textbf{Learning type}  & \textbf{Transformer role} & \textbf{Complexity} & \textbf{Best Scenario Fit} \\
\hline
Offline, Policy-based \newline (sequence modeling) & Imitation / offline   & Sequence modeling for action prediction & Moderate & Low-data settings, pre-collected demonstrations, safe offline policy learning \\
\hline
\Ac{SAC} & \ac{AC} \newline (off-policy)  & Transformer encodes state-action for policy/value nets & High & Continuous control, exploration-demanding UAV missions \\
\hline
\Ac{PG}  & Policy-based \newline (on-policy)  & Transformer augments state representation for policy learning & Moderate & Flight path optimization, single UAV trajectory planning \\
\hline
\Ac{MADDPG} & \ac{AC} \newline (off-policy)  & Transformer models agent interactions or environment state & High & Cooperative tasks, swarm coordination, pursuit-evasion \\
\hline
Multi-agent \ac{AC} & \ac{AC} \newline (on/off-policy)  & Attention enhances agent communication / centralized critic & High & Distributed UAV control, coordinated surveillance \\
\hline
Multi-agent Transformer with virtual agent & \ac{AC} with Attention  & Transformer encodes all agents including virtual one for centralized policy & Very high & Communication-constrained environments, partial observability, emergent behavior modeling \\
\hline
\Acf{DQN} & Value-based & Transformer processes temporal or spatial observations & Low & Discrete action tasks like grid-based navigation or decision-making \\
\hline
\end{tabular}
\end{table*}

The paper \cite{jung2024maneuver} proposed a Maneuver-Conditioned Decision Transformer (MCDT) for \ac{UAV} tactical in-flight decision-making. It employed a decoder-only Transformer with causal self-attention over sequences of states, actions, returns, and maneuver tokens. The method followed offline, policy-based, imitation-style learning via supervised sequence modeling, avoiding traditional \ac{DRL} techniques. Evaluated on the artificial Dogfight dataset, MCDT outperformed \ac{SAC} and standard decision Transformers, especially in low-data regimes. Its limitations included lack of online adaptability, dependence on maneuver token discretization, and sensitivity to domain-specific configurations. Moving on, \cite{jiang2024autonomous} presented TOSRL, a Transformer-based observation sequence \ac{DRL} model for UAV navigation in a partially observable Markov decision process (POMDP) setting. It combined the \ac{SAC} strategy with a Transformer encoder—composed of positional encoding, \ac{MSA}, and feedforward layers—to extract temporal features from partial observations. TOSRL enabled obstacle avoidance and target tracking in dynamic environments, and outperformed recurrent \ac{SAC} (RSAC) in success rate, trajectory stability, and collision avoidance. The work proposed by Zhu et al.~\cite{zhu2022uav} introduced a novel Transformer-weighted A* (TWA*) algorithm to address the \ac{UAV} trajectory planning problem for age-of-information (AoI)-minimal data collection in \ac{UAV}-assisted internet-of-things (IoT) networks. The Transformer model adopted an encoder-decoder structure, leveraging \ac{MSA} to represent the UAV-IoT environment as a sequence. It was trained using a \Ac{PG}-based \ac{DRL} technique, known as REINFORCE, to optimize the \ac{UAV} visiting order. The weighted A* algorithm was then employed to refine the selection of hovering points. However, the method suffered from scalability issues, a fixed flight model, and the lack of dynamic re-planning during the mission. In \cite{li2023trans}, the authors presented a Transformer-enhanced multi-agent \ac{DRL} framework in which \ac{MADDPG} served as the foundational algorithm. \ac{MADDPG} enabled centralized training with decentralized execution, allowing \acp{UAV} to learn coordinated behaviors using a shared critic that considered global states and actions. The Transformer replaced traditional \ac{RNN}s in both the actor and critic networks, leveraging self-attention to model inter-agent and temporal dependencies. This architecture enhanced multi-task execution, removed action space constraints, and improved generalization in complex, dynamic \ac{UAV} environments. However, the proposed scheme was limited by simulation-based evaluation and the absence of real-environment uncertainty handling.

In \cite{gao2023sequential}, the authors proposed the MA2DBT algorithm, which reformulated each agent’s composite action space as a sequential decision process. This was handled by a Transformer architecture with self- and masked-attention mechanisms, enhanced by DeLighT modules for efficient depth. Integrated into a multi-agent actor–critic \ac{DRL} framework, MA2DBT reduced variance and improved coordination in multi-\ac{UAV} electronic warfare tasks. Similarly, in \cite{jiang2023short}, the work proposed MAT-V, a multi-agent \ac{DRL} algorithm for short-range UAV air combat, which introduced a virtual agent into the Transformer-based model. It enhanced the traditional multi-agent Transformer (MAT) by leveraging virtual interactions to improve cooperative decision-making. The model applied centralized training with decentralized execution and adapted attention mechanisms for air combat scenarios, achieving better convergence, higher win rates, and improved strategy diversity compared to baseline methods. The work in \cite{neves2024multimodal} proposed a multimodal \ac{UAV} landing system, called ViTAL, which combined a Transformer-based perception module and a \Ac{DQN} \ac{DRL} algorithm for autonomous decision-making. The Transformer integrated visual, thermal, and \ac{LiDAR} data for robust marker detection. Specifically, ViTAL adopted a \ac{ViT}-inspired architecture that combined a Transformer encoder with \ac{CNN}-based feature extraction and \ac{MLP} fusion in a multimodal design. A discretized \Ac{DQN} guided precise landings via 5 action choices in a 3D grid. The framework was validated in simulation and real-world conditions, showing accurate landings and resilience to environmental disturbances.

\subsection{Siamese and Swin-Transformer based UAV}

\noindent\textbf{-- Siamese-based models:} combine the matching capability of Siamese networks with the global contextual modeling strength of Transformers to enhance \ac{UAV} object tracking performance. These models leverage two parallel branches to extract feature representations from both the target template and the search region, enabling robust similarity learning under appearance variations and dynamic scenes. The working principle of Siamese--Transformer models can be summarized as follows:

\begin{equation*}
Z_T, Z_S \xrightarrow{\text{Transformer}} Z_L \xrightarrow{\text{Similarity Head}} y
\end{equation*}

\noindent where $Z_T$ and $Z_S$ denote the tokenized features extracted from the target template and the search region, respectively. Initially, convolutional backbones (e.g., \ac{CNN} or lightweight encoders) process the input template and search images to produce spatial feature maps, which are then flattened into token sequences. The Transformer block applies \ac{MSA} and cross-attention mechanisms to jointly model the interaction between $Z_T$ and $Z_S$, capturing global dependencies and enhancing feature alignment across frames. The output sequence $Z_L$ is then passed to a similarity head, typically implemented as a correlation or matching module, to produce the final tracking prediction $y$, which indicates the location of the target within the search region. This hybrid design allows Siamese--Transformer models to achieve accurate and robust \ac{UAV} tracking by integrating fine-grained feature matching with long-range contextual reasoning.

In light of previous developments, \cite{xing2022siamese2} proposed SiamTPN, a lightweight Siamese Transformer framework for \ac{UAV} tracking. Unlike traditional trackers, SiamTPN fuses multi-scale pyramid features through a centralized Transformer-based module before classification and regression. This design enables richer target-specific representations by modeling global context across scales. To avoid the high complexity of standard attention mechanisms, a pooling attention module was introduced, which reduces memory and computational load by downsampling keys and values. While extensive benchmarks confirm real-time performance on CPUs, a potential limitation is that lightweight backbones may struggle under extreme conditions such as small, fast-moving objects or severe appearance changes.

Subsequently, \cite{xing2022siamese} introduced SiamATN, a dynamic Siamese adaptive Transformer network also tailored for real-time \ac{UAV} tracking on resource-constrained platforms. It features a multi-stage architecture where each stage progressively enhances feature encoding using Transformer-based attention. The network adaptively selects the optimal stage based on scene complexity, measured by the KL-divergence between predicted and expected localization distributions. Shallow stages suffice for simple scenes, while deeper stages are triggered in complex scenarios. Additionally, spatial adaptive attention minimizes redundant computations. Experimental results demonstrate that SiamATN achieves state-of-the-art tracking performance at high CPU speeds, confirming its robustness and efficiency. Nevertheless, handling extremely rapid scene changes remains challenging despite the adaptive inference stages.

Building on these efforts, \cite{hu2023siamhsft} proposed SiamHSFT, a lightweight Siamese network-based tracker combining hierarchical sparse fusion, \ac{CBAM}, and Transformer enhancements. The \ac{CBAM} module refines feature fusion by emphasizing informative spatial and channel components, which is crucial for detecting small \ac{UAV} targets. Furthermore, SiamHSFT improves Transformer capabilities by incorporating a modulation enhancement layer using triplet attention, reinforcing inter-layer feature dependencies. Enhanced low- and high-level features are passed through a hierarchical Transformer to boost global context understanding. This architecture achieves superior robustness, accuracy, and real-time performance across challenging benchmarks. However, the increased model complexity may slightly impact deployment on resource-limited \ac{UAV}s. Similarly, \cite{wang2023learning} embedded Transformer-based self-attention within the Siamese framework by replacing traditional feature correlation with feature concatenation followed by attention, thus enabling global interaction. This modification improved tracking precision while maintaining a lightweight design suitable for \ac{UAV}s. The model demonstrated real-time capability and improved generalization, though it encountered challenges under extreme \ac{UAV} dynamics, heavy occlusions, and adverse lighting conditions.

In addition to tracking, Transformer-based Siamese frameworks have also been applied to change detection tasks. \cite{zhai2024cas} introduced CAS-Net, a Siamese network tailored for \ac{UAV} change detection. It incorporates Transformer-inspired contrastive attention modules (CAMs) at each Siamese stage. These modules apply local and global contrastive attention to effectively capture fine-grained spatial differences. CAS-Net fuses bitemporal features using CAMs and a difference augmenting module (DAM), which compresses and refines the representation for final classification. This enhances robustness and accuracy, especially in the presence of image noise in UAV datasets.

Lastly, \cite{chen2024spatial} proposed SiamST, which connects Siamese networks with Transformer principles to improve \ac{UAV} tracking. It employs a two-stage enhancement: a high-order multiscale spatial module for refined spatial feature extraction and a temporal template Transformer that adaptively updates the template using temporal context. Unlike traditional fixed-template Siamese models, this hybrid design allows dynamic and fine-grained spatial-temporal feature learning. However, SiamST still struggles under severe occlusions and lacks a trajectory prediction mechanism—highlighting areas for future improvement.


\vspace{0.5cm}


\noindent\textbf{-- Swin--Transformer-based models:} employ a hierarchical \ac{ViT} architecture with shifted window-based self-attention to efficiently capture both local and global context for \ac{UAV} tracking tasks. These models enhance feature representation by computing self-attention within non-overlapping local windows while enabling cross-window connections through window shifting, providing a balanced trade-off between accuracy and computational efficiency. The working principle of Swin--Transformer models can be summarized as follows:

\begin{equation*}
X \xrightarrow{\text{Patch Partition}} P_0 \xrightarrow{\text{Swin Transformer Blocks}} P_L \xrightarrow{\text{Head}} y
\end{equation*}

\noindent where $X$ denotes the input image or search region, which is partitioned into non-overlapping patches to form the initial sequence $P_0$. The Swin Transformer blocks apply \textit{window-based} \ac{MSA} and \textit{shifted window-based} \ac{MSA} to capture both local features and long-range dependencies across the image. Through a hierarchical structure, the feature maps are progressively merged, reducing spatial dimensions while increasing feature richness, resulting in $P_L$. Finally, a task-specific output head processes $P_L$ to generate the prediction $y$, such as the target location or class score. Swin--Transformer-based models are particularly suitable for \ac{UAV} tracking due to their scalability, efficient computation, and powerful representation ability across varying object scales, motions, and complex aerial scenes.

\begin{figure}
    \centering
    \includegraphics[width=0.9\linewidth]{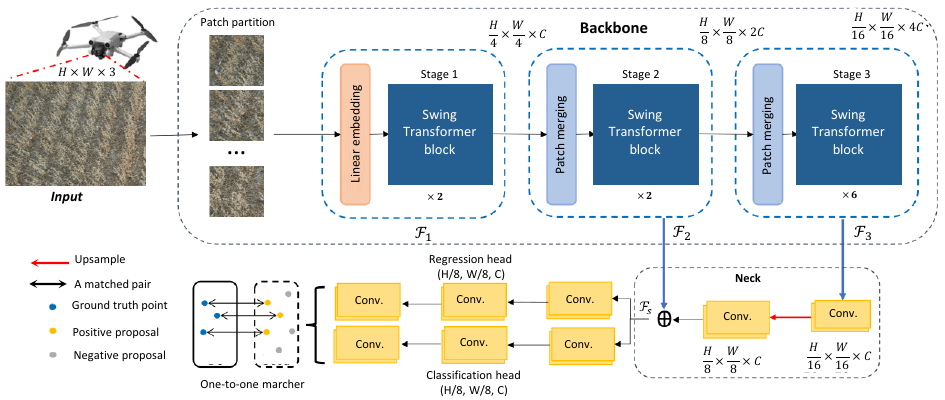}
    \caption{The architecture of SoybeanNet, designed for counting soybean pods, is built upon a Transformer-based backbone. It extracts deep feature maps using upsampling and skip connection techniques. Subsequently, two separate head branches — one dedicated to regression and the other to classification — generate a set of predicted point proposals accompanied by their confidence scores. To ensure accurate correspondence between the predicted points and the ground truth, a one-to-one matching strategy is applied \cite{li2024soybeannet}.}
    \label{fig:SwinExample}
\end{figure}

Swin-Tiny (Swin-T) is the most commonly employed variant of the Swin Transformer in \ac{UAV} vision tasks due to its favorable trade-off between performance and computational efficiency, making it particularly suitable for edge devices onboard drones. For example, Zhang et al. in \cite{zhang2023swint} propose the SwinT-YOLO architecture, in which Swin-T replaces the conventional convolutional backbone in YOLOv4 to enhance the detection of densely clustered maize tassels in \ac{UAV} imagery. The model further integrates depthwise separable convolutions to keep computational costs low while improving accuracy. Similarly, Swin-T is used in Swin-T-NFC CRFs scheme \cite{wang2023swin} as the encoder in a dual-task architecture that fuses point cloud super-resolution with image semantic segmentation. The use of shifted windows, pyramid pooling, and fully connected conditional random fields (CRFs) allows the model to capture both fine local structures and global layout for precise \ac{UAV} positioning. In \cite{meng2024swin}, Swin-T is embedded within a Fast R-CNN pipeline, enhanced with a \ac{CAM} and binary self-attention. These modifications significantly improve object localization in cluttered aerial environments, such as electrical infrastructure. Across these studies, Swin-T is consistently selected for its balance of lightweight design and powerful hierarchical feature extraction, especially when further enhanced by modular attention techniques. Swin-T offers efficiency but has limited representational capacity, making it less effective in complex scenes. It often requires auxiliary modules to perform well and may miss broader context due to its smaller receptive field.

Swin-B (Base) and other general-purpose Swin variants offer higher model capacity and are often chosen for more demanding \ac{UAV} perception tasks where model complexity is acceptable. For example, FEA-Swin \cite{xu2022fea} introduces a foreground enhancement attention mechanism atop a standard Swin backbone, combining it with a bi-directional \ac{FPN} and skip connections to boost detection accuracy of small and densely packed objects in \ac{UAV} imagery. In  \cite{lu2023automated}, Swin replaces the \ac{CNN} backbone within a Mask R-CNN framework to improve segmentation of hazardous areas at construction sites. The hierarchical attention and shifted windows in Swin allow for fine-grained recognition of weak terrain patterns that signal danger zones. SoybeanNet \cite{li2024soybeannet} also adopts a general Swin-based backbone as a feature extractor in a point-based counting framework for soybean pods. The model capitalizes on Swin’s ability to learn spatial hierarchies in complex field conditions, improving counting precision under natural lighting and dense foliage. Figure \ref{fig:SwinExample} illustrates the proposed SoybeanNet model, which is based on the Swin-B (Base) architecture.  Collectively, these implementations show how Swin-B and its unmodified variants serve as versatile and powerful replacements for \acp{CNN} in \ac{UAV} visual understanding tasks, especially when enriched with auxiliary modules or fused with dense prediction heads. However, Swin-B increases computational and memory demands, limiting real-time \ac{UAV} deployment. It also requires careful tuning to avoid overfitting in data-scarce environments.

Several studies have extended the original Swin architecture into novel variants tailored for real-time, low-power, or clutter-intensive \ac{UAV} settings. One notable adaptation is the Ghost Swin Transformer, introduced in forest fire detection \cite{wang2024forest}. This model combines Swin Transformer with Ghost convolution modules and a reparameterized rotational attention mechanism to enable efficient multi-scale feature extraction, particularly useful for detecting small flames and obscured smoke in forest imagery. Additionally, the inclusion of a custom loss function, EAGIoU, improves localization precision and convergence. In a different context, \ac{UAV} swarm tracking \cite{wang2024target} incorporates Swin blocks into the network neck and supplements them with an \ac{ECA} module to boost tracking accuracy of multiple drones. This architecture significantly improves data association and trajectory completeness under dynamic aerial scenarios. These customized Swin variants underscore an emerging trend: modifying the Swin architecture with lightweight modules and context-aware attention mechanisms can dramatically enhance its robustness in real-time \ac{UAV} applications without compromising performance. Nevertheless, these customized variants add design complexity and may not generalize well across \ac{UAV} tasks. Their performance is sensitive to training data quality and may drop on low-resolution inputs.

Linear Swin Transformer represents a more computationally efficient evolution of the original design, aimed at reducing the quadratic complexity of standard self-attention while retaining hierarchical learning capabilities. In \cite{li2024lswinsr}, a linear Swin Transformer architecture is proposed for \ac{UAV} image super-resolution. This model utilizes residual linear Swin blocks (RLSTB) to reconstruct high-resolution imagery from low-resolution \ac{UAV} inputs, achieving excellent visual fidelity while maintaining low computational demand. Beyond improving traditional image quality metrics like \ac{PSNR} and \ac{SSIM}, the model also demonstrates downstream benefits, particularly in enhancing the accuracy of semantic segmentation tasks on super-resolved \ac{UAV} images. This shows that Linear Swin is not only lightweight but also well-suited for cascading vision tasks where resource efficiency and image enhancement are both critical. While efficient, Linear Swin may underperform in capturing fine details due to simplified attention. It also lacks the maturity and support of full-attention models.

Swin Transformer’s hierarchical attention and patch-based representation also make it an ideal encoder in multimodal fusion networks. A compelling example is the Swin-UetFuse model developed in \cite{li2023observing}, which fuses visible and infrared \ac{UAV} imagery for ecological monitoring. The model applies Swin as a dual-stream encoder, using patch-expanding layers and skip connections to fuse features across modalities while preserving both spectral specificity and spatial resolution. This fusion enables simultaneous recognition of animal movement and environmental context, which is particularly useful for tracking endangered species in forested environments. The Swin-based fusion strategy demonstrates the model’s strength in integrating heterogeneous data sources, making it highly valuable for multi-sensor \ac{UAV} platforms tasked with comprehensive environmental perception. Multimodal Swin models face challenges in aligning different sensor inputs and often require complex calibration. Their dual-stream design may also hinder real-time processing on \acp{UAV}.

\subsection{Unsupervised Transformer-based UAV}

This kind of Transformer is designed to extract meaningful patterns from sequential aerial data without relying on labeled annotations. These models typically operate through predictive tasks such as next-frame reconstruction, motion forecasting, or masked input recovery, enabling them to autonomously learn representations that support downstream tracking, detection, or anomaly identification. The prediction Transformer-based \ac{UAV} can be summarized as follows: 

\begin{equation*}
\{X_1, X_2, \dots, X_T\} \xrightarrow{\text{Encoder}} Z \xrightarrow{\text{Prediction Transformer}} \hat{X}_{T+1:T+k}.
\end{equation*}

In this structure, the model encodes an input sequence into latent features \( Z \), and then generates predicted outputs \( \hat{X}_{T+1:T+k} \), capturing progression and continuity across \ac{UAV}-captured sequences.

The use of Transformers for modeling sequential \ac{UAV} data for unsupervised task has been investigated in \cite{wu2023developing} and \cite{ahmad2024transformer}. The study investigated by Wu et al. \cite{wu2023developing} represents one of the earlier efforts in this domain. It employs a \ac{TST} model to predict vehicle speeds based on lane-level data extracted from \ac{UAV} videos, utilizing \ac{YOLO} for object detection and Deep-SORT for tracking. While the method effectively captures temporal dependencies better than \ac{LSTM}-based approaches, its primary limitation lies in its domain specificity. The model is tailored to freeway interchanges in Chinese infrastructure and may not generalize well to different road geometries, traffic cultures, or environmental conditions without significant retraining. Additionally, the system relies heavily on accurate object detection and tracking performance, which may degrade under occlusions or poor \ac{UAV} video quality. A more recent study by Ahmad et al. \cite{ahmad2024transformer} introduces a valuable improvement by using Attention-based networks to predict and classify different time-ordered  sensor failures in \acp{UAV}. The study also contributes the BASiC dataset, which includes data from sensors such as \ac{GPS}, remote control, accelerometer, gyroscope, compass, and barometer, summarized into 17 key features. Its Transformer model enables early detection, which is critical for autonomous flight safety. However, the study’s limitation lies in the complexity of real-world \ac{UAV} deployments. The dataset, while extensive, is simulated or semi-structured, and may not account for edge-case sensor failures encountered in high-stress or long-duration flights. The method also presumes consistent sensor sampling rates and synchronization, which can be a fragile assumption in multi-vendor \ac{UAV} platforms.

\acp{ViT} have also made their way into unsupervised \ac{DA} tasks in \ac{UAV} imaging. For example, the paper \cite{dos2022unsupervised} proposes a novel strategy to reduce manual labeling using dilation-based pseudo masks and Transformer-based semantic segmentation,   DAFormer \cite{hoyer2022daformer}, the authors show improved generalization across farms with different visual characteristics. However, this method has two key limitations. First, the quality of pseudo-labels depends on initial manual annotations and dilation parameters, which may not always reflect accurate plant boundaries. Second, while the model generalizes across domains, it still requires a reasonably aligned visual structure (e.g., crop row patterns), limiting its scalability to unstructured vegetation or irregular planting schemes. On the more advanced side of \ac{DA}, the  study \cite{wei2024unsupervised} addresses the challenging unsupervised task of \ac{UAV}-based tracking at night by adapting knowledge from labeled daytime datasets. The architecture—TransffCAR—incorporates Swin Transformer V2 modules, hierarchical feature fusion, and adversarial learning to tackle domain shift. Its major strength is the ability to operate in low-light and occlusion-heavy environments. However, it has significant computational demands due to multiple Attention layers and fusion mechanisms, making real-time deployment on edge \ac{UAV} devices difficult. Additionally, the model’s performance still partially hinges on the accuracy of initial pseudo-label generation, which remains susceptible to visual noise or poor initialization.


\subsection{Spatio-Temporal Transformer-based UAV}

\ac{STT}-based models extend conventional Transformer architectures to jointly model spatial and temporal dependencies for \ac{UAV} tasks. These models are specifically designed to handle video sequences or multi-frame data, where spatial feature extraction and temporal dynamics modeling are equally crucial for accurate and robust target tracking in aerial environments. The working principle of \ac{STT} models can be summarized as follows:

\begin{equation*}
\{X_1, X_2, \dots, X_T\} \xrightarrow{\text{Spatial Encoder}} Z_0 \xrightarrow{\text{Temporal Transformer}} Z_L \xrightarrow{\text{Head}} y
\end{equation*}

\noindent where $\{X_1, X_2, \dots, X_T\}$ represents the sequence of input frames over $T$ time steps. A spatial encoder (e.g., CNN or patch-based embedding) processes each frame $X_t$ to extract local spatial features, which are then aggregated into the initial token sequence $Z_0$.

The Temporal Transformer applies \ac{MSA} across the temporal dimension to capture motion cues, long-term dependencies, and target trajectory patterns. This enables the model to understand object motion, appearance variations, and contextual scene dynamics across frames. The final representation $Z_L$ is processed by a task-specific head to output $y$, which may correspond to the predicted target location, object class, or motion estimation. \ac{STT}-based models are particularly effective in \ac{UAV} tracking scenarios involving rapid motion, occlusion, and long-term target dynamics, as they leverage both intra-frame spatial structure and inter-frame temporal continuity.

\begin{figure}
    \centering
    \includegraphics[width=0.9\linewidth]{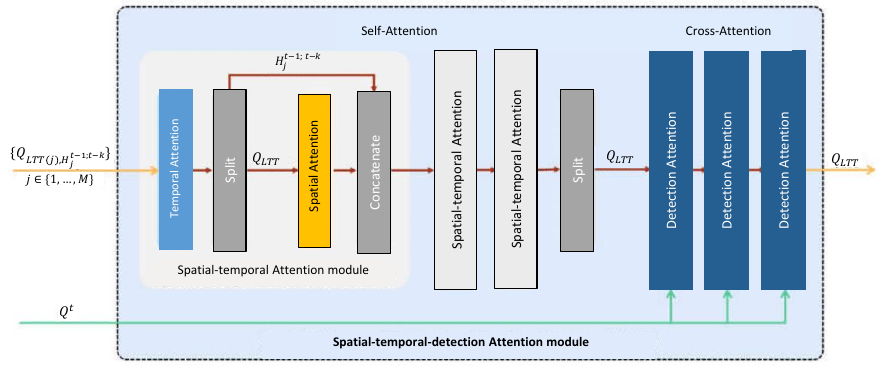}
    \caption{Example of a \ac{STT}. It concatenated the learnable trajectory tokens \( Q_{\text{LTT}} \) with the historical tracking features \( H^{t-1:t-k} \), then iteratively updated \( Q_{\text{LTT}} \) through spatial-temporal attention layers. The updated tokens \( Q_{\text{LTT}} \) interacted with the detection features \( O_t \) via cross-attention, and the STD module output the final refined trajectory tokens \( Q_{\text{LTT}} \) \cite{hu2023stdformer}.}
    \label{fig:STT}
\end{figure}

Hu et al.~\cite{hu2023stdformer} proposed a \ac{STT} model named STDFormer for multi-object \ac{UAV} tracking, which relied solely on motion cues, without using appearance features. At the core of the design was the spatial-temporal-detection (STD) module, which sequentially integrated temporal, spatial, and detection information through attention layers to enable precise motion modeling. A learnable trajectory token aggregated historical dynamics to strengthen object association. The model was evaluated on \ac{UAV} datasets, where it demonstrated strong tracking capabilities under complex motion and occlusion conditions. STDFormer thereby highlighted the effectiveness of Transformer architectures for motion-driven \ac{UAV} tracking. Figure~\ref{fig:STT} illustrates the principle of the STD module as an example of a \ac{STT}.

To address persistent challenges in \ac{UAV} tracking, such as fast motion, occlusion, scale variation, and low resolution, Zhu et al.~\cite{zhu2023spatio} introduced an efficient  spatio-temporal hierarchical feature Transformer (STHFT). Unlike local correlation-based methods, STHFT fused multi-level spatio-temporal features via attention mechanisms. The encoder processed high-resolution inputs to capture spatial-temporal interdependencies, while the decoder generated low-resolution semantic features. An information fusion layer further enhanced robustness across spatial scales. This architecture significantly improved global context modeling, achieving superior tracking performance without post-processing.

Similarly, Siyuan et al.~\cite{siyuan2024learning} proposed a \ac{STT} for short-term 2-D \ac{UAV} trajectory forecasting in vision-based systems. To mitigate overfitting in small datasets, the authors introduced a degenerate self-attention mechanism, characterized by lightweight, single-head attention without normalization layers. This design captured key spatial–temporal dependencies efficiently. By independently modeling azimuth and elevation dimensions, the model learned nonlinear variation patterns representative of complex \ac{UAV} maneuvers. It achieved accurate trajectory forecasting in detection-only environments, overcoming the limitations of lacking tracking supervision.

Lastly, Shen et al.~\cite{shen2023adaptive} proposed DUTrack, a dual-template Transformer-based tracker for \ac{UAV} multi-object tracking. The architecture included a memory dynamic update module, which adaptively refreshed the dynamic template using spatio-temporal cues from previous frames. This mechanism improved robustness to occlusions and appearance variations. In parallel, a spatio-temporal context mapping module fused response maps through an encoder-decoder Transformer structure, enabling resilient temporal continuity and accurate tracking in complex aerial environments.


\subsection{ViT and YOLO-based UAV}

\noindent\textbf{-- ViT--based models:} directly apply Transformer architectures to image patches without relying on convolutional layers, enabling global feature modeling from the earliest stages of the network. In \ac{UAV} tracking tasks, \ac{ViT}-based models process the target template and search region using patch embedding and self-attention mechanisms to capture global contextual information and improve tracking robustness against challenging aerial scenarios. The working principle of \ac{ViT}--based models can be summarized as follows:

\begin{equation}
X \xrightarrow{\text{Patch Partition}} P_0 \xrightarrow{\text{Linear Embedding}} Z_0 \xrightarrow{\text{Transformer}} Z_L \xrightarrow{\text{Head}} y.
\end{equation}

\noindent where, $X$ represents the input image (either the target template or search region). The image is divided into non-overlapping patches, which are flattened and linearly projected to form the initial token sequence $P_0$. These tokens, enriched with positional encodings, constitute $Z_0$. A standard Transformer encoder, composed of multiple layers of \ac{MSA} and feed-forward networks, processes $Z_0$ to produce the final representation $Z_L$. The output head then generates the tracking prediction $y$, such as the target location within the search region. In Table \ref{table:8} we provides a detailed summary of \ac{ViT}-based \ac{UAV}.

To illustrate, the work in \cite{li2023adaptive} introduced Aba-\ac{ViT}, representing the first attempt to apply efficient \ac{ViT}s for real-time \ac{UAV} tracking. Specifically, the Aba-\ac{ViT} framework leveraged adaptive and background-aware token computation, halting background tokens earlier based on learned probabilities. This strategy significantly reduced computational overhead. Furthermore, by generalizing the ponder loss to assign higher weights to background tokens, the method enhanced token sparsification and efficiency. In comparison with traditional approaches, Aba-\ac{ViT} unified feature learning and template-search coupling into a streamlined, efficient pipeline, eliminating the need for separate or computationally heavy modules.

Similarly, the work in \cite{munyer2022foreign} focused on \ac{UAV} applications by proposing a novel approach for detecting \ac{FOD} on airport runways. Unlike the supervised techniques discussed earlier, which relied on extensive annotated datasets, this method utilized self-supervised learning with \ac{ViT}s. By learning from clean runway images, it effectively localized anomalies as potential debris, addressing the challenge of arbitrary object detection. Additionally, the integration of \ac{ViT}s boosted both localization accuracy and generalization across various environments. Moreover, the introduction of an economical data collection framework, using uncrewed aircraft systems to gather high-resolution images, highlighted the practical applicability of this method for real-world scenarios.

Building on the theme of efficient \ac{UAV} tracking, the paper \cite{lilearning} presented AVTrack, an advanced framework that optimized computational efficiency while maintaining high tracking precision. Through its activation module, AVTrack dynamically activated Transformer blocks, thereby addressing computational constraints. To overcome challenges like extreme viewing angle variations, it incorporated \ac{MI} maximization, enabling robust, view-invariant representations. Consequently, this framework achieved real-time tracking speeds without compromising accuracy. Notably, evaluations on five benchmarks revealed AVTrack's superior precision and efficiency, positioning it as a state-of-the-art solution. Expanding further, the framework introduced in \cite{ferdous2022uncertainty} took a different angle by addressing uncertainty-aware \ac{MTL} for \ac{UAV}-based object re-identification (ReID). Unlike the previously mentioned methods, this approach employed a convolution-free architecture with a modified pyramid \ac{ViT} (PVT) backbone to extract hierarchical features across varying altitudes and camera angles. Furthermore, the integration of spatial and \ac{CAM}s enhanced feature discrimination while reducing noise. Additional components, such as batch instance normalization (BIN) and multitask optimization with uncertainty-aware recognition, ensured robust performance. Evaluations on PRAI-1581 and VRAI datasets reinforced the framework’s effectiveness under challenging aerial surveillance conditions.

In contrast to convolution-based models, the study in \cite{zhu2021vitt} pioneered the use of Transformers for \ac{MOT}. The \ac{ViT} Tracker (ViTT) utilized a pure Transformer encoder as its backbone, allowing direct image input processing. Moreover, ViTT addressed issues like occlusion and complex scenarios through global context modeling. By combining object detection and embedding extraction in a unified multi-task learning framework, this method demonstrated improved performance, as shown in its results on the MOT16 dataset.

The study \cite{castellano2023weed} took a novel direction by applying lightweight \ac{ViT}s for weed mapping in precision agriculture. Unlike the other applications discussed, this approach focused on semantic segmentation of multispectral drone imagery. By adapting pre-trained RGB weights to non-visible channels such as near-infrared and red edge, the proposed models achieved exceptional performance on the WeedMap dataset. Additionally, variants like DoubleLawin and SplitLawin optimized inference time and segmentation accuracy, making them well-suited for resource-constrained \ac{UAV} platforms. This method not only advanced sustainable farming practices by reducing herbicide usage but also demonstrated strong generalization across diverse agricultural fields, emphasizing its scalability and efficiency.

\begin{figure}
    \centering
    \includegraphics[width=0.9\linewidth]{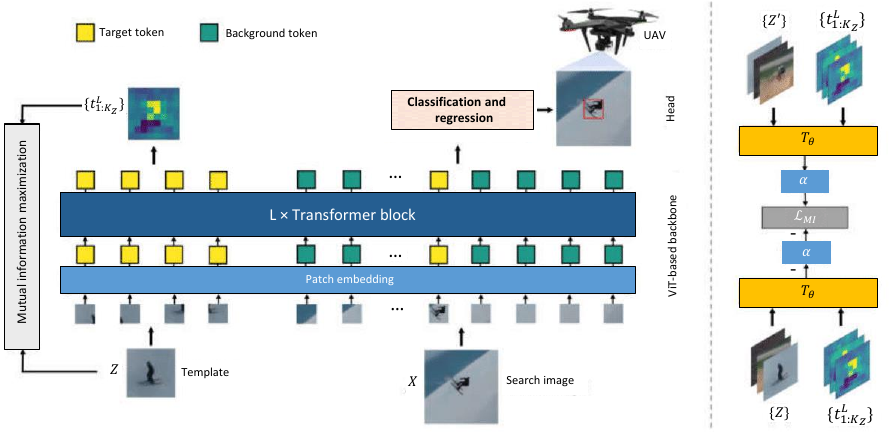}
 \caption{
As an example of a ViT-based \ac{UAV} tracker. The left part shows the overall framework, which consists of a single-stream ViT backbone that jointly performs feature learning and template--search coupling from the input template \( Z \) and search image \( X \), followed by a localization head to predict the target bounding box. The right part illustrates the \ac{MI} (MI) maximization module used during training to enforce target awareness. Here, \( \{Z\} \) denotes a batch of template images, and \( \{Z'\} \) is a randomly shuffled version used to simulate the marginal distribution. These are processed alongside their corresponding token embeddings from the ViT to compute \ac{MI} via a neural estimator \( T_\theta \), which maps input pairs to scalar scores. The softplus function \( \alpha(z) = \log(1 + e^z) \) is applied to these scores as part of the Jensen--Shannon divergence-based \ac{MI} estimator. The final \ac{MI} loss \( L_{\text{MI}} \) is defined as the negative estimated \ac{MI} and is minimized to maximize the dependency between the template and its learned representation, thereby enhancing the model’s ability to preserve critical target information amidst abundant background tokens 
\cite{li2024learning}.}
    \label{fig:exViT}
\end{figure}

In addition, \cite{li2024learning} introduced TATrack illustrated in Figure \ref{fig:exViT}, a Transformer-based \ac{UAV} tracker that unified feature learning and template–search coupling within a one-stream ViT backbone. To tackle the degradation of target information due to dominant background tokens, it maximized \ac{MI} between the template and its learned representation during training. Additionally, an MI-based knowledge distillation method compressed the model for improved efficiency. While it achieved state-of-the-art precision and speed, limitations included dependency on carefully tuned \ac{MI} loss weighting and decreased robustness in extreme occlusion or ambiguous background scenarios.

\vspace{0.5cm}

\noindent\textbf{-- YOLO-Transformers-based:}  combine the real-time object detection capability of the \ac{YOLO} framework with the global context modeling power of Transformers to enhance \ac{UAV} tracking and detection tasks. These hybrid models benefit from the fast and dense prediction mechanism of \ac{YOLO} while incorporating Transformer layers to improve feature representation, object localization, and robustness in complex aerial environments. The working principle of YOLO--Transformer models can be summarized as follows:

\begin{equation}
X \xrightarrow{\text{Backbone (CNN)}} F_0 \xrightarrow{\text{Transformer Encoder}} F_L \xrightarrow{\text{Detection Head}} y.
\end{equation}

In this illustration, $X$ denotes the input image, typically from the \ac{UAV} onboard camera. The CNN-based backbone extracts low-level and mid-level feature maps $F_0$. The Transformer encoder is then employed to enhance these features by modeling long-range dependencies and global relationships across different regions of the image, producing the refined feature representation $F_L$. The detection head, often following the \ac{YOLO} paradigm, performs multi-scale prediction to output $y$, which may include bounding box coordinates, objectness scores, and class probabilities — allowing the system to detect and track objects within the aerial scene efficiently.

Classic \ac{YOLO} models (YOLOv1 to YOLOv4) have been widely used in \ac{UAV}-based object detection due to their real-time performance and lightweight CNN-based architecture. However, they do not incorporate Transformer mechanisms. From YOLOv5 onward, Transformer-based YOLO variants have emerged, enhancing \ac{UAV} tracking and detection tasks by integrating attention modules and Transformer components. These improvements provide better context modeling and robustness, making them more suitable for complex aerial scenes where precise object localization is critical.

A number of works have investigated enhancing YOLOv5 with Transformer-based components. For instance, the study \cite{li2023mcd} introduced MCD-YOLOv5, an enhanced YOLOv5-based model that integrated three components: multi-level fusion feature weighting, \ac{CBAM}, and a deep effective Transformer fusion (DETF) mechanism. These improvements enhanced real-time pest and disease detection in \ac{UAV}-acquired images. The model demonstrated superior performance on a fusion dataset, highlighting the synergy between \acp{UAV} and Transformer-augmented YOLOv5 for agricultural applications. Moreover, \cite{zhao2022concrete} proposed a Swin Transformer-enhanced YOLOv5s-HSC model tailored for structural damage detection in concrete dams. The integration of Swin Transformer blocks and \ac{CA} modules significantly improved detection under complex visual conditions. The model, supported by photogrammetric 3D reconstruction, achieved a 3.8\% improvement in mean \ac{AP}, offering spatially precise damage visualization.  Furthermore, \cite{xing2023improved} enhanced YOLOv5 for pavement crack detection using \acp{UAV}, integrating a Swin Transformer in the backbone and a bi-directional \ac{FPN} in the neck. This architecture achieved real-time performance while detecting cracks as narrow as 1.2 mm, although its effectiveness decreased under noisy or low-resolution conditions. In addition, \cite{zhu2024natca} proposed NATCA YOLO, a YOLOv5-based model augmented with a NAT Transformer and \ac{CA} to capture global and long-range spatial-channel dependencies. The model significantly enhancing detection performance for small objects in aerial scenes. The authors in \cite{wang2023consumer} introduced SR-YOLO, a Transformer-augmented YOLOv5/v8 model featuring a recursive bottleneck Transformer, CARAFE upsampling, and Shape-CIoU loss. This configuration achieved high \ac{mAP} on dense-scene datasets such as VisDrone and TinyPerson, albeit at the cost of increased computational complexity, limiting its real-time deployment on consumer-grade \acp{UAV}. 

The combination of YOLOv6 and Transformer modules has been the focus of multiple recent studies. For example, the study \cite{li2023object} improved YOLOv6 for detecting small, densely packed targets in \ac{UAV} imagery. It incorporated a Transformer-based prediction head (TPH) leveraging a Transformer encoder block (TEB) for global context capture. Moreover, it introduced a multi-anchor assignment (MAA) mechanism, \ac{FPN} for feature alignment, and additional modules to mitigate false positives and enhance feature integration across scales, effectively boosting detection accuracy with minimal computational overhead.

Transformer-enhanced YOLOv7 models have been widely adopted in \ac{UAV}-based research. To illustrate, The work  in \cite{liao2024improved} enhanced YOLOv7 for wind turbine blade damage detection using \ac{UAV} imagery. It introduced an \ac{ECA} module, a down-sample block, and a dynamic group convolution shuffle Transformer (DGST) to improve the extraction of contour, texture, and subtle damage features. As a result, the proposed model outperformed subsequent YOLO models in terms of accuracy, speed, and robustness. Moreover, \cite{zhang2023ats} proposed ATS-YOLOv7, an enhanced model integrating an adaptive fusion \ac{FPN} and a Transformer-based detection head. These components significantly improved multi-scale detection, particularly in cluttered and occluded \ac{UAV} scenes. A new SIoU  loss with angular regression further accelerated convergence and localization precision. Similarly,  \cite{zhao2023ms} introduced MS-YOLOv7 for small object detection in \ac{UAV} imagery. The model embedded a Swin Transformer for global feature awareness, \ac{CBAM} for spatial-channel refinement, and a Soft-NMS algorithm to reduce detection overlap errors. The use of Mish activation further contributed to improved convergence and expressiveness. Besides, \cite{ma2024uav} presented a two-stage pipeline combining YOLOv7 with $\alpha$-\ac{IoU} loss for initial localization and a ResNet-ViT-CAM hybrid classifier for species recognition in forest-based Cervidae monitoring. The integration of \ac{ViT} into the classification stage enabled global context modeling, reducing background interference and enhancing recognition accuracy.

For YOLOv8s-Transformer,  \cite{do2023human} introduced a framework for human detection using \ac{UAV} imagery. It incorporated the SC3T Transformer backbone to enhance global feature awareness and contextual sensitivity. Additionally, \ac{SPP} modules were employed to improve multi-scale feature capture under complex illumination and environmental variability. Similarly, \cite{tahir2024pvswin} proposed a combining YOLOv8s with Swin Transformer architecture in the backbone, and \ac{CBAM} in the neck. Furthermore, non-maximum suppression  (Soft-NMS) post-processing step replaced the standard non-maximum suppression to address object occlusion, enabling improved pedestrian and vehicle detection in smart city surveillance scenarios. In addition, \cite{wang2023consumer} hybridized YOLOv8 with recursive Transformer modules and shape-aware loss functions in the SR-YOLO model, targeting densely packed scenes. Although the model offered significant accuracy improvements, its high complexity constrained real-time performance on embedded \ac{UAV} platforms.

\begin{table}[t!]
\centering
\caption{Summary of ViT, YOLO, and LLM-based \ac{UAV} methods. }
\scalebox{0.74}{
\begin{tabular}{m{0.5cm} m{2cm} m{6.5cm} m{2cm} m{2.5cm} m{6cm}m{1cm}}
\label{table:8} \\
\hline
\textbf{Ref} & \textbf{AI method} & \textbf{Contributions} & \textbf{Used data} & \textbf{Best result } & \textbf{Limitations} & Code\\
\hline

\cite{li2023adaptive} & Aba-ViT & Exploration of efficient adaptive and background-aware ViTs for \ac{UAV} tracking. & UAV123 & Pre= 86.4 \newline AUC= 66.4  & Dependency on predefined distributions for halting tokens. & \href{https://github.com/xyyang317/Aba-ViTrack}{Github}\\
\hline

\cite{munyer2022foreign} &  ViT & Proposed a self-supervised \ac{FOD} detection method. & FOD images & Acc= 99.94 \newline DR= 82.7 & Dependent on high-resolution data; lighting and shadow variations. & --- \\
\hline

\cite{lilearning} & ViT & Developed AVTrack for real-time UAV tracking. & VisDrone2018 & Pre=86.0 \newline Speed= 220 FPS & Potential efficiency trade-offs in resource-constrained environments. & \href{https://github.com/wuyou3474/AVTrack}{Github}\\
\hline

\cite{ferdous2022uncertainty} & Pyramid \ac{ViT} & Designed \ac{MTL} framework with uncertainty modeling for UAV object ReID. & PRAI-1581 and VRAI & mAP= 82.86 \newline Rank-1= 84.47 & Limited generalization to highly variable environmental conditions. &  --- \\
\hline

\cite{zhu2021vitt} & ViT  & Multi-task learning MOT model using a Transformer backbone. & MOT16 & MOTA=  65.7 \newline Speed= 15 FPS & Requires large datasets for optimal performance. & \href{https://github.com/jiayannan/VITT}{Github}\\
\hline

\cite{castellano2023weed} & ViT & Developed Lightweight ViT for multispectral mapping using RGB transfer learning. & WeedMap  & F1 = 86.50 & Lower generalization on eschikon subset due to fewer weed samples. & \href{https://github.com/pasqualedem/LWViTs-for-weedmapping}{Github}\\
\hline
\cite{li2024learning} & ViT & Introduced efficient mutual information-based Attention for accurate real-time UAV tracking & VisDrone2018 & Pre= 88\newline Sr= 66.9  & Struggles with occlusion, similar objects, fog, and fast motion. & \href{https://github.com/xyyang317/TATrack}{Github}\\
\hline
\cite{li2023mcd} & MCD-Yolov5 & Optimized UAV model for real-time pest and disease detection in agricultural environments. & Plant-village and rice data & Acc= 95.7 & Limitations with occlusion and lighting variations in UAV-collected images. & --- \\
\hline

\cite{zhao2022concrete} & YOLOv5s-HSC & Developed a YOLO-Swin Transformer  for concrete dam damage detection and localization. & UAV dams images & mAP=  79.8 \newline Speed= 17 FPS & Generalizability limits; skewed boxes, blurred boundaries reduce accuracy. & --- \\
\hline

\cite{li2023object} & YOLOv6-ViT & Improved detection of small, dense targets by using ViT into the model architecture. & VisDrone2019-DET & AP$_{50}$= 59.73 & High computational complexity and potential overfitting. &  --- \\
\hline

\cite{liao2024improved} & YOLOv7-DGST & Developed a robust model that improves UAV-based wind turbine blade damage detection. & WTB damage image &  mAP$_{50}$= 78.3 \newline Speed= 94.3 FPS & Slightly lower recall than YOLOv8l and YOLOv10x; added model complexity. & --- \\
\hline

\cite{zhang2023ats} & ATS-YOLOv7 & YOLO-MHA model for improved object detection accuracy in UAV imagery applications. & DIOR dataset & mAP= 87 \newline Speed= 94.2 FPS & Limited testing on diverse weather and extreme occlusion scenarios. & --- \\
\hline

\cite{zhao2023ms} & MS-YOLOv7 & YOLO-Swin  model for enhanced detection of small, dense objects in UAV images. & VisDrone2019 & mAP$_{0.5}$= 53.1 & Struggles with aerial images featuring severe occlusion. & --- \\
\hline

\cite{tahir2024pvswin} & PVswin-YOLOv8s & YOLOv8-Swin framework to enhance efficiency in UAV object detection tasks. & VisDrone2019 & Pre= 54.5\newline mAP$_{50}$= 43.3 & Limited performance in detecting extremely small objects. & ---\\

\hline

\cite{zhu2024natca} & NATCA YOLO & Proposed neighborhood attention Transformer-enhanced model for improved small object detection in aerial images. & VisDrone2019-DET  & mAP$_{50}$=66.27\newline Speed= 37 FPS   & Suffer from reduced performance in foggy weather and difficulty distinguishing similar objects. & --- \\

\hline


\cite{sankararao2024uc} & CNN-ViT & Released crop HSI datasets, modeled growth stages using Transformer fusion. & Crop
growth stage & Acc= 98.86 & Model complexity increases training time and demands high computational resources. & \href{https://github.com/sankaraug/CrHyperS}{Github}\\\hline

\cite{bashmal2021uav} & Improved ViT & Improves UAV scene labeling using data-efficient Transformers and data augmentation. & Civezzano & Spe= 97.91\newline mAP= 69.79 & Requires long training time and heavy data augmentation use. & --- \\\hline
\cite{lu2023cascaded} & ViT variant & A ViT-inspired model with customized attention mechanisms designed to address UAV nighttime image denoising. & DarkTrack2021 & Pre= 57.4\newline Sr= 44.2 & Limited performance under sudden drastic lighting changes during nighttime tracking. & ---\\\hline
\cite{qian2024efficient} & YOLOv5-GA & Combines deformable convolutions and attention in YOLO for UAV detection. & AU-AIR & AP$_{50}$= 52.6 \newline Speed= 46.8 FPS & Higher computation, slower speed, and reduced large-object detection. & --- \\
\hline
\cite{yao2024can} & LLM (GPT-3.5, ChatGLM) & Introduces CNER-UAV, fine-grained dataset for address resolution, comparing human and LLM annotation, evaluating NER models. & CNER-UAV & Acc > 90 (human annot.) & LLMs cannot fully substitute human annotation; struggle with fine-grained tags and diverse/complex text errors & \href{https://github.com/zhhvvv/CNER-UAV}{Github} \\\hline
\cite{lykov2024flockgpt}& LLM (GPT4) & LLM-guided drone flocking with natural language for shape formation; used SDFs and flocking algorithms for swarm pattern control. &Simulation  & Mean Acc= 80 &  Variability in LLM-generated patterns affecting recognition for some shapes& \href{https://github.com/Taintedy/flock_gpt}{Github} \\\hline
\cite{wang2024multi}& LLM (GPT) & LLM-driven multi-UAV placement for integrated access and backhaul; iteratively optimizes UAV positions for robust QoS &  Simulation & > 82\% of theoretical optimal score & Performance drop (~15\%) compared to DRL baseline & --- \\\hline
\cite{trigg2024natural}& GPT-4o-DRL & NavChat framework enhances understanding of decisions by the DRL navigation agent. & Simulation & Acc=100  & Relies primarily on OpenAI's LLMs; further LLMs and visual explanations needed & \href{https://github.com/tantak/NavChat.github.io}{Github} \\\hline
\cite{yuan2024patrol} &VLM-YOLO-LLM& Proposes patrol agent, autonomous UAV enabling urban patrol and tracking with VLM, YOLO, cloud LLM. & Simulation & CCs= 66.10 \newline CAs= 89.47 & Slow response speed due to cloud-based LLM; Limited scene variety compared to the real world. & ---  \\\hline
\cite{xiang2024real}& LLM-DRL & Proposes reward shaping with fine-tuned ChatGLM2-6B, knowledge base to boost DRL agent performance and training efficiency in UAV air combat. & Simulation &WR=70 (2v2 missile combat
task) & Limited by training data quality; high resource costs and focus only on 1v1/2v2 scenarios &---  \\\hline
\end{tabular}
}
\begin{flushleft}
\scriptsize{\textbf{Abbreviations:} YOLO-Transformer (YOLO-T); global Attention (GA). Correct captions (CCs); Correct actions (CAs);Win rate (WR) \\ 
\textbf{Note:} All AI metrics are in percentage (\%)  } 
\end{flushleft}
\end{table}

\subsection{LLM-based UAV}

\noindent
\Ac{LLM}--UAV frameworks integrate \acp{LLM} with the multi-module architecture of modern \ac{UAV} systems to realize agentic capabilities, including autonomous perception, reasoning, memory, planning, and human-like interaction. These frameworks enhance traditional \ac{UAV} functional modules---perception, navigation, planning, control, communication, interaction, and payload---by fusing real-time multimodal sensor data with natural language instructions and context-aware reasoning. A detailed LLM--UAV working principle  can be formulated as:
\[
\begin{cases}
X_s \xrightarrow{\text{Perception Module}} F_s, \\
P \xrightarrow{\text{Prompt Encoder}} F_p, \\
\big[ F_s, F_p \big] \xrightarrow{\text{LLM Core}} F_{\text{LLM}}, \\
F_{\text{LLM}} \xrightarrow{\text{Memory and Reasoning}} F_{\text{plan}}, \\
F_{\text{plan}} \xrightarrow{\text{Planning Module}} T, \\
T \xrightarrow{\text{Navigation \& Control}} A.
\end{cases}
\]

In \ac{LLM}-based framework, $X_s$ represents the raw sensor data collected by the \ac{UAV}, including inputs from RGB cameras, \ac{LiDAR}, radar, \ac{IMU}, and other onboard sensors. The \textit{perception module} processes this raw input $X_s$ and transforms it into rich perception features $F_s$ through advanced multimodal fusion techniques. Meanwhile, $P$ denotes the human-issued or system-generated prompt---such as a mission objective, instruction, or query---which is handled by the \textit{prompt encoder} that tokenizes and embeds this input into a language representation $F_p$. The \textit{LLM core} then combines the perception features $F_s$ and the encoded prompt $F_p$ into a unified contextual representation $F_{\text{LLM}}$. This is further processed by the \textit{memory and reasoning unit}, which maintains situational knowledge, decomposes complex tasks, and infers actionable steps. Next, the \textit{planning module} uses this inferred context to generate concrete, executable trajectories $T$ derived from the high-level plan $F_{\text{plan}}$. Finally, the \textit{navigation and control module} translates these planned trajectories $T$ into precise low-level actuation commands $A$ that control the \ac{UAV}’s motors and actuators to carry out the mission effectively.

\begin{figure}[ht!]
    \centering
    \includegraphics[width=0.85\linewidth]{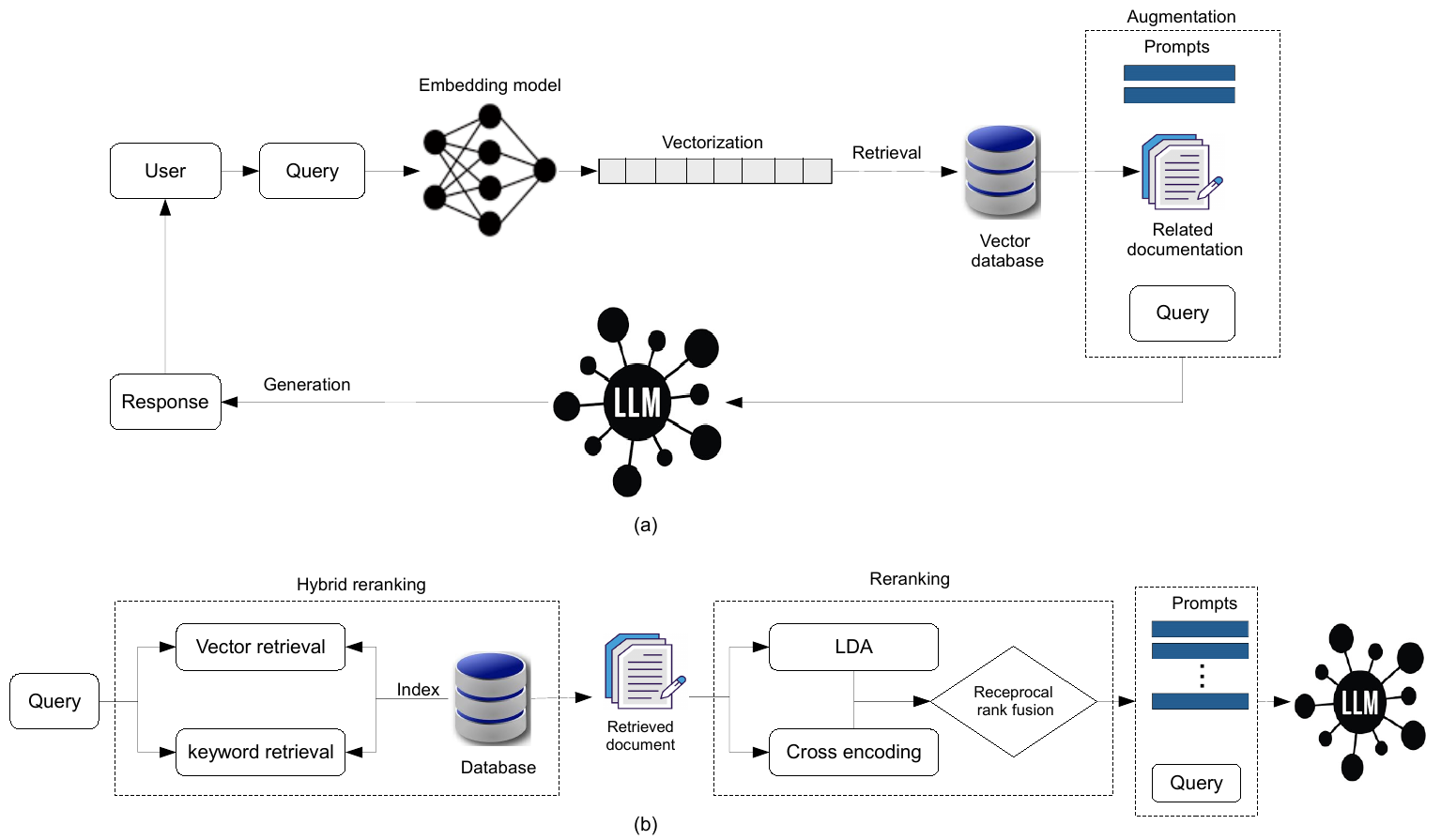}
    \caption{Hallucination-mitigation framework for UAV tasks: (a) Classical \ac{RAG}; (b) Enhanced \ac{RAG} system combining hybrid search and advanced reranking \cite{hao2025advancing}.}
    \label{fig:rag-llm}
\end{figure}

A growing body of research has investigated the integration of \acp{LLM} to support diverse tasks in \ac{UAV} applications.  One line of research focused on leveraging \ac{BERT} models for semantic understanding. For example, Mbaye et al. \cite{mbaye2023bert} presented a method that combined \ac{BERT}-based topic modeling with semantic information retrieval to enable the safe integration of \acp{UAV} in wildfire response operations. In this work, the \ac{UAV}’s main tasks included surveillance and hazard monitoring. The proposed \ac{LLM} framework processed unstructured historical incident reports from the aviation safety reporting system (ASRS) dataset and output structured fishbone diagrams to provide traceable root causes and risks, thus supporting systems engineers in identifying and mitigating potential \ac{UAV} failure scenarios. Several studies explored \ac{RAG} frameworks and domain-specific \acp{LLM} in order to reduce the challenging hallucination phenomenon. Hao et al. \cite{hao2025advancing} developed a domain-specific \ac{LLM} for biological \ac{UAV} swarm control and decision-making, building on an enhanced \ac{RAG} architecture with hybrid retrieval (vector and keyword) and reranking. The base \ac{LLM} employed the Moka massive mixed embedding (M3E) model for embeddings. Their \ac{RAG} system retrieved relevant domain knowledge and augmented generation to reduce hallucinations, improving factual accuracy and context relevance. \Ac{LDA} topic modeling boosted semantic reranking precision. The approach was evaluated using Recall, F1, BLEU, and Cosine Similarity on LeCaRDv2 and private \ac{UAV} datasets, showing strong performance but facing challenges due to domain data scarcity. Figure \ref{fig:rag-llm} illustrates the classical \ac{RAG} and the enhanced \ac{RAG} system.

A significant number of studies employed general-purpose \acp{LLM} such as GPT-series models, DeepSeek, or Gemini for UAV tasks. Moraga et al. \cite{moraga2025ai} integrated the experimental \textit{Gemini-2.0-Flash} \ac{LLM} with a real-time architecture combining the SUMO simulator and \ac{IoT} sensors for \ac{UAV}-assisted urban traffic optimization. The \ac{LLM} generated adaptive vehicle speed recommendations to reduce congestion and CO$_2$ emissions, though inference latency and computational costs remained barriers to real-time deployment. Fernandes et al. \cite{fernandes2025deepseek} evaluated DeepSeek-V3, GPT-4, Phi-4, and LLaMA-3.3 for generating correct Python code for UAV placement and received power calculations in long-range \ac{IoT} scenarios. Using zero-shot natural language prompts, they tested each model on tasks with progressively higher complexity. Performance was measured on a 0–5 correctness scale with syntactic, runtime, and output checks, showing DeepSeek-V3 and Phi-4 performed best overall. Wu et al. \cite{wu2025evolutionary} proposed the ARE-LLM framework, embedding GPT-3.5-turbo and GPT-4 into an evolutionary attraction–repulsion algorithm inspired by lion pride dynamics for \ac{UAV} swarm task allocation. The framework evolved labor division strategies via prompt engineering and bidirectional feedback. Tested on dynamic reconnaissance and attack scenarios with heterogeneous \acp{UAV}, ARE-LLM surpassed advanced baselines like ALPDA in cost efficiency and adaptability. Li et al. \cite{li2025large} designed an LLM-enabled decomposition-based multi-objective evolutionary algorithm that used GPT-3.5 Turbo to optimize \ac{UAV} deployment and power control in integrated sensing and communication systems. By decomposing the optimization problem into tractable sub-problems and solving them with prompt-engineered \acp{LLM}, the approach improved hypervolume performance for scenarios with quasi-static ground users. Zhou et al. \cite{zhou2025llm} presented LLM-QL, combining ChatGPT-4o with classical Q-Learning to schedule multiple parallel \acp{UAV} for the multiple flying sidekicks traveling salesman problem. Their framework integrated heuristic item generation via \ac{LLM} prompts and adaptive Q-table updates, demonstrating improved solution quality on real and synthetic datasets. Chang et al. \cite{chang2024iteratively} introduced a novel \ac{UAV} trajectory planning system that incorporated GPT-4 into a stochastic optimizer. Human operators provided latent knowledge through natural language instructions and map annotations, which the \ac{LLM} translated into new terms for the optimization objective. Experiments in wildfire search scenarios showed improved waypoint compliance and user learning. Yao et al. \cite{yao2025incorporating} combined GPT-4 with YOLOv11 and reinforcement learning to create an intelligent \ac{UAV}-assisted traffic light control system. The architecture integrated real-time vision-based detection, a clearance urgency indicator, and \ac{LLM}-driven adaptive signal control, yielding better performance in average speed and travel time in SUMO-simulated intersections. De et al. \cite{de2023semantic} integrated GPT-3 with CLIP and YOLOv7 to enable robust zero-shot semantic scene understanding for \acp{UAV}, generating detailed textual descriptions from aerial video streams. The output’s readability was evaluated using the \textit{GUNNING Fog Index} and \textit{FLESCH} scores.

Emerging works have proposed collaborative or multi-model architectures. Han et al.  \cite{han2025swarmchain} developed SwarmChain, a system combining StableLM and Llama2 for \ac{UAV} swarm control. The framework leveraged CoLLM for tensor-parallel distributed inference and an adaptive load scheduling algorithm to balance computation across resource-limited \acp{UAV}. AirSim simulations demonstrated improved instruction parsing and execution speed. Zhu et al. \cite{zhu2024task} introduced LLM-QTRAN, which combined DeepSeek with QTRAN, a value decomposition method for cooperative multi-agent \ac{DRL}, along with graph convolutional networks for UAV-assisted edge computing. Their decentralized POMDP formulation for UAV trajectory planning yielded faster convergence and higher task success rates compared to QMIX and standard QTRAN baselines.

Nevertheless, several limitations have been faced in all the suggested \ac{LLM}-based \ac{UAV} frameworks. These include dependency on prompt design quality, the computational costs associated with large \ac{LLM} inference, and latency challenges that can hinder real-time deployment. In addition, some methods face task-specific constraints, such as reliance on domain data availability, the accuracy of vision modules, or the static nature of test scenarios. Addressing these challenges remains essential for future research to ensure robust, scalable, and real-time \ac{LLM}-enabled \ac{UAV} systems.

\subsection{Lessons learned and key takeaways}

Based on the comparative analysis of Transformer-based \ac{UAV} models, several important lessons emerge:

\begin{itemize}
    \item \textbf{No one-size-fits-all architecture:} Different transformer variants suit different \ac{UAV} tasks. For example, \textit{\acp{ViT}} and \textit{Swin Transformers} excel in vision-based segmentation and classification, while \textit{\acp{STT}} and \textit{RL-based Transformers} are better suited for trajectory prediction and adaptive control.
    \item \textbf{Attention mechanisms are critical:} The type and granularity of Attention (e.g., multi-scale, hierarchical, pixel-wise) significantly impact model performance, particularly in dynamic and occlusion-prone scenarios. Attention mechanisms tailored to specific sensing modalities (e.g., IR, LiDAR) enhance robustness.
    \item \textbf{CNN--Transformer hybrids offer balance:} These hybrids effectively combine local feature extraction with global context modeling. They often outperform pure \acp{CNN} or Transformers in resource-constrained \ac{UAV} environments, striking a trade-off between performance and real-time feasibility.

    \item \textbf{LLMs are promising but underexplored:} While \acp{LLM} show potential for multimodal fusion and high-level decision-making, their deployment in \acp{UAV} is still in early stages due to latency and resource constraints.

    \item \textbf{Evaluation remains fragmented:} Many studies use different datasets and inconsistent evaluation protocols. This fragmentation hinders direct comparison and slows progress toward standardized benchmarks.

    \item \textbf{Real-time deployment challenges persist:} High-performing models often require computational resources that exceed those available on embedded \ac{UAV} hardware. This highlights the need for lightweight Transformer variants or knowledge distillation methods.
\end{itemize}

These insights also inform the broader research agenda discussed in Section~\ref{sec7}, particularly in areas such as energy-efficient architectures, multimodal fusion, and \ac{DTL} for deployment in real-world \ac{UAV} systems.

\section{Background}
\label{sec3}

\subsection{UAV Simulators}

UAV simulators play a pivotal role in advancing Transformer-based \ac{UAV} research by enabling controlled, repeatable, and scalable experimentation. In the context of \ac{DRL} and computer vision, simulators offer a safe and cost-effective alternative to real-world data collection, allowing researchers to model complex flight dynamics, environmental interactions, and sensor configurations. This is particularly beneficial when dealing with scenarios that are hazardous, rare, or prohibitively expensive to reproduce in physical settings—such as obstacle-rich urban landscapes, multi-agent coordination under uncertainty, or aggressive flight maneuvers. For Transformer architectures, which often require large volumes of diverse and annotated data, simulation environments can be scripted to generate synthetic datasets with fine-grained labels including bounding boxes, segmentation masks, depth, and inertial sensor streams. These datasets support supervised pretraining, domain adaptation, and multi-task learning strategies. Moreover, simulators facilitate the evaluation of generalization and robustness by enabling controlled perturbations to visual appearance, lighting, motion blur, and occlusion. Table~\ref{tab:uav_simulators} summarizes a set of widely adopted \ac{UAV} simulators that support such tasks, each offering distinct advantages in terms of realism, modularity, and integration with control stacks such as PX4 or \ac{DRL} libraries.

\begin{table}[ht!]
\centering
\caption{Comparison of popular UAV simulators with features, use cases, and access links.}
\label{tab:uav_simulators}
\scalebox{0.65}{
\begin{tabular}{m{2.2cm}m{9cm}m{7.2cm}m{5cm}m{2cm}}
\hline
\textbf{Simulator} & \textbf{Key Features} & \textbf{Use Cases} & \textbf{Target Users} & \textbf{Access Link} \\
\hline

\textbf{AirSim } & Unreal/Unity, photorealistic, physics-based, multi-agent & AI training, computer vision, reinforcement learning & Researchers, developers, AI engineers & \href{https://github.com/microsoft/AirSim}{Available} \\\hline

\textbf{Gazebo + PX4} & ROS-integrated, modular UAVs, high-fidelity physics & Autonomous navigation, swarm robotics, \ac{UAV} control & Robotics, ROS users & \href{https://docs.px4.io/}{Gazebo}, \href{https://gazebosim.org/}{PX4} \\\hline

\textbf{RotorS} & Lightweight ROS/Gazebo sim with quadrotor support & SLAM, planning, vision research & Academic researchers & \href{https://github.com/ethz-asl/rotors_simulator}{Available} \\\hline

\textbf{JSBSim} & Flight dynamics engine, no GUI by default & Control design, autopilot development & Flight dynamics engineers & \href{https://github.com/JSBSim-Team/jsbsim}{Available} \\\hline

\textbf{Flightmare} & Unity + physics, photorealistic + real-time & Perception, RL, navigation research & AI/ML researchers & \href{https://github.com/uzh-rpg/flightmare}{Available} \\\hline

\textbf{DJI Flight } & Official DJI sim, supports GPS/ATTI, wind effects & Pilot training, commercial ops & DJI users, enterprise trainers & \href{https://www.dji.com/simulator}{Available} \\\hline

\textbf{X-Plane 12} & FAA-certified, realistic physics, plugin support for custom UAVs & High-fidelity simulation and R\&D & Professionals, sim developers & \href{https://www.x-plane.com/}{Available} \\\hline

\textbf{RealFlight} & RC and drone flight, physics engine, controller integration & Hobbyist training, flight basics & Beginners, enthusiasts & \href{https://www.realflight.com/}{Available} \\\hline

\textbf{UAV Pilot} & Simple interface, indoor/outdoor environments & Educational use, control testing & Students, instructors & \href{http://www.uavpilot.org/}{Available} \\\hline

\textbf{FlightGoggles} & Photorealistic drone racing/perception sim, inertial rendering & Vision testing, drone racing & UAV perception researchers & \href{https://github.com/mit-fast/FlightGoggles}{Available} \\\hline

\textbf{DroneSim Pro} & Affordable, beginner-friendly, basic drone control training & Learning to fly, basic skills & New pilots, learners & \href{https://dronesimpro.com/}{Available} \\
\hline
\end{tabular}}
\end{table}

\subsection{Metrics}

In \ac{UAV} research, performance evaluation relies on a diverse range of metrics that can be broadly grouped into seven main categories based on the specific task or application: semantic segmentation \cite{lu2024lightweight,huang2023easy}, object detection and classification \cite{ye2022ct,liu2024action}, tracking \cite{cao2022tctrack,xing2022siamese2,siyuan2024learning,shen2023adaptive,wei2024unsupervised,zhu2021vitt,zhu2023spatio}, \ac{DRL} and decision-making \cite{fang2024online,jiang2023short,poorvi2024securing}, localization and regression \cite{wang2023autonomous,wu2023developing,he2024dcd,chen2023stexplorer,li2025large}, image quality and fusion \cite{li2024lswinsr}, efficiency and real-time performance \cite{zhou2023hybrid}, and \ac{LLM}-based \ac{UAV} metrics \cite{hao2025advancing,de2023semantic}. Each of these categories addresses a unique aspect of system capability — from measuring how well predicted regions align with ground truth, to assessing the accuracy of object identification, the consistency of object identity across frames, the precision of trajectory or positional estimates, the perceptual quality of generated images, or the system’s computational demands in time-critical scenarios.

For example, semantic segmentation metrics such as \ac{IoU} and \ac{mIoU} ensure accurate region prediction in UAV-captured images and maps. Object detection and classification tasks use metrics like \ac{mAP},  precision, recall, and Top-1 Accuracy to evaluate how reliably \ac{UAV} systems detect, localize, and categorize targets such as vehicles or people from aerial views. Tracking performance is commonly assessed by \ac{MOTA}, IDF1, and \ac{CLE}, which together quantify how well \acp{UAV}  maintain target identities and positions across video frames. \ac{DRL} and decision-making tasks are measured with cumulative reward and win rate, reflecting how effectively \ac{UAV} agents learn and optimize mission policies in dynamic environments. Localization and regression applications rely on \ac{RMSE}, \ac{MAE}, \ac{FDE}, \ac{ADE}, Chamfer Distance, and Age of Information (AoI) to assess how precisely \acp{UAV}  estimate coordinates, paths, or structural layouts. Image quality and fusion tasks are evaluated with \ac{PSNR}, \ac{SSIM}, pixel accuracy, $R^2$, and Pearson’s $r$, ensuring that UAV-captured or generated imagery retains visual fidelity and detail. Efficiency and real-time performance are benchmarked with \ac{FPS}, \ac{FLOPs}, and model parameter counts, which indicate the computational speed and resource needs critical for time-sensitive \ac{UAV} missions. Finally, \ac{LLM}-based \ac{UAV} tasks employ text similarity and readability metrics such as BLEU, Cosine Similarity, Flesch Reading Ease, and Gunning Fog Index to verify that auto-generated \ac{UAV} mission reports, captions, or instructions align semantically and remain accessible for human operators.

\begin{figure}[h]
    \centering
    \includegraphics[width=1\linewidth]{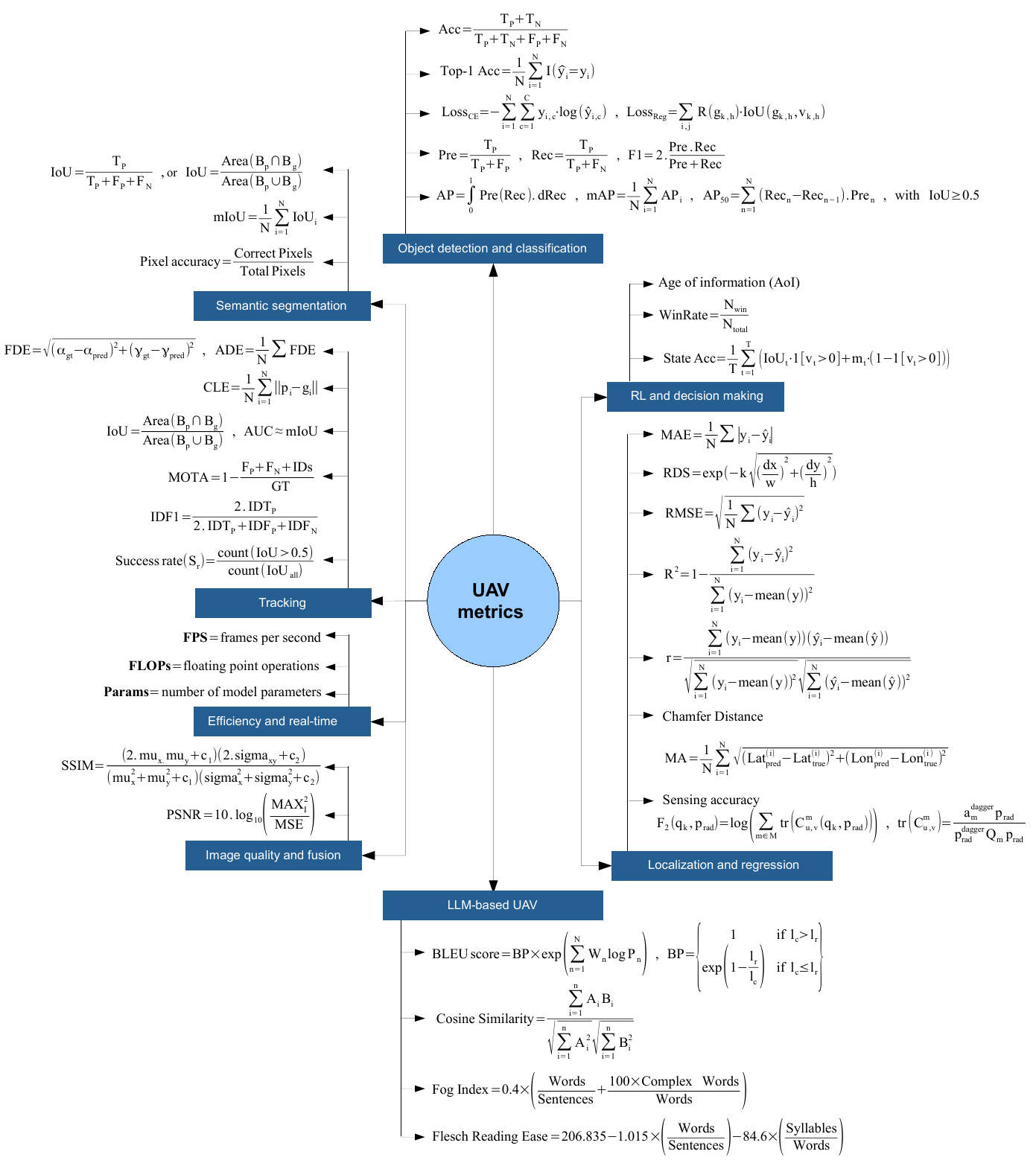}
    \caption{Taxonomy of metrics used in Transformer-based \ac{UAV} research domain. }
    \label{fig:metrics}
\end{figure}

Importantly, some metrics naturally overlap multiple categories; for instance, \ac{IoU} is central to both segmentation and detection tasks, and is also critical for evaluating tracking quality. Likewise, some measures are aggregate forms of others — such as mIoU and mAP — which summarize detailed per-instance results into a single representative score. Most of these metrics and their relationships are summarized visually in Figure \ref{fig:metrics}, while the exact parameter definitions, implementation details, and practical usage are thoroughly explained in the original works cited for each category above.

\subsection{Datasets}

The availability and quality of datasets play a pivotal role in the design, training, and rigorous evaluation of Transformer-based \ac{UAV} applications. Given the diverse operational scenarios and mission profiles of \ac{UAV} s, the datasets employed in this field encompass a broad spectrum of environments, sensing modalities, and annotation levels. Such diversity is essential for developing models that are robust and adaptable to the unique challenges of aerial data, including varying object scales, dynamic viewpoints, motion blur, cluttered backgrounds, and the demands of real-time processing. Table \ref{tab:datasets} provides an overview of representative \ac{UAV} datasets that address key tasks such as object detection, multi-object tracking, semantic segmentation, \ac{GL}, anomaly detection, and more. The table reveals dataset popularity both horizontally (by references) and vertically (by Transformer usage). Horizontally, UAVid, UDD6, UAV123 and it variant UAV123@10fps, Visdrone and it variant Visdrone-2019,  University-1652, and UAV-Human are the most  referenced datasets, indicating widespread use in Transformer-based \ac{UAV} studies. Vertically, \ac{CNN}-based, Attention-based, Siamese, YOLO-based and Swin Transformers dominate in terms of usage, while the proposed YOLO-Transformers techniques are focusing mostly on VisDrone2019 and AU-AIR datasets. \Ac{LLM} represent a recent Transformer technique, their application in \ac{UAV} has only recently begun to be explored by researchers, with most datasets being derived from simulations. From another point of view, the most frequently conducted task is tracking, featured in datasets like UAV123, DTB70, and UAVDT, highlighting its central role in \ac{UAV} applications. Classification, object detection, and semantic segmentation are also well represented. In contrast, tasks like anomaly detection (AD), routing, and swarm control have seen limited exploration, offering future research opportunities. These datasets serve as valuable benchmarks for advancing and validating Transformer-based approaches in \ac{UAV} research under realistic and complex conditions.

\begin{table}[h]
\scriptsize
\caption{Summary of the employed datasets for evaluating Transformer and \acp{LLM}-based \ac{UAV} capabilities. The symbol \CIRCLE{} indicates that the dataset has been used by  AI algorithm, while \Circle{} denotes that it has not been utilized. }
\label{tab:datasets}
\centering
\begin{tabular}
{m{1.8cm}m{0.5cm}m{1cm}m{0.3cm}m{0.3cm}m{1cm}m{0.1mm}m{0.2cm}m{0.5cm}m{0.6cm}m{0.6cm}m{0.4cm}m{0.4cm}m{0.5cm}m{0.3cm}cm{2.3cm}m{0.5cm}}
\toprule
 \multirow{2}{*}{Dataset}  & \multicolumn{5}{c}{Dataset overview} & \multicolumn{9}{c}{Transformer types using the dataset (In literature)} &  & \multicolumn{2}{c}{Availability} \\
 \cline{2-6} 
 \cline{8-15}
 \cline{17-18}
  &  DT &  Samples &  NoC &   L? & Task & & LLM  & C-T& Atten. & Siamese & Swin & STT & Y-T & ViT  & & Used in &  Link \\
\hline

Trento    & Image & 3415     & 13          & Yes & C & &  \Circle{} & \Circle{} & \Circle{}  & \Circle{} & \Circle{}  & \Circle{}  & \Circle{} & \CIRCLE{}  & & \cite{bashmal2021uav} & \href{https://www.mdpi.com/2076-3417/11/9/3974}{Yes} \\\hline

Civezzano    & Image & 3415    & 14          & Yes & C & & \Circle{} & \Circle{} & \Circle{}  & \Circle{} & \Circle{}  & \Circle{}  & \Circle{} & \CIRCLE{}  & & \cite{bashmal2021uav} &  \href{https://www.mdpi.com/2076-3417/11/9/3974}{Yes} \\\hline

UAVid    & Image & 420    & 8          & Yes & SS& & \Circle{} & \CIRCLE{} & \CIRCLE{}  & \Circle{} & \CIRCLE{}  & \Circle{}  & \Circle{} & \Circle{}  & & \cite{lu2024lightweight,yi2023uavformer,zhou2023hybrid,li2024lswinsr,kumar2022semantic}  & \href{https://uavid.nl/}{Yes} \\\hline

AID   & Image & 10K   & 30          & Yes & SS & & \Circle{} & \Circle{} & \CIRCLE{} & \Circle{} & \CIRCLE{}   & \Circle{}  & \Circle{} & \Circle{}  & & \cite{li2024lswinsr}  & \href{https://arxiv.org/pdf/1608.05167}{Yes} \\\hline

UDD6   & Image & 205   & 6         & Yes & SS& & \Circle{} & \CIRCLE{}  & \CIRCLE{}  & \Circle{} & \CIRCLE{}  & \Circle{}  & \Circle{} & \Circle{}  & & \cite{lu2024lightweight,zhang2025uav,yi2023uavformer,wang2023swin}  & \href{https://github.com/MarcWong/UDD}{Yes}\\\hline

MS COCO    & Image & $\approx$200k   & 171   & Yes & OD & & \Circle{} & \CIRCLE{} & \Circle{}  & \Circle{} & \Circle{}  & \Circle{}  & \Circle{} & \Circle{}  & & \cite{ye2022ct}  & \href{http://mscoco.org/dataset/\#upload}{Yes} \\\hline

ImageNet    & Image & 14M     & 21k  & Yes & OD & & \Circle{} & \CIRCLE{} & \Circle{}  & \Circle{} & \Circle{}  & \Circle{}  & \Circle{} & \Circle{}  & & \cite{ye2022ct} & \href{https://ieeexplore.ieee.org/abstract/document/5206848}{Yes} \\\hline

VOC    & Image & $\approx$19,694    & 20 & Yes & OD& & \Circle{} & \CIRCLE{} & \Circle{}  & \Circle{} & \Circle{}  & \Circle{}  & \Circle{} & \Circle{}  & & \cite{ye2022ct}  & \href{http://host.robots.ox.ac.uk/pascal/VOC/voc2012/}{Yes} \\\hline

WHU aerial    & Image & 	8139    & 2 & Yes & SS & & \Circle{} & \CIRCLE{} & \Circle{}  & \Circle{} & \Circle{}  & \Circle{} & \Circle{} &\Circle{}  & & \cite{huang2023easy}  & \href{https://gpcv.whu.edu.cn/data/}{Yes} \\\hline

WHU satellite   & Image  & 3135    & 2  & Yes &  SS & & \Circle{} & \CIRCLE{} & \Circle{}  & \Circle{} & \Circle{}  & \Circle{} & \Circle{} &\Circle{} & & \cite{huang2023easy}  & \href{https://gpcv.whu.edu.cn/data/}{Yes} \\\hline

UAV123    & Video &  123   & 1 & Yes & Tr & & \Circle{} & \CIRCLE{} & \Circle{}  & \CIRCLE{} & \Circle{}  & \CIRCLE{}  & \Circle{} & \CIRCLE{}  & & \cite{cao2022tctrack,hu2024tfitrack,zhu2023spatio,xing2022siamese2,li2023boosting,li2023adaptive}  & \href{https://ivul.kaust.edu.sa/benchmark-and-simulator-uav-tracking-dataset}{Yes} \\\hline

DTB70    & Video & 70   & 3 & Yes & Tr & & \Circle{} & \CIRCLE{} & \Circle{}  & \CIRCLE{} & \Circle{}  & \Circle{}  & \Circle{} & \Circle{}  & & \cite{cao2022tctrack,chen2024spatial}  & \href{https://github.com/flyers/drone-tracking?tab=readme-ov-file}{Yes} \\\hline

UAVTrack112L    & Video & 112 & 1 & Yes & Tr & & \Circle{} & \CIRCLE{} & \CIRCLE{}  & \Circle{} & \Circle{}  & \Circle{}  & \Circle{} & \Circle{}  & & \cite{cao2022tctrack,li2023boosting}  & \href{https://github.com/vision4robotics/SiamAPN/tree/master/UAVTrack112}{Yes} \\\hline

UAV20L    & Video & 20    & 5  & Yes & Tr & & \Circle{} & \CIRCLE{} & \CIRCLE{}  & \Circle{} & \Circle{}  & \Circle{}  & \Circle{} & \Circle{}  & & \cite{hu2024tfitrack,chen2024spatial} & \href{https://ivul.kaust.edu.sa/benchmark-and-simulator-uav-tracking-dataset}{Yes} \\\hline

UAV123@10fps    & Video & 123   & 1 & Yes & Tr & & \Circle{} & \CIRCLE{} & \CIRCLE{}  & \Circle{} & \Circle{}  & \Circle{}  & \Circle{} & \Circle{}  & & \cite{hu2024tfitrack,chen2024spatial,li2023boosting}  & \href{https://ivul.kaust.edu.sa/benchmark-and-simulator-uav-tracking-dataset}{Yes} \\\hline

Aeroscapes    & Image & 3269    & 11 & Yes & SS & & \Circle{} & \CIRCLE{} & \Circle{}  & \Circle{} & \Circle{}  & \Circle{}  & \Circle{} & \Circle{}& & \cite{zhou2023hybrid}  & \href{https://github.com/ishann/aeroscapes}{Yes} \\\hline

VOT2018    & Video & 60    & 1  & Yes & Tr & & \Circle{} & \Circle{} & \Circle{}  & \CIRCLE{} & \Circle{}  & \Circle{}  & \Circle{} & \Circle{}  & & \cite{xing2022siamese2}  & \href{https://www.votchallenge.net/}{Yes} \\\hline

OTB    & Video & 100    & 16 & Yes  & Tr & & \Circle{} & \Circle{} & \Circle{}  & \CIRCLE{} & \Circle{}  & \Circle{}  & \Circle{} & \Circle{} & & \cite{xing2022siamese2}  & \href{https://opendatalab.com/OpenDataLab/OTB100}{Yes} \\\hline

LaSOT   & Video & 1400   & 85  & Yes & Tr & & \Circle{} & \Circle{} & \Circle{}  & \CIRCLE{} & \Circle{}  & \Circle{}  & \Circle{} & \Circle{}  & & \cite{xing2022siamese2}  & \href{http://vision.cs.stonybrook.edu/~lasot/}{Yes} \\\hline

Got10K    & Video & 10k   & >563 & Yes & Tr & & \Circle{} & \Circle{} & \Circle{}  & \CIRCLE{} & \Circle{}  & \Circle{}  & \Circle{} & \Circle{}  & & \cite{xing2022siamese2}  & \href{http://got-10k.aitestunion.com/}{Yes} \\\hline

AU-AIR    & Video & 8    & 8  & Yes & T& & \Circle{} & \Circle{} & \Circle{}  & \Circle{} & \Circle{}  & \Circle{}  & \CIRCLE{} & \Circle{}  & & \cite{qian2024efficient}&  \href{https://bozcani.github.io/auairdataset}{Yes} \\\hline

UAV-SD    & Image & 10,000    & 2  & Yes & C & & \Circle{} & \Circle{} & \Circle{}  & \CIRCLE{} & \Circle{}  & \Circle{}  & \Circle{} & \Circle{} & & \cite{zhai2024cas}  & \href{https://github.com/tulingLab/CAS-Net}{OR}\\\hline

VisDrone2019     & Video & 288    & 10  & Yes & Tr, OD& & \Circle{} & \Circle{} & \Circle{}  & \Circle{} & \CIRCLE{}  & \CIRCLE{}  & \CIRCLE{} & \CIRCLE{}   & & \cite{hu2023stdformer,shen2023adaptive,li2023object,zhao2023ms,tahir2024pvswin}&  \href{https://github.com/VisDrone/VisDrone-Dataset}{Yes} \\\hline

3rd Anti-UAV   & Video &  600   & 1  & Yes  & Tr & & \Circle{} & \Circle{} & \Circle{}  & \Circle{} & \Circle{}  & \CIRCLE{}  & \Circle{} & \Circle{}  & & \cite{siyuan2024learning}  & \href{https://anti-uav.github.io/}{Yes} \\\hline

UAV lane-level     & Video & 600K    & 12  & No & P & & \Circle{} & \Circle{} & \CIRCLE{}  & \Circle{} & \Circle{}  & \Circle{}  & \Circle{} & \Circle{}  & & \cite{wu2023developing}  & \href{https://www.sciencedirect.com/science/article/pii/S0045790623002847}{OR} \\\hline

NAT2021-test    & Video & 180    &  1 & Yes & Tr & & \Circle{} & \Circle{} & \Circle{}  & \Circle{} & \CIRCLE{}  & \Circle{} & \Circle{} & \Circle{}  & & \cite{wei2024unsupervised}  & \href{https://vision4robotics.github.io/NAT2021/}{Yes} \\\hline

UAVDark70    & Video & 70   & 1  & Yes & Tr & & \Circle{} & \Circle{} & \Circle{}  & \Circle{} & \CIRCLE{}  & \Circle{} & \Circle{} & \Circle{}  & & \cite{wei2024unsupervised}  & \href{https://pan.baidu.com/s/1PTFwNoSxwZBmUSzDD3ti2A}{Yes} \\\hline

NAT2021-L-test    & Video & 23    &  1 & Yes & Tr & & \Circle{} & \Circle{} & \Circle{}  & \Circle{} & \CIRCLE{}  & \Circle{} & \Circle{} & \Circle{}  & & \cite{wei2024unsupervised}  & \href{https://vision4robotics.github.io/NAT2021/}{Yes} \\\hline

BASiC    & TSSD & 70    & 7  & Yes & C, P, AD& & \Circle{} & \Circle{} & \CIRCLE{}  & \Circle{} & \Circle{}  & \Circle{}  & \Circle{} & \Circle{}  & & \cite{ahmad2024transformer}&  \href{http://dx.doi.org/10.5281/zenodo.8195067}{Yes} \\\hline

University-1652   & Image & 50,218   & 701  & Yes & Lo & & \Circle{} & \Circle{} & \Circle{}  & \Circle{} & \CIRCLE{}  & \Circle{} & \Circle{} & \Circle{}  & & \cite{chen2023cross,fan2024cross,lv2024direction,li2023transformer,zhuang2022semantic,dai2021transformer}  & \href{https://github.com/layumi/University1652-Baseline}{Yes} \\\hline

FODAnomalyData   & Image & 81632    & -- & Yes & OD & & \Circle{} & \Circle{} & \Circle{}  & \Circle{} & \Circle{}  & \Circle{} & \Circle{} & \CIRCLE{}  & & \cite{munyer2022foreign}&  \href{https://github.com/FOD-UNOmaha/FODAnomalyData}{Yes} \\\hline

Seattle city   & Data & 640    & --  & No & Routing & & \CIRCLE{} & \Circle{} & \Circle{}  & \Circle{} & \Circle{}  & \Circle{}& \Circle{}& \Circle{}  &  & \cite{zhou2025llm}  & \href{https://github.com/optimatorlab/mFSTSP?utm_source=chatgpt.com}{Yes} \\\hline

UAVSwarm   & Image & 12,598     & 1  & Yes & Tr  & & \Circle{} & \Circle{} & \CIRCLE{}  & \Circle{} & \CIRCLE{}  & \Circle{}  & \Circle{} & \Circle{} & & \cite{wang2024target}  & \href{https://github.com/UAVSwarm/UAVSwarm-dataset}{Yes} \\\hline

UAV-Human   & Video & 67,428    & 119 & Yes & AR & & \Circle{} & \Circle{} & \CIRCLE{}   & \Circle{} & \Circle{}  & \CIRCLE{}  & \Circle{} & \Circle{}  & & \cite{xin2023really,xin2023skeleton,sun2021msst}  & \href{https://openaccess.thecvf.com/content/CVPR2021/html/Li_UAV-Human_A_Large_Benchmark_for_Human_Behavior_Understanding_With_Unmanned_CVPR_2021_paper.html}{Yes} \\\hline
WeedMap    & Image & 18,746    & 3  & Yes & SS & & \Circle{} & \Circle{} & \Circle{}  & \Circle{} & \Circle{} & \Circle{}  & \Circle{} & \CIRCLE{}  & & \cite{castellano2023weed}  &  \href{https://projects.asl.ethz.ch/datasets/doku.php?id=weedmap:remotesensing2018weedmap}{Yes} \\\hline

Visdrone    & Image & 10,209    & 10  & Yes & OD& & \Circle{} & \Circle{} & \Circle{}  & \Circle{} & \CIRCLE{}  & \Circle{} & \Circle{} & \Circle{}  & & \cite{xu2022fea}&  \href{http://aiskyeye.com/}{Yes} \\\hline
NWPU VHR-10    & Image & 800    & 10  & Yes  & OD& &  \Circle{} &  \Circle{} &  \Circle{}  &  \Circle{} & \CIRCLE{}  &  \Circle{}  &  \Circle{} &  \Circle{} & & \cite{xu2022fea} &  \href{https://labelbox.com/datasets/nwpu-vhr-10/}{Yes} \\\hline
ASRS   & Text & $\approx$ 2M    & --  & Yes & AD & & \CIRCLE{}  & \Circle{} & \Circle{}  & \Circle{} & \Circle{}  & \Circle{}  & \Circle{} & \Circle{}  & & \cite{mbaye2023bert}  & \href{https://www.kaggle.com/datasets/drxc75/nasa-asrs?utm_source=chatgpt.com}{Yes} \\\hline
LeCaRDv2   & Text & 16,701    & 50  & Yes & SC & & \CIRCLE{}  & \Circle{} & \Circle{}  & \Circle{}& \Circle{}  & \Circle{}  & \Circle{} & \Circle{}  & & \cite{hao2025advancing}  & \href{https://github.com/THUIR/LeCaRDv2}{Yes} \\
\bottomrule
\end{tabular}
\begin{flushleft}
\scriptsize{
  Abbreviations: Data type (DT);   Semantic segmentation (SS);  Swarm control (SC);Classification (C); Prediction (P); Number of classes (NoC); labeled (L); Tracking (Tr); Time-series sensor data (TSSD)    YOLO-Transformer (Y-T); CNN-Transformer (C-T); Object detection (OD);  Anomaly detection (AD); Localisation (Lo) Spatio-temporel Transformer (STT); On request (OR).}
\end{flushleft}
\end{table}

\section{Application of Transformer-based UAV}
\label{sec5}

This section explores the main investigated \ac{UAV} application domains where Transformers have shown the greatest impact, including action recognition, precise localization, target tracking and detection, scene segmentation, security monitoring and anomaly detection, as well as precision agriculture and emerging use cases. Across these tasks, Transformers offer enhanced modeling of spatial and temporal dependencies, enabling \ac{UAV}s to perform more intelligent perception, accurate decision-making, and adaptive mission planning under diverse and dynamic conditions. 

\subsection{Action recognition}
\Ac{AR} in \ac{UAV} footage posed unique challenges due to high-altitude perspectives, requiring advanced models like Transformers to capture spatio-temporal features effectively. For example, the paper \cite{xin2023skeleton} applied \ac{AR} to \ac{UAV}-based human monitoring, which was crucial for crowd analysis, disaster response, and anomaly detection from aerial views. It proposed Skeleton MixFormer, a dual-branch Transformer combining a local branch for short-term dynamics and a global branch for long-range temporal patterns. This composition integrated cross-joint and cross-frame attention, enhancing \ac{UAV} capability to detect complex human activities from sparse skeleton data. However, the scheme may struggled with rare or subtle motion variations in low-resolution skeleton data. Moving on, the work in \cite{liu2024action} introduced CI-STFormer, a Transformer-based network for skeleton-based \ac{AR}, targeting joint jitter and multiscale feature fusion. It integrated spatial sparse optimization, primal-level fusion, and a scale-level temporal convolutional network. By refining joint representation and temporal interactions, CI-STFormer achieved state-of-the-art results on several datasets, significantly improving robustness and accuracy in \ac{UAV}-based recognition. Similarly, \cite{sun2021msst} investigated Transformer-based \ac{AR} in \ac{UAV} contexts, addressing challenges from aerial viewpoints such as occlusion and motion blur. It introduced a dual-stream spatial-temporal model combining CNN and Transformer to capture global dependencies and fine-grained motion. Applied to the \ac{UAV}-Human dataset, the approach enhanced recognition accuracy, supporting \ac{UAV} applications in surveillance, disaster response, and behavior monitoring.

\subsection{Localization}

\ac{UAV} localization, particularly in GPS-denied or challenging environments, represented a critical application area where Transformer models delivered notable advancements. A central approach in this field was cross-view \ac{GL}, which involved matching images captured by a \ac{UAV} with geo-tagged satellite or ground-level imagery to determine the drone's position. This task remained challenging due to significant differences in viewpoint, scale, rotation, and environmental conditions between the image sources.

Several studies employed ViT or standard Transformer architectures as backbone networks for feature extraction in cross-view image matching. Li et al.~\cite{li2023transformer} proposed a method for UAV-satellite image matching using an adaptive semantic aggregation (ASA) module that aggregated patches into part-level features via soft partition-based attention maps. Zhuang et al.~\cite{zhuang2022semantic} introduced a semantic guidance module (SGM) leveraging pixel attention to align semantic parts across modalities, enhancing robustness to scale and spatial offset. Wang et al.~\cite{wang2024accurate} presented KFCM-Net, a Transformer-driven matching network incorporating a key texture recognition module (KTR-Net) and linear attention with mask filtering to strengthen position-related features under complex scenes. Fan et al.~\cite{fan2024cross} developed the SSPT method, which combined self-attention and cross-attention mechanisms for feature fusion and utilized a pyramid head structure for heatmap upsampling, improving precision in localization tasks. These approaches achieved competitive performance on datasets such as University-1652 and UL14.

Other contributions explored hierarchical Transformer variants such as the Swin Transformer. Lv et al.~\cite{lv2024direction} introduced a Swin-based model integrating a radial slicer network (RSN) and a local pattern network (LPN) to extract directional and environmental cues. By fusing multiscale features, their method improved robustness across varying \ac{UAV} altitudes and achieved state-of-the-art results on University-1652 and SUES-200 in both drone-view target localization and navigation tasks. Similarly, in airport runway monitoring, the study in~\cite{niu2024automatic} used a Swin Transformer-enhanced YOLOv5 model for the detection and predictive \ac{GL} of \ac{FOD}. \acp{UAV}  were deployed to capture high-resolution imagery under variable environmental conditions, and the Transformer component enabled robust detection of small-scale objects, demonstrating strong localization accuracy even in noisy scenes.

Further improvements to Transformer-based \ac{GL} were achieved through the incorporation of local attention mechanisms or graph-based enhancements. Guo et al.~\cite{guo2023agcosplace} proposed AGCosPlace, a visual positioning algorithm that augmented a Transformer backbone with a single-layer graph coding module to aggregate contextual information and boost image retrieval recall. Dai et al.~\cite{dai2021transformer} introduced FSRA, a method that segmented regions based on heatmap distribution and aligned them across views for fine-grained feature extraction. Similarly, Bui et al.~\cite{bui2023cross} modified the ViT structure to incorporate local token awareness, particularly leveraging the classification token to enhance matching accuracy by combining local and global feature representations. 
Yet, the proposed scheme lacked evaluation in GPS-denied environments and on unseen \ac{FOD} types. In bridge inspection, the work in~\cite{yin2024bridge} applied UAV-based multi-view image capture and 3D reconstruction to generate panoramic bridge images. A Swin Transformer was integrated into an improved YOLOv8 model to detect surface defects across ultra-high-resolution views, significantly improving detection and localization of fine-scale structural anomalies. However, the approach required high computational resources due to 3D reconstruction and ultra-high-resolution inputs.

While \ac{UAV} imagery is valuable in infrastructure monitoring and situational awareness, its real-time coverage remains limited for large-scale traffic prediction. In this context, the study in~\cite{islam2023transformer} shifted toward connected vehicle trajectory data for crash localization. A Transformer-Conformer ensemble was proposed to model spatial and temporal driving dynamics. The Transformer captured long-range dependencies, while the Conformer modeled local sequential behaviors. Though \ac{UAV} video data was acknowledged for its high resolution, the authors emphasized the superior scalability and continuity of connected vehicle data in real-time crash likelihood localization. Nevertheless, the model was limited by low connected vehicle data coverage and region-specific validation.

\subsection{Tracking and detection}

Autonomous tracking and detection were fundamental capabilities for \acp{UAV} across a range of applications, including surveillance and object monitoring. The inherent challenges of aerial imagery---such as varying object scales, complex backgrounds, and dynamic motion---drove researchers to adopt advanced deep learning techniques. In particular, Transformer-based models emerged as powerful solutions, demonstrating significant advancements in handling spatial and temporal contexts critical for these tasks. Recent works explored various Transformer architectures to enhance \ac{UAV} tracking and detection performance in diverse operational scenarios.

A substantial body of research concentrated on Transformer-driven detection methods aimed at boosting accuracy and robustness in aerial imaging. For example, Fan et al.~\cite{fan2022object} proposed a rotary-wing UAV object detection algorithm based on an AWin Transformer network, introducing a novel self-attention mechanism with annular windows to leverage local contextual information. This approach improved detection accuracy, achieving a 1.7\% increase in mAP on a custom rotary-wing UAV dataset. Li and Hussin~\cite{li2024application} enhanced a YOLOv7-based detector with the MobileVITv3 Transformer and attention mechanisms, alongside an improved Kalman filter for real-time UAV tracking. Tests on a custom UAV dataset showed an AP of 89.31\% for detection and improved tracking success rates, demonstrating robustness in both simple and complex backgrounds. Liang et al.~\cite{liang2023cross} developed CTPFNet, which incorporated Transformer blocks for global feature extraction and lightweight modules for small object detection, yielding significant performance gains on VisDrone2021-DET with fewer parameters than conventional detectors. Other contributions, such as VIP-Det~\cite{chen2024drone} and the YOLOX-L based wildland fire smoke detector~\cite{zhan2021pdam}, also focused on fusing multi-modal information and leveraging Transformer-based pyramids to enhance small target detection, particularly under challenging conditions; VIP-Det in particular demonstrated improved accuracy on the DroneVehicle dataset under night and fog scenarios.

For object detection in agriculture, Liang et al. \cite{liang2023detection} propose YOLOv5-SBiC, an enhanced object detection model tailored to detect late-autumn litchi shoots—small, visually ambiguous targets that typically elude standard detectors. The model modifies YOLOv5 by integrating a Swin Transformer module into the neck to improve multi-scale feature fusion and preserve fine-grained object details. The Swin Transformer introduces a hierarchical architecture with shifted window attention, significantly reducing computational overhead while maintaining the global feature modeling benefits of classical Transformers. Combined with a bi-directional \ac{FPN} and a \ac{CBAM}, this design enhances the network’s ability to localize small targets in complex orchard environments. The proposed method  outperforming standard YOLOv5 and a pure Transformer baseline. Similarly, Zhang et al. \cite{zhang2023transformer} present a comprehensive image detection method for monitoring grassland conditions in alpine meadows, where terrain, vegetation coverage, and degradation signs are difficult to segment using conventional methods. Their model, Am-mask, incorporates the Swin Transformer into a Mask R-CNN framework for semantic segmentation. Unlike CNNs, the Swin Transformer utilizes local self-attention confined within shifted windows, offering both scalability and efficiency. The Transformer is pretrained and then fine-tuned using task-specific prefixes and attention sparsity strategies to enhance generalization and reduce overfitting. Image stitching is used as a preprocessing step to obtain panoramic views, and the segmentation achieves an \ac{AP} of 95.4\%, significantly surpassing CNN-only baselines. The study demonstrates that Transformer-based segmentation is highly effective in capturing the diverse spatial patterns and degradation signals across large, heterogeneous grassland regions.

Building on these advances in detection, another major line of research addressed the development of Transformer-based tracking frameworks tailored to the dynamic requirements of UAV operations. Liu et al.~\cite{liu2023bactrack} introduced BACTrack, a lightweight aerial tracking system using mixed-temporal Transformer attention for efficient multi-template fusion, which achieved superior accuracy and speed on challenging aerial tracking benchmarks. Wang et al.~\cite{wang2024contextual} presented a Siamese network for \ac{UAV} visual tracking, utilizing a contextual enhancement-interaction module and multi-scale weighted fusion  to handle similar targets and scale variations effectively, achieving effective and stable aerial tracking. Wang et al.~\cite{wang2024dynamic} proposed a dynamic region-aware Transformer for \ac{UAV} tracking, utilizing sparse attention and DropKey for generalization, and reported 66.5\% AUC on the UAV123 benchmark. Yuan et al.~\cite{yuan2024multi} developed MT-Track, a multi-step temporal modeling framework using a temporal correlation module and mutual Transformer, which was evaluated on four \ac{UAV} benchmarks and achieved superior performance and real-time speed. Yu et al.~\cite{yu2023unified} further contributed a unified Transformer-based tracker for robust anti-UAV tracking in thermal infrared (TIR) mode, integrating four modules to handle multi-region local tracking, global detection, background correction, and dynamic small object detection for challenging scenarios. Similarly, the work in \cite{jeon2024autonomous} proposed a fully autonomous \ac{UAV} inspection system for overhead power transmission infrastructure using multimodal sensors—namely 3D \ac{LiDAR} and optical cameras. Central to this system was the RoMP Transformer, a novel object detection architecture tailored for identifying infrastructure with extreme aspect ratios (e.g., towers and insulator strings). It employed rotational bounding boxes to enable orientation-aware detection and integrated a multi-scale, multi-level feature pyramid to enhance spatial representation. Transformer-based attention modules were applied over the fused features to capture long-range contextual dependencies. The system was evaluated using an actual UAV-based dataset collected across 31 transmission towers and 9,000+ manually labeled instances, achieving a \ac{mAP} of 92.4\% in object detection for transmission tower detection task. However, its performance degraded under poor lighting or \ac{LiDAR} occlusion conditions.

Transitioning to even more advanced approaches, several studies combined detection and tracking or introduced cross-modal, hierarchical, and multi-task Transformer architectures to further enhance robustness and contextual awareness. For instance, Ke et al.~\cite{ke2024cross} designed a cross-scale feature enhancement (CFE) framework within a YOLOv8n backbone for UAV-based cotton seedling detection; this approach, incorporating a global information extraction (GIE) module, achieved an F1 score of 95.3\% on a collected dataset, outperforming standalone YOLOv8n, especially for small, overlapping, and blurred seedlings. In contrast, Yang et al.~\cite{yang2023echoformer} utilized raw radar echoes as input and introduced Echoformer, featuring a multi-layer Transformer extractor with \ac{MSA}, temporal embeddings, and \ac{MLP} blocks for comprehensive global feature learning, achieving an F1 score of 97.6\% on a real pulse-Doppler radar dataset. Wu et al.~\cite{wu2022gcevt} addressed vehicle tracking in \ac{UAV} videos by combining non-local blocks with a channel-wise Transformer enhancer (CTE), and reported an IDF1 score of 50.6\% on VisDrone2021 and 68.6\% on UAVDT, excelling even with similar or occluded targets. Finally, Wang et al.~\cite{wang2023hierarchical} introduced the hierarchical feature pooling Transformer (HFPT), integrating hierarchical CNN features and pooling in Transformer layers to reduce computational cost and improve tracking, particularly for small objects; on the DTB70 benchmark, HFPT achieved its best performance with a precision score of 79.0\% and a success score of 59.2\%.  These studies collectively demonstrated that Transformer-based models significantly improved \ac{UAV} detection and tracking by effectively capturing spatial, temporal, and multi-modal information to overcome the complex challenges of aerial vision tasks.

\subsection{Segmentation}

In recent \ac{UAV}-based image segmentation studies, hybrid CNN-Transformer architectures gained traction for their ability to model both local and global contextual features. UAVformer \cite{yi2023uavformer} exemplified this trend by introducing a composite backbone built on aggregation windows \ac{MSA}, coupled with a V-shaped decoder and positional attention mechanisms to enhance dense predictions in urban scenes. This architecture achieved a \ac{mIoU} of 53.2\% on the UAVid dataset. Similarly, the model presented in \cite{kumar2022semantic} incorporated a token spatial information fusion (TSIF) module, effectively blending convolutional networks for local detail with Transformer layers for long-range dependencies. This design yielded an \ac{mIoU} of 61.93\% on UAVid and 73.65\% on the Urban Drone dataset, highlighting the value of hybrid context modeling.

Efforts to improve multi-scale representation and boundary segmentation in \ac{UAV} image analysis led to the development of cascaded or composite Transformer models. For instance, CCTSeg \cite{yi2023cctseg}, designed for \ac{UAV} visual perception, adopted a cascade of Swin Transformers to progressively refine semantic features from coarse to fine scales. This approach resulted in enhanced boundary delineation and superior segmentation performance compared to CNN-based baselines such as DeepLabv3+. In the context of \ac{UAV}-based forest health monitoring, \cite{liu2024clusterformer} proposed Clusterformer to target pine tree disease identification by integrating cluster-based attention mechanisms and spatial-channel feed-forward networks. This design enabled more precise segmentation under complex backgrounds, outperforming conventional \ac{CNN} architectures in terms of boundary accuracy and robustness.

Transformers were also adapted for task-specific segmentation challenges. HSI-TransUNet \cite{niu2022hsi} addressed the complexity of crop mapping from \ac{UAV}-based hyperspectral imagery by integrating a \ac{ViT} with \ac{CNN} encoders. This hybrid setup allowed simultaneous exploitation of spectral and spatial features for improved accuracy. Similarly, in \cite{gibril2023large}, the segmentation of date palm trees from multi-scale \ac{UAV} imagery benefited from the Swin Transformer backbone, which outperformed classical \acp{CNN} such as Unet++ and DeepLabv3+. In the context of infrastructure monitoring, TLSUNet \cite{he2023transmission} combined lightweight GhostNet encoders with Transformer blocks to enhance long-range dependency modeling, achieving a notable \ac{mIoU} of 86.46\%, surpassing the baseline UNet performance of 79.75\%.

Lastly, several models focused on combining Transformer mechanisms with \ac{MSFF} techniques. In the marine monitoring domain, a Transformer-enhanced multi-scale fusion framework (MSFF + Transformer) \cite{liu2023image} integrated \ac{CNN}-based hierarchical features with attention modules to achieve smoother object boundaries and improved robustness in challenging scenarios, outperforming CNNs like PSPNet. Additionally, the self-cascade Transformer segmentation architecture proposed in \cite{chu2024transformer} employed a Transformer-guided cascade refinement strategy to ensure semantic consistency across scales, offering improved segmentation quality in complex \ac{UAV}-acquired imagery.

\color{black}

\subsection{Security and anomalies detection}

In the realm of aerial surveillance and anomaly detection, the work \cite{jin2022anomaly} introduced the ANDT model, a Transformer-based approach that treated aerial video streams as sequences of tubelets to detect temporal anomalies. The model employed an encoder-decoder Transformer structure to predict future video frames and flagged discrepancies as anomalies, thus enabling unsupervised anomaly detection. This work also contributed the Drone-Anomaly dataset, consisting of over 87,000 frames across seven real-world \ac{UAV} scenarios. However, the model was computationally intensive due to spatiotemporal encoding, and its generalization outside the Drone-Anomaly dataset remained unverified. Complementing this, \cite{zhao2024security} focused on \ac{UAV} swarm network security using TransReSE, a hybrid Transformer model enhanced with ResNeXt and Squeeze-and-Excitation mechanisms. Designed to assess situational awareness, TransReSE extracted multi-scale features and assigned importance to different channels, significantly boosting accuracy, recall, and F1 scores on four benchmark datasets. Nevertheless, the model’s integration complexity might have hindered deployment on resource-constrained \ac{UAV}s, and its performance under real-time distributed swarm conditions was not evaluated.

In heavily contested \ac{UAV} communication environments, Transformers proved crucial for proactive defense. The framework in \cite{elleuch2024leveraging} presented a hybrid anti-jamming strategy that combined pseudo-random (PR) channel selection with a Transformer module trained to predict jammer behavior. Designed for both \ac{UAV}s and high-altitude platform stations, the model operated in both offline and online modes, achieving over 70\% success rate in transmission under high-density jamming attacks. However, the approach assumed availability of historical jamming data for effective offline training, and real-world scalability under dynamic jamming strategies remained untested. In parallel, \cite{poorvi2024securing} tackled the dual challenge of data freshness and physical layer security in multi-\ac{UAV} systems using a gated Transformer-enhanced \ac{DRL} architecture. This system jointly optimized \ac{UAV} trajectories and \ac{IRS}-aided beamforming to combat jamming and eavesdropping, while minimizing the \ac{AoI}. The Transformer module enabled efficient task offloading and improved communication reliability. Yet, the model relied on accurate channel state and jammer location estimates, which might have been difficult to obtain in adversarial conditions; the \ac{IRS} deployment overhead was also not addressed.

Transformer-based architectures also showed significant promise in \ac{UAV} visual sensing applications, particularly in adversarial settings. Feng et al. \cite{feng2024security} introduced the SC-RTDETR framework for secure target recognition in \ac{UAV} forestry remote sensing. This system integrated a real-time detection Transformer with a soft-threshold adaptive filter and a cascaded group Attention (CGA) mechanism to filter out noise and improve feature learning under adversarial attacks. The framework demonstrated marked improvement in \ac{mAP} on pine wilt disease datasets. However, the framework was tailored to a specific type of dataset and disease scenario, limiting broader applicability; computational demands might have exceeded embedded \ac{UAV} capacities. In contrast, \cite{zhang2024empowering} explored the vulnerabilities of \ac{ViT}-based detectors in \ac{UAV} imagery by proposing a novel physical attack strategy. The authors introduced Jacobian matrix regularization, combining feature variance and Attention Weight regularization to create transferable adversarial patches. These patches degraded detector performance even under physical constraints, exposing critical flaws in \ac{ViT}-based models. Nonetheless, the proposed attack required precise patch placement and might have struggled in highly dynamic environments; defensive countermeasures against such attacks were not proposed. Table \ref{table:9} summarizes the performance of Transformer-based \ac{UAV} security approaches and quantified their improvements relative to baseline models.

To illustrate the diversity of Transformer-based architectures in UAV applications, the work in \cite{wang2023anomaly} introduced GTAF, an advanced anomaly detection model for multivariate time-series data from unmanned systems such as \ac{UAV}. It combined a multi-channel Transformer for contextual feature extraction with a graph attention network (GAT) to capture dependencies across data dimensions. A BiLSTM-based channel attention module fused outputs from different channels to enhance prediction quality. The model was evaluated using Precision, Recall, and F1-score. GTAF achieved outstanding performance, with F1-scores of 96.59\% and 92.83\% on two real-world \ac{UAV} datasets, demonstrating superior anomaly detection capabilities. Nevertheless, the model complexity and reliance on high-quality data hindered real-time deployment scalability.

\begin{table*}[ht]
\centering
\caption{Performance gain (\%) of Transformer-based \ac{UAV} Methods over baseline for several applications. For some schemes, results showing no improvement indicate that the method achieves competitive or state-of-the-art performance.}
\label{table:9}
{\fontsize{6}{7}\selectfont
\begin{tabular}{|l|l|l|c|c|c|l|}
\hline
Apps. & \textbf{Ref.} & \textbf{Transformer Method} &  \textbf{Dataset} & \textbf{Metric}  & \textbf{ Result (GCB)}  & \textbf{Task} \\
\hline

\multirow{6}{*}{\rotatebox{90}{Localization}}     & \cite{li2023transformer} & ASA     & University-1652           & Recall & 86.58 (+3.28)     & UAV-satellite image matching for Urban navigation\\
\cline{2-7}
& \cite{zhuang2022semantic} & SGM-Transformer                         & University-1652           & Accuracy & 84.7 (+8)         & Cross-view GL for UAV navigation and remote sensing\\
\cline{2-7}
& \cite{fan2024cross} & SSPT-Transformer                        & UL14                      & RDS      & 84.40 (+8.15)     & Cross-view Localization for UAV navigation\\
\cline{2-7}
& \cite{niu2024automatic} & YOLOv5 + Swin-T & Custom UAV FOD    & mAP & 89.6 (+2.4)       & FOD detection and localization for airport runway safety \\
\cline{2-7}
 & \cite{guo2023agcosplace} & AGCosPlace Transformer                  & SF-XL                     & Recall & 85.5 (+3.5)       & UAV visual Positioning for  image retrieval\\
 \cline{2-7}
 & \cite{dai2021transformer} & FSRA  & University-1652  & Recall & 87.3         & UAV-satellite matching for  instance matching \\\cline{2-7}
 & \cite{zhu2023uav} &  ViT-BERT & SUES-200& AP  & 93.63 (+9.1)    & Cross-view \ac{GL} using multimodal Transformer fusion \\

\hline\hline

 & \cite{lu2024lightweight} & LAPNet  & UDD6        & mIoU & 76.6 (+0.3)   & Semantic segmentation of low-altitude UAV aerial imagery. \\\cline{2-7}
& \cite{zhou2023hybrid} & HTCViT   & UAVid        & mIoU & 75.42 (+1.4)    & Semantic segmentation of remote-sensing UAV aerial images.\\\cline{2-7}
 & \cite{wang2023swin} & Swin-T-NFC CRFs   & UDD6        & mIoU & 74.59 (+1.41)   & Semantic segmentation and 3D UAV positioning precision. \\\cline{2-7}
& \cite{li2024lswinsr} & LSwinSR   & AID          & SSIM & 94.01 (+0.24)    & UAV image super-resolution for semantic segmentation. \\\cline{2-7}
\multirow{6}{*}{\rotatebox{90}{Segmentation}}    & \cite{castellano2023weed} & Lawin    & WeedMap        & F1 & 86.5 (+2.8)   & Semantic weed segmentation from UAV multispectral imagery. \\\cline{2-7}
& \cite{ghali2022deep} & TransUNet   & FLAME         & F1 & 99.90 (+0.24)   & UAV wildfire classification and segmentation. \\\cline{2-7}
& \cite{dos2022unsupervised} & SegFormer   & Acquired    & F1 & 83.93 (+0.18)    & Semantic segmentation of crop structure using UAV imagery \\\cline{2-7}

& \cite{yi2023cctseg} & CCTseg   & AERO       & mIoU & 70.1 (+1.8)    & Semantic segmentation for UAV autonomous visual perception. \\\cline{2-7}
& \cite{liu2024clusterformer}  &  Clusterformer   & Hubei       & IoU & 78.35 (+1.55)    & Pine disease segmentation using UAV remote sensing.\\\cline{2-7}
& \cite{niu2022hsi} & HSI-TransUNet   & UAV-HSI-Crop        & Accuracy & 86.05 (+7.41)    & Semantic crop mapping using UAV hyperspectral imagery. \\\cline{2-7}
& \cite{gibril2023large} & Segformer+UperNet-Swin   & Acquired       & mIoU & 86.36 (+0.91)    & Date palm segmentation using UAV Transformer models. \\\cline{2-7}
& \cite{liu2023image} & EMFNet   & CITYSCAPES        & mIoU & 76.0 (+0.7)    & Marine image segmentation using UAV and Transformer. \\\cline{2-7}
& \cite{chu2024transformer} & Cascade CATransUNet   & CrackLS315        & mIoU & 89.83 (+9.03)    & UAV-based high-resolution bridge crack segmentation system. \\

\hline \hline
  & \cite{fan2022object} &  AWin Transformer & Custom Rotor-Drone  & mAP & 89.5 (+1.7)   & Rotary-wing UAV object detection \\\cline{2-7}
    & \cite{li2024application} & MobileVITv3 & Custom UAV dataset  &  AP & 89.31 (+1.72)  & UAV image detection and real-time tracking \\\cline{2-7}
\multirow{6}{*}{\rotatebox{90}{Tracking and detection}}    & \cite{liu2023bactrack} &  Mixed-Temporal Transformer & DTB70  & Pre & 84.3 (+3.4)  & Robust aerial target tracking with multi-template fusion \\\cline{2-7}
 & \cite{wang2024contextual}  &   Siamese  & DTB70 & AUC & 65.5 (+0.9)   & UAV tracking with contextual interaction and multi-scale fusion  \\\cline{2-7}
  & \cite{liang2023cross} &   ELAN-Trans  & VisDrone2021-DET  & mAP & 18.1 (+2.8)  & Small object detection in UAV with altitude, density variation. \\\cline{2-7}

  & \cite{chen2024drone} &    ViT & DroneVehicle & mAP & 75.5 (+1.3)   &Visible–thermal UAV object detection under adverse conditions \\\cline{2-7}
   & \cite{wang2024dynamic} & DRAT & UAV123  & AUC & 66.5   & Real-time UAV tracking with sparse region-aware attention \\\cline{2-7}
  & \cite{yuan2024multi} &  Mutual Transformer & DTB70  & Pre & 85.9 (+4.2)   & Robust UAV tracking with  real-time efficiency \\\cline{2-7}
& \cite{zhan2021pdam} &   STPN  & Custom Dataset  & mAP & 77.86   & UAV-based forest smoke detection for early fire monitoring \\\cline{2-7}
& \cite{yu2023unified} & Unified Transformer &  Anti-UAV test-dev & AUC & 77.9 (+4.7)  & Thermal UAV target tracking with variation and occlusion.
 \\\cline{2-7}

\hline\hline
\multirow{3}{*}{\rotatebox{90}{AR}} 
& \cite{liu2024action} &  CI-STFormer & UAV-Human & Top-1 Acc. & 89.3 (+2.5) & Robust skeleton-based AR with joint jitter suppression \\
\cline{2-7}
& \cite{xin2023skeleton} & Skeleton MixFormer & UAV-Human Skeleton  & Top-1 Acc. & 88.5 (+2.7) & Human AR from aerial skeleton data \\
\cline{2-7}
& \cite{sun2021msst}& CNN-Transformer & UAV-Human & Top-1 Acc. & 86.0 (+3.4) & Human AR from UAV video streams \\

\hline \hline
 & \cite{poorvi2024securing}  &  Gated Transformer + DRL & Simulated multi-UAV  & TSR &  78.0 (+12.0) &   AoI and secrecy performance in task offloading \\
\cline{2-7}
\multirow{6}{*}{\rotatebox{90}{Security}} & \cite{feng2024security} &  SC-RTDETR  & Pine wilt disease UAV imagery& mAP   & 89.1 (+12.9) &   Detection performance under adversarial attacks \\
\cline{2-7}
& \cite{zhao2024security} & TransReSE  & Four UAV security datasets & F1 & 90.8 (+7.6) &   Efficient across multiple UAV security datasets \\
\cline{2-7}
& \cite{elleuch2024leveraging} & Transformer + PR selection & Simulated jamming  & CSR  & 70.0 (+12.0)&   CSR under heavy jamming (vs. static methods) \\
\cline{2-7}
& \cite{jin2022anomaly} & ANDT  & Drone-anomaly &  FPA  & 91.3 (+7.3)&   FPA in anomaly detection \\
\cline{2-7}

& \cite{zhang2024empowering}  & ViT + Jacobian matrix reg. & UAV remote sensing  & ATSR &  84.0 (+19.0)&   Adversarial transferability in physical attack scenario \\\cline{2-7}

 & \cite{anidjar2023stethoscope} &  Wav2Vec2 & In-house 3-hour UAV audio  & F1 & 88.4 (+1.4)  & Identifying flight anomalies via UAV sound patterns \\\cline{2-7}
& \cite{golam2024blm} & CNN-LLM blockchain & f 5G-NIDD & F1 & 93.85 (+1.56) & AI-driven blockchain framework for UAV threat resistance.\\\cline{2-7}
& \cite{piggott2023net}& Llama-2-7B & Network traffic & Accuracy & 97.03 (+19.01) & LLM-based chatbot conducts UAV man-in-the-middle attacks.\\
\hline \hline
\multirow{4}{*}{\rotatebox{90}{Agriculture}} 
& \cite{hafeez2023abnormal} & ViT & Cotton field, and public  & Accuracy & 97.56  (+2.44) & Abnormal image classification for smart agriculture \\
\cline{2-7}
& \cite{zhang2023swint} &  SwinT-YOLO & Custom UAV maize tassels  & AP & 95.11 (+1.3) & Object detection of maize tassels \\
\cline{2-7}
& \cite{gao2024aerial} & LoFTR (Retrained) & Wheat field (ETH Zurich) & RE & 27 cm (avg. error) & Inter-day image alignment and monitoring \\
\cline{2-7}
& \cite{liang2023detection}& Swin-T  & Litchi orchard UAV  & AP & 79.6 (+4.0) & Detection of late-autumn litchi shoots \\
\cline{2-7}
& \cite{zhang2023transformer} & Swin-T  & Alpine meadow UAV imagery & AP & 95.4 (+10) & Grassland segmentation and degradation detection \\

\hline
\end{tabular}
\begin{flushleft}
 \textbf{Abbreviations:} Channel success rate (CSR); Frame prediction accuracy (FPA); Task success rate (TSR); Attack transfer success rate (ATSR); Gain compared to the baseline (GCB); registration error (RE); Dynamic Region-Aware Transformer (DRAT); Small-Scale Transformer Feature Pyramid Network (STPN)
\end{flushleft}}
\end{table*}

\subsection{Other applications}

Beyond surveillance and \ac{AR}, Transformer-based models have been increasingly adopted in \ac{UAV}-enabled environmental monitoring tasks. Their ability to extract contextual and multi-scale features has enabled new directions in object detection, restoration, and environmental assessment from aerial imagery. For example, the work \cite{fu2024floating} introduced a \ac{ViT}-based framework for \ac{UAV}-assisted floating waste detection in water environments. It applied object-centric learning using a dual-branch architecture: one for encoding general object features and another for capturing instance-level semantics. By incorporating a request-driven mechanism and attention-based feature aggregation, the method enabled precise waste localization and discovery. This enhanced \ac{UAV} applications in environmental monitoring, supporting sustainable water resource management and automated pollution detection. However, its performance is limited by reliance on well-defined user queries and clean visual contexts.\ Similarly,  \cite{kieu2024enhancing} presented an interesting application of Transformer models for image inpainting. The authors evaluated two methods: deep image prior (DIP) and decoupled \ac{STT} for recovering occluded turbidity plume information from \ac{UAV} imagery. This allowed for better estimation of water quality in obstructed regions. However, the main limitation lay in the model sensitivity to obstruction complexity and motion dynamics.  Decoupled \ac{STT} performed poorly when occlusion exceeded a moderate threshold, and DIP, while more robust, operated slowly due to its optimization-from-scratch nature. Additionally, both methods assumed that the visual structure of the unobstructed region was sufficient to infer the missing areas—a premise that may not have held in highly dynamic coastal scenes with shifting turbidity and marine activity. Moving on, in \cite{gao2024aerial}, Gao et al. present a novel inter-day registration framework for crop imagery using \ac{UAV} photogrammetry. Central to their method LoFTR, a state-of-the-art dense image matching Transformer, originally designed for static urban scenes. The authors retrain LoFTR on a dynamic agricultural dataset, incorporating a novel supervision strategy that replaces height values in 3D landmarks to account for crop growth. This adaptation enables accurate feature matching despite significant inter-day variations in plant shape and lighting conditions. The refined pipeline achieves a registration error as low as 27 cm over a growing season, allowing near-plant-level monitoring for extended time periods without repeated ground control point usage.

\subsection{Lessons learned and key takeaways}

The surveyed applications of Transformer-based models in \ac{UAV} systems reveal several important insights and emerging design principles:

\begin{enumerate}
    \item \textbf{Enhanced perception and control:} Transformer architectures significantly improve \ac{UAV} perception tasks such as object detection, semantic segmentation, and trajectory tracking. Their attention mechanisms enable superior contextual awareness by modeling long-range dependencies—surpassing traditional \ac{CNN}-based approaches, especially in dynamic environments.

    \item \textbf{Strength of hybrid models:} \ac{CNN}–Transformer hybrids offer an effective balance between local and global feature extraction. This combination enhances robustness across varying aerial conditions and proves especially valuable in practical scenarios such as precision agriculture, surveillance, and disaster response.

    \item \textbf{Scalability through hierarchical attention:} Models like Swin Transformers employ window-based attention mechanisms that reduce computational load while maintaining high performance. These are particularly suitable for embedded \ac{UAV} systems dealing with high-resolution imagery, though they require tuning for domain-specific constraints.

    \item \textbf{Spatio–temporal reasoning for dynamic environments:} \acp{STT} excel in modeling motion patterns and temporal dependencies, making them ideal for real-time navigation, crowd monitoring, and coordinated multi-agent behavior. However, their real-time deployment is still hindered by inference delays and hardware limitations.

    \item \textbf{Emerging role of LLMs:} \acp{LLM} introduce high-level reasoning, mission planning, and human–UAV interaction via natural language. While current adoption is limited by size and latency, LLMs show strong potential for intuitive control in collaborative defense and civilian applications, especially with advances in prompt tuning and model compression.
\end{enumerate}

These insights underscore that successful deployment of Transformer-based \ac{UAV} systems requires a trade-off between model performance and operational feasibility. Future work should focus on cross-domain generalization, real-world benchmarking, and edge-efficient Transformer design.

\section{Case studies}
\label{sec6}

This section presents two representative case studies in the domain of \ac{UAV} intelligence enhancement: the first focuses on a Transformer-based \ac{UAV}  tracking framework, while the second explores a \ac{LLM}-based \ac{UAV} framework. By examining these systems in detail, the section highlights key architectural components, and practical deployment considerations. The inclusion of publicly available source code for both case studies enhances reproducibility and transparency, making these works valuable references, and excellent entry point for junior researchers seeking to explore and contribute to the growing field of intelligent \ac{UAV} systems.

\begin{figure}[ht!]
    \centering
    \includegraphics[width=1\linewidth]{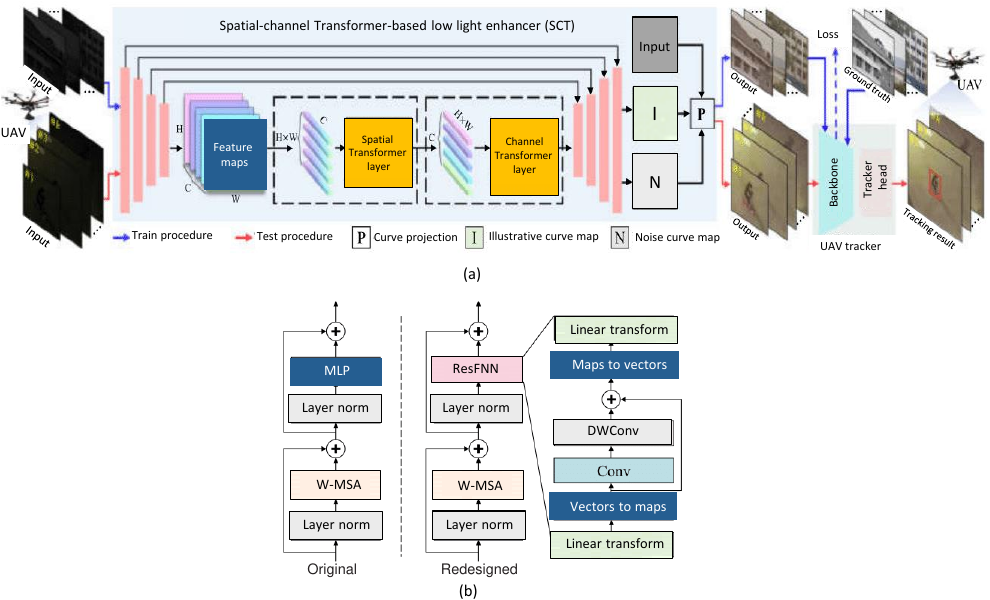}
    \caption{Overview of the SCT-enhanced Transformer framework for nighttime \ac{UAV} tracking and Transformer module design comparison.
(a) The SCT module is inserted before the tracking network to enhance low-light inputs by estimating illumination and noise maps, followed by a robust curve projection for simultaneous denoising and brightening of template and search patches. Trained with a task-specific perceptual loss, SCT aligns image enhancement with the needs of tracking, improving feature quality and object localization in nighttime \ac{UAV} applications.
(b) Transformer layer design comparison: the original version uses window-based W-MSA followed by an \ac{MLP}, which lacks local detail preservation. The redesigned version replaces the \ac{MLP} with a ResFFN, combining convolutional operations with attention to better retain local features and boost enhancement and tracking performance.}
    \label{fig:11}
\end{figure}

\noindent \textbf{-- Transformer-based UAV:} As a case study of Transformer-based \ac{UAV} tracking, the work \cite{ye2022tracker} introduces \acf{SCT}, a novel low-light enhancement module designed to improve visual tracking in nighttime \ac{UAV} applications. The pipeline, illustrated in Figure \ref{fig:11} (a), begins by preprocessing incoming low-light images using SCT before feeding them into a tracking network. SCT enhances both the template and search patches by estimating two key components: an illumination curve map and a noise curve map. These maps are generated using a hybrid CNN-Transformer architecture that captures both local and global contextual features. A robust non-linear curve projection mechanism then utilizes these maps to simultaneously brighten and denoise the input frames, improving the quality of features extracted downstream.

A central contribution is the Transformer layer design, detailed in Figure \ref{fig:11} (b), which departs from the traditional \ac{MLP}-based feedforward structure. Instead, it employs a residual convolutional feedforward network (ResFFN). This allows the model to retain and enhance fine-grained local structures—crucial for visual tracking under low-light conditions. The self-attention mechanism used is a window-based \ac{MSA} (W-MSA) module that operates within non-overlapping spatial windows, significantly reducing computational overhead while modeling long-range dependencies.

The overall architecture effectively integrates spatial and channel attention, yielding improved perception of object features in darkness. Training is task-driven, guided by a perceptual loss derived from feature maps of a tracker backbone (e.g., AlexNet), thus aligning enhancement with the needs of tracking rather than generic image restoration. In deployment, \ac{SCT} acts as a plug-and-play module, offering substantial tracking accuracy boosts across various benchmark trackers without requiring additional nighttime-labeled data. The authors have made their implementation and the newly constructed DarkTrack2021 benchmark publicly available, which facilitates reproducibility and further experimentation. This openness significantly enhances the framework’s practical value, enabling both academic and industry researchers to benchmark, extend, or deploy the \ac{SCT} module in real-world \ac{UAV} scenarios. While \ac{SCT} adopts a Transformer-based enhancement strategy, its objective shares conceptual similarities with recent encoder–decoder-based image enhancement models such as R-REDNet \cite{boucherit2025reinforced}, which employs deeper convolutional layers and averaging-based skip connections to improve real-world image denoising. However, \ac{SCT} differs by integrating task-specific attention mechanisms and curve-based projection directly tailored for \ac{UAV} tracking under low-light conditions.

\begin{figure}
    \centering
    \includegraphics[width=0.8\linewidth]{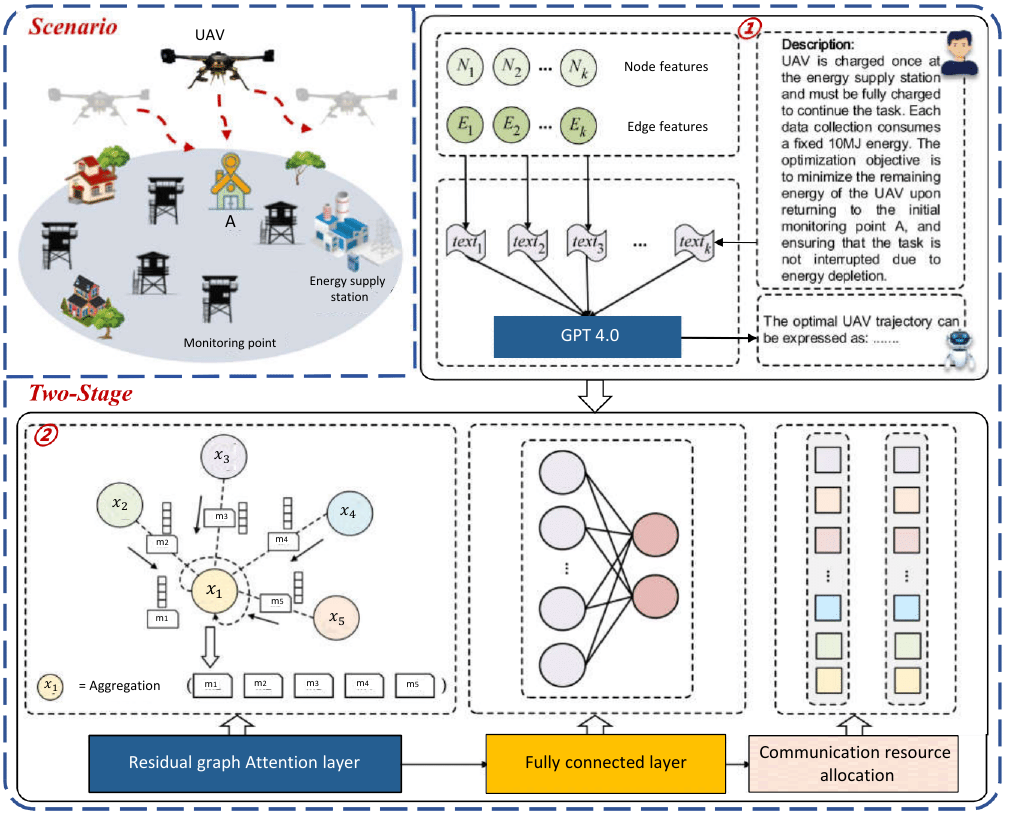}
    \caption{The \ac{LLM} case study describes a scenario module that uses a two-stage joint optimization strategy: first, a \ac{LLM} generates the UAV’s flight trajectory, and then a \ac{GNN} allocates communication resources. These two stages work together, ensuring that the \ac{UAV}’s path and resource allocation satisfy the mission’s objectives and constraints \cite{sun2024large}.}
    \label{fig:CaseLLM}
\end{figure}

\noindent \textbf{-- LLM-based UAV:} The \ac{LLM} case study depicted in Figure \ref{fig:CaseLLM} demonstrates the proposed framework through a \ac{UAV} networking scenario designed to optimize both trajectory planning and communication resource allocation. The scenario involves a \ac{UAV} that must visit several monitoring points, collect data, recharge at an energy supply station, and return to its starting point while managing its energy efficiently. The system adopts a two-stage joint optimization strategy. 

In the \textit{\textbf{first stage}}, an \ac{LLM} block (e.g., GPT-4) transforms extracted features — such as \ac{UAV} start position, distances between points, and data volumes — into textual inputs. This LLM generates an initial \ac{UAV} trajectory plan based on semantic understanding of these features. In the \textit{\textbf{second stage}}, a \ac{GNN} block processes the combined information (the initial trajectory and graph data) to allocate communication resources such as bandwidth and power. Within the GNN, an attention layer dynamically weighs the importance of different nodes and edges, ensuring that critical monitoring points or links receive appropriate resources while balancing overall energy consumption. The residual connections between the \ac{LLM} and \ac{GNN} blocks help maintain information flow and reinforce feature learning. The output block verifies results and optimizes both path and resource allocation under constraints. Experiments show that the LLM+GNN framework outperforms LLM+Node2Vec and LLM+GAT baselines, demonstrating robust energy efficiency and stable performance even as task sizes increase — though the attention mechanism can sometimes over-prioritize high-importance nodes, which requires careful tuning. Importantly, the authors have made their implementation publicly available to support reproducibility, which further strengthens the framework’s practical value \cite{sun2024large}.

\section{Research challenges and future directions}
\label{sec7}

 This section explores the key limitations facing current Transformer-\ac{UAV} integrations and outlines promising research directions aimed at overcoming these challenges and suggesting future endeavors while enhancing scalability, efficiency, and adaptability in real-world \ac{UAV} operations.

\subsection{Research challenges}

The integration of Transformer, and \acp{LLM} architectures into \ac{UAV} systems has opened new frontiers in perception, decision-making, and autonomous control. Despite their impressive capabilities in modeling long-range dependencies and processing multi-modal data, several challenges hinder the widespread adoption and full potential of Transformer-, and \ac{LLM}-based \ac{UAV}. A deeper examination of their current limitations and operational constraints is presented below: 

\begin{itemize}[leftmargin=0.5cm]
    
    \item [(a)] \textbf{Real-time deployment of Transformers for UAVs:} Transformer models, while powerful in capturing long-range dependencies and contextual information, are computationally intensive due to their self-attention mechanism, which scales quadratically with input length. In \ac{UAV} applications, where real-time decision-making is critical, such complexity poses a significant limitation. The high number of parameters increases latency and memory demands, making deployment on resource-constrained onboard hardware challenging. This often necessitates reliance on edge servers or cloud computing, introducing communication delays that further hinder responsiveness. Moreover, the energy consumption associated with Transformer inference can drain \ac{UAV} battery life, limiting operational duration. Real-time performance is especially affected in tasks like path planning \cite{zhang2024real}, swarm coordination \cite{javed2024state}, or  and tracking \cite{qin2025pptracker}, where milliseconds matter. To overcome these issues, future research must focus on model compression techniques, efficient attention mechanisms (e.g., linear or sparse attention), and hardware-aware Transformer design. These solutions are essential to ensure low-latency, high-accuracy Transformer performance compatible with real-time UAV missions.

    \item [(b)] \textbf{Data scarcity and annotation cost:}  Transformer-based models are inherently data-hungry due to their high capacity and lack of inductive biases, making them particularly susceptible to overfitting when applied to \ac{UAV} tasks with limited annotated datasets. In this context, high-quality data annotation plays a pivotal role, as the performance and generalization ability of Transformers heavily depend not only on the quantity but also on the accuracy and consistency of labeled training data \cite{hegde2024importance}.  This issue is further compounded in \ac{UAV} contexts where data collection and labeling present unique challenges. \ac{UAV} imagery often exhibits varying object scales (due to altitude changes), dynamic viewpoints (from rapid aerial maneuvers), motion blur (caused by platform speed or wind), and cluttered backgrounds (e.g., urban environments or natural terrains). These visual complexities make manual annotation both time-consuming and costly, as they often require expert intervention and multiple verification stages. While a number of synthetic and simulated datasets, such as those developed for \ac{UAV} monitoring  \cite{barisic2022sim2air},  have been proposed to alleviate the data scarcity issue, many of these fail to accurately capture the variability and unpredictability of real-world scenarios, leading to domain gaps that impair generalization. As a result, Transformer models trained solely on synthetic \ac{UAV} data may underperform in practical deployments. Furthermore, the scarcity of diverse and high-quality \ac{UAV} datasets hampers the generalization ability of Transformer models, which typically excel when trained on large-scale, well-annotated corpora. As a result, the field suffers from a bottleneck in supervised learning progress, especially for fine-grained tasks like object detection, semantic segmentation, and \ac{AR} from aerial views. To overcome these challenges, future research must explore cost-efficient annotation strategies, including active learning, semi-supervised learning, and synthetic data generation through simulation or generative models. However, because simulated datasets often fail to fully replicate the complexity of real-world UAV environments, their use alone may not guarantee robust generalization. Thus, domain adaptation techniques, transfer learning from related domains, and realistic data augmentation are essential to narrowing the synthetic–real gap. Bridging this gap remains a critical research direction for enabling the scalability and real-world deployment of Transformer-based UAV systems.

    \item [(c)] \textbf{Multi-modal sensor fusion:} Despite the considerable potential of multi-modal sensor fusion to improve perception accuracy, situational awareness, and robustness in \ac{UAV} operations \cite{ye2023review}, its implementation in Transformer-based \ac{UAV} systems presents several critical challenges. A primary challenge is the need for precise calibration and temporal synchronization across diverse sensors such as \ac{LiDAR}, RGB cameras, \acp{IMU}, and GPS; any misalignment can degrade the performance of attention-based fusion mechanisms by introducing temporal or spatial noise into the input embeddings. Transformer architectures, while powerful in modeling long-range dependencies, are highly sensitive to input consistency and may propagate sensor misalignments through the entire attention pipeline. Furthermore, the real-time integration of heterogeneous, high-dimensional sensor data imposes substantial computational and memory demands—constraints that conflict with the limited onboard processing capabilities typical of small \ac{UAV} platforms. Environmental variability (e.g., fog, rain, or low light) can also disrupt modality-specific inputs, making it difficult for Transformers to maintain reliable attention weights across modalities. In addition, these models often require large labeled datasets for training and struggle to generalize to unseen sensor configurations or mission domains without extensive fine-tuning. Lastly, aligning multimodal data representations (e.g., 2D visual streams with 3D point clouds) remains a non-trivial task, especially when attempting to encode them into unified token sequences suitable for Transformer inputs. These challenges underscore the need for efficient, domain-adaptive, and resource-aware sensor fusion strategies specifically tailored for Transformer-based \ac{UAV} architectures. To address the unique challenges of multi-modal sensor fusion in Transformer-based UAV systems, future research should focus on designing fusion frameworks that are both modality-aware and computationally efficient. Transformer architectures offer strong potential for modeling complex cross-modal relationships through self-attention mechanisms; however, adapting them to UAV platforms requires optimizing their efficiency and robustness. Strategies such as modality-specific tokenization, cross-attention fusion modules, and hierarchical encoding can enhance the network's ability to learn meaningful correlations across disparate sensor inputs (e.g., \ac{LiDAR}, RGB, \ac{IMU}). Incorporating uncertainty modeling and confidence weighting into attention layers can also improve the resilience of fusion against sensor noise or failure. Furthermore, lightweight Transformer variants (e.g., MobileViT, Linformer) and edge computing techniques, including FPGA or TPU acceleration, are essential for meeting real-time constraints on resource-limited UAVs. Standardized multi-sensor alignment procedures and domain-adaptive pretraining on heterogeneous \acp{UAV} datasets will be key to improving generalization and transferability across missions and environments.

    \item [(d)] \textbf{Swarm and multi-agent coordination:} In Transformer-based \ac{UAV} systems present significant challenges. Real-time communication and synchronization among agents are hindered by limited bandwidth, latency, and packet loss. Transformers, while powerful, are computationally intensive, restricting their deployment on resource-constrained UAVs. Achieving decentralized coordination without sacrificing global task coherence is complex, especially in dynamic and unpredictable environments. Moreover, adapting Transformers to process multi-modal data from heterogeneous UAV sensors for shared situational awareness remains an open issue. Scalability is another concern, as coordinating larger swarms exacerbates data flow, model complexity, and decision latency. To address these challenges in Transformer-based UAV systems, several strategies can be employed. First, efficient Transformer variants such as sparse attention, linear attention, or lightweight architectures (e.g., Linformer, Performer) can reduce computational demands, enabling real-time execution on UAV hardware. Second, hierarchical or federated coordination frameworks can facilitate decentralized control while preserving global coherence through periodic synchronization. Third, adaptive bandwidth management and compression techniques can alleviate communication bottlenecks and reduce packet loss. Fourth, multi-modal sensor fusion modules with attention gating can help prioritize reliable sensor streams, improving robustness to environmental variability. Finally, scalable swarm learning algorithms that combine local learning with global updates can support coordination across large UAV fleets with minimal latency and model overhead.

    \item [(e)] \textbf{Interpretability and explainability:} Interpretability and explainability of Transformer-based UAV models pose critical challenges. Transformers, with their deep attention layers and complex architectures, function as black boxes, making it difficult to understand how decisions are derived. In UAV missions, where safety and real-time responses are crucial, the lack of transparency hinders trust and adoption. Visualizing attention maps offers some insight, but interpreting them in the context of dynamic flight scenarios remains difficult. Furthermore, explaining multimodal fusion and temporal dependencies in UAV tasks adds to the complexity. Achieving interpretable yet performant models demands novel explainable AI (XAI) approaches tailored for aerial decision-making contexts. To overcome these interpretability challenges, several targeted solutions can be implemented. First, developing domain-specific XAI methods, such as attention flow tracing or saliency-based heatmaps aligned with flight telemetry, can provide more intuitive insights into model decisions. Second, incorporating symbolic reasoning or rule-based modules alongside Transformer outputs can enhance transparency by grounding decisions in interpretable logic. Additionally, embedding uncertainty estimation within attention mechanisms can help indicate confidence levels, supporting safer and more accountable UAV operations. These strategies collectively enhance trust and facilitate deployment in mission-critical scenarios.

    \item [(f)] \textbf{UAV  sensors sensitivity challenge:}  UAV sensor sensitivity poses critical challenges for Transformer-based models. Sensors such as \ac{LiDAR}, IMUs, GPS, and  rgb-d cameras are vulnerable to noise, drift, and occlusion, which can degrade the quality of input data. Since Transformers rely on attention mechanisms that weigh all inputs, any distortion can misguide attention distributions, leading to poor feature extraction, temporal misalignment, and inaccurate decisions. The impact is more pronounced in dynamic environments where reliable perception is vital for navigation and coordination. When extended to \acp{LLM}, additional concerns arise. LLMs may misinterpret noisy or ambiguous sensor-derived prompts, leading to hallucinations or incorrect reasoning. Their sensitivity to input phrasing and lack of grounded understanding makes them especially prone to error when sensor data is translated into text. Moreover, LLMs typically lack real-time feedback mechanisms, making them less responsive to fluctuating sensor reliability. Addressing these issues requires sensor-aware prompting strategies, uncertainty quantification, and hybrid architectures that fuse symbolic reasoning with perception-driven models for robust UAV applications.

    \item [(g)] \textbf{Transformer-based UAV models  threats: }  Transformer‑based UAV  face evolving threats. Flight variability and spectrum congestion allow attackers to vary timing, protocol and modulation, undermining learned attention patterns. Sophisticated adversaries craft perturbations in radio frequency waveforms or telemetry streams, subtly shifting token distributions to trigger misclassification, while replay and spoofing exploits leverage temporal self‑attention windows. Limited training data for rare signal types amplifies overfitting, degrading generalisation to unseen intrusions. \ac{LLM}‑driven IDS add further vulnerabilities. Textual reasoning layers can hallucinate interpretations of malicious traces, especially when prompts derive from noisy, compressed sensor logs. Prompt‑injection or adversarial suffixes may redirect the model’s focus, suppressing alarms. The black‑box nature complicates forensic auditing, and the parameter space enlarges the attack surface for trojaned weights or backdoor activations. To mitigate these challenges, several solutions have been proposed. Robust adversarial training using a mix of synthetic and real attack patterns can improve model generalization and resistance to sophisticated intrusions. Incorporating multi-modal fusion with uncertainty quantification allows the system to assess and adjust for sensor reliability, reducing the impact of noisy or manipulated inputs before processing by Transformer or \ac{LLM} models. Additionally, deploying lightweight explainability tools and runtime monitoring mechanisms can help detect abnormal attention patterns or prompt manipulations, enhancing both the interpretability and the security of \ac{UAV} intrusion detection systems in real-time operations.

\end{itemize}

\subsection{Future directions}

To overcome the aforementioned limitations and advance the capabilities of Transformer- and \ac{LLM}-based \ac{UAV} systems, it is crucial to explore forward-looking strategies and innovative solutions. The following future directions, drawn from both the reviewed studies and recent developments in related fields, offer valuable insights that can be adapted to enhance performance, reliability, and scalability in UAV contexts. These future directions aim to pave the way for more intelligent, autonomous, and resilient UAV applications: 

\begin{itemize}[leftmargin=0.5cm]
\item [(a)] \textbf{Lightweight Transformer- and LLM-based UAV variants:} One of the most critical directions for advancing Transformer- and \ac{LLM}-based \ac{UAV} systems lies in the development and deployment of lightweight and computationally efficient Transformer architectures. Standard models such as \ac{BERT} or \ac{ViT}, while powerful, impose substantial memory and compute requirements that exceed the constraints of edge devices onboard \acp{UAV}. To bridge this gap, emerging architectures like MobileViT \cite{mehta2021mobilevit}, TinyBERT, DistilBERT \cite{kheddar2025transformers}, Linformer \cite{wang2020linformer}, and Performer offer promising trade-offs between performance and resource efficiency. These variants reduce model complexity through techniques such as knowledge distillation, low-rank approximation, sparse or linear attention mechanisms, significantly lowering latency and power consumption. A complementary strategy involves designing \acp{SLM} \cite{lu2024small}  tailored to \ac{UAV}-specific tasks—such as mission summarization, instruction parsing, and inter-agent communication—while accommodating constraints like limited input length, real-time responsiveness, and domain-specific vocabulary. In this context, Transformer-based \ac{DTL} \cite{peng2025bayesian} and domain adaptation \cite{sohail2025advancing} play a pivotal role in creating lightweight yet task-effective models. Instead of training from scratch, \ac{UAV} developers can fine-tune pre-trained Transformer backbones (e.g., BERT, ViT, or T5) on UAV-relevant datasets, thereby leveraging knowledge learned from large corpora and adapting it to aerial domains. Moreover, cross-modal domain adaptation techniques allow Transformers to generalize across sensor inputs (e.g., visual, thermal, radar) or environmental variations (e.g., urban vs. rural, day vs. night), increasing robustness while reducing data collection costs. These latter methods help compress model size, reduce training time, and boost performance under real-world \ac{UAV} constraints.

\item [(b)] \textbf{ Exploring self-supervised and semi-supervised learning in UAV:} Despite the rapid progress in applying Transformers and \acp{LLM} to \ac{UAV} systems, current approaches overwhelmingly rely on supervised learning, which demands large amounts of labeled data—often difficult to obtain in aerial contexts. In contrast, \ac{SSL}-  and \ac{semi-SL}-based Transformer \cite{chen2021empirical,xing2023svformer} offer promising yet underexplored paradigms for enabling data-efficient model development. In the context of LLM-based \ac{UAV} applications—such as mission instruction parsing, human-\ac{UAV} interaction, or cross-agent communication—\ac{SSL} techniques like masked token prediction, contrastive pretraining, and next-sentence prediction can be used to pretrain models on massive corpora of unstructured aerial mission logs, telemetry data, or inter-agent dialogues. These self-supervised objectives allow \acp{LLM} to capture structural and semantic patterns without requiring manual annotation, thus supporting flexible adaptation to downstream UAV-specific tasks. Similarly, semi-supervised Transformers and \acp{LLM}  can leverage limited labeled data alongside large-scale unlabeled \ac{UAV} inputs (e.g., video frames, sensor readings, environmental reports) using methods like pseudo-labeling, consistency training, and teacher-student architectures. These approaches are particularly useful for fine-tuning \acp{VLM} or instruction-following \acp{LLM} in real-time scenarios, where fully annotated datasets are scarce. Moreover, combining \ac{SSL} and \ac{semi-SL} \cite{zhai2019s4l} with continual learning strategies enables \acp{UAV} to incrementally improve their models based on edge-collected data, enhancing robustness to domain shifts and operational variability. Future research should focus on integrating \ac{SSL} and \ac{semi-SL} frameworks into lightweight Transformer and \ac{LLM} variants, tailored for real-time \ac{UAV} deployment. This includes designing pretext tasks specific to aerial perception and communication, minimizing labeling costs, and maximizing generalization across diverse missions and environments. Advancing this direction will be essential to achieving scalable, autonomous, and intelligent \ac{UAV} systems that learn and adapt with minimal supervision.

\item [(c)] \textbf{Federated, collaborative,  and edge learning:} Transformer and LLM-based models offer powerful capabilities for perception, decision-making, and communication in \ac{UAV} systems. However, their training and inference typically require centralized data and substantial computational resources—assumptions that are not always feasible in real-world UAV deployments characterized by distributed agents, limited connectivity, and privacy concerns. Yet, federated learning, collaborative learning, and edge learning remain largely unexplored in the context of Transformer-based \ac{UAV} applications. These paradigms enable decentralized model training across multiple \ac{UAV} or edge nodes without requiring raw data sharing, making them highly suitable for \ac{UAV} swarms engaged in tasks like cooperative surveillance, distributed mapping, or autonomous coordination. Incorporating federated Transformers \cite{li2023fedtp} allows \acp{UAV} to collectively improve their models—e.g., visual Transformers or LLMs—based on local experiences, while only exchanging model updates. This preserves data privacy and reduces bandwidth usage, which are critical in contested or bandwidth-limited environments. Collaborative learning, where \acp{UAV} actively share partial representations, context, or knowledge graphs, could enhance semantic alignment and mutual understanding in swarm-level operations. Meanwhile, edge learning enables real-time adaptation of Transformers on-device, allowing \acp{UAV} to adjust to local conditions without depending on the cloud. When integrated with lightweight Transformer variants and self-supervised or semi-supervised pretraining, these distributed learning strategies can dramatically improve scalability, robustness, and generalization across diverse missions and environments. Future research should explore novel architectures and protocols that support federated fine-tuning of Transformer backbones, dynamic model aggregation, cross-modal collaboration among heterogeneous \acp{UAV}, and energy-efficient edge deployment. Addressing challenges such as model drift, communication efficiency, and security in federated Transformer training is crucial. Advancing this direction will pave the way for fully autonomous, intelligent, and cooperative \ac{UAV} networks capable of learning and evolving in decentralized, real-world settings.

\item [(h)] \textbf{Exploit unused DRL techniques:} Although recent advancements have integrated sophisticated \ac{DRL} methods—such as \ac{SAC}, \ac{MADDPG}, and multi-agent \ac{AC} frameworks—with Transformers in UAV applications, several foundational and effective \ac{DRL} techniques remain largely absent from the literature. Notably, classical value-based methods such as state–action–reward–state– action (SARSA), Q-Learning, asynchronous advantage actor–critic (A3C), double DQN (DDQN), and prioritized DQN (pri-DQN) \cite{kheddar2024reinforcement,basnet2025advanced}, have yet to be explored in conjunction with Transformer architectures for \acp{UAV}. These methods, while simpler, offer stable learning dynamics and can benefit significantly from Transformer-based temporal or spatial state encoding, especially in discrete action spaces like grid-based navigation or decision-tree decision-making. In addition, enhanced Q-learning approaches like double DQN (DDQN) and prioritized DQN (pri-DQN)—which address issues such as overestimation bias and inefficient experience replay—are promising candidates for integration with Transformers, particularly in partially observable environments where attention mechanisms can augment the value estimation process. Furthermore, asynchronous \ac{AC} methods like A3C, known for enabling parallelized learning and better exploration, have not yet been adapted with Transformer-enhanced policy/value networks in \ac{UAV} scenarios. These methods could unlock new possibilities in large-scale, real-time, or mission-critical multi-UAV applications. Future research should investigate how these underutilized \ac{DRL} frameworks can be reimagined using Transformer components to improve sample efficiency, stability, and task-specific generalization across various \ac{UAV} missions.

\item [(j)] \textbf{Enhancing UAV visual perception through descriptor fusion:} A compelling future direction for \ac{UAV} systems is the enhancement of visual perception capabilities by combining local and global feature descriptors within CNN-Transformer hybrid models, alongside task-specific denoising techniques. In visually complex \ac{UAV} missions—such as long-range surveillance, target tracking, or inspection—robust feature extraction is crucial for consistent recognition and scene understanding. Traditional descriptors like local maximal occurrence (LOMO) offer strong local texture sensitivity \cite{chouchane2024multilinear}, while CNN-Transformer fusion models enable hierarchical and context-aware representations that capture both spatial detail and global semantics. Integrating these via a high-dimensional feature fusion (HDFF) framework, such as tensor-based schemes that unify heterogeneous descriptors, can lead to more discriminative and resilient visual encoding, even in challenging aerial conditions. 

\item [(i)] \textbf{Visual language model for task-specific:}  A promising future direction for Transformer-based \ac{UAV} systems lies in the development of task-specific \acp{VLM} that jointly process visual inputs (e.g., aerial images, video streams) and textual data (e.g., mission instructions, annotations, or environmental reports). While general-purpose \acp{VLM} like BLIP \cite{li2022blip}, Flamingo \cite{alayrac2022flamingo}, and GPT-4V \cite{yao2025efficient} have demonstrated impressive performance across a range of vision-language tasks, their direct application to \ac{UAV} missions is limited by domain mismatch and computational overhead. Task-specific \acp{VLM}, trained or fine-tuned on UAV-relevant data, can bridge this gap by aligning semantic understanding with mission-critical objectives such as target recognition, damage assessment, path description, or instruction-grounded control. These models enable \acp{UAV} to interpret multi-modal inputs in real time and generate mission-aware outputs—such as narrating surroundings, reporting anomalies, or following natural language commands with visual grounding. Future research should focus on curating specialized aerial vision-language datasets, designing lightweight architectures suited for onboard inference, and exploring multi-task and cross-modal attention mechanisms to improve alignment between visual perception and linguistic reasoning. Task-specific \acp{VLM} will be pivotal in enabling interactive, explainable, and human-aligned autonomy in next-generation \ac{UAV} systems.
    
\item [(d)] \textbf{Cross-modal and \ac{MTL}:} A promising yet underdeveloped direction in Transformer- and LLM-based \ac{UAV} research lies in the integration of cross-modal and \ac{MTL} paradigms \cite{yan2023cross}. \acp{UAV}  typically operate in environments saturated with heterogeneous data sources, such as RGB/thermal imagery, \ac{LiDAR}, \ac{IMU} readings, environmental metadata, and textual instructions, making them ideal candidates for cross-modal Transformer architectures capable of learning rich, unified representations from multiple modalities. Existing approaches often process each modality in isolation or employ late fusion strategies, which limit the model's ability to reason jointly across inputs. Emerging multimodal models such as ViLBERT \cite{lu2019vilbert}, and BLIP \cite{li2022blip} can be adapted for \ac{UAV} use cases to enhance context-awareness, real-time scene understanding, and human-machine interaction through shared attention mechanisms across modalities. In parallel, \ac{MTL} offers a scalable and sample-efficient framework for training a single Transformer \cite{shridhar2023perceiver} or \ac{LLM} model \cite{xia2024efficient} to perform multiple \ac{UAV}-related tasks—such as navigation, target recognition, obstacle avoidance, instruction following, and swarm coordination—within a shared architecture. This encourages the development of more generalizable and efficient models, capable of transferring knowledge across related tasks and improving robustness in unseen scenarios. The combination of cross-modal and \ac{MTL}, such as cross-modal multitask Transformers \cite{srivastava2024omnivec2}, presents an especially powerful direction for UAV autonomy. By leveraging shared representations across modalities and objectives, such models can simultaneously interpret complex multimodal inputs (e.g., video and text) and execute diverse tasks (e.g., path planning, threat assessment, and report generation) within a single cohesive framework. This not only reduces computational overhead and deployment complexity but also improves adaptability and decision coherence in dynamic, real-world \ac{UAV} missions.

\item [(e)] \textbf{Human-UAV collaboration with Transformers, LLMs, and digital twins: } A critical future direction for autonomous aerial systems is the development of intelligent frameworks that support effective human-UAV collaboration, enabled by Transformer architectures, \acp{LLM}, and digital twin-based \ac{UAV} technologies \cite{lv2021digital}. As \acp{UAV} are increasingly deployed in complex, high-stakes environments—such as disaster response, precision agriculture, surveillance, and infrastructure inspection—there is a growing need for systems that can interpret human intent, adapt to dynamic objectives, and communicate transparently. Traditional control interfaces based on manual inputs and rigid protocols often lack the flexibility and accessibility required for seamless cooperation. In contrast, \acp{LLM} and \acp{VLM} offer a powerful means of enabling \acp{UAV} to process natural language commands, interpret contextual cues, and generate semantically aligned responses in real time. Moreover, the integration of digital twins—virtual replicas of \ac{UAV} systems and their operating environments—provides a shared, interpretable space for real-time monitoring, simulation, and human-in-the-loop control. Digital twins can visualize the \ac{UAV}’s state, planned trajectory, environmental conditions, and mission progress, allowing human operators to better understand and intervene when needed \cite{oulefki2025innovative}. Combined with multimodal Transformer models that fuse vision, language, and sensor data, such systems support shared autonomy, where control responsibilities fluidly transition between human and machine based on context and confidence. Future research should focus on developing lightweight, explainable Transformer-LLM architectures that interface seamlessly with digital twin platforms, enabling bidirectional communication, adaptive planning, and trust-driven collaboration. This integration will be foundational to building the next generation of human-centered, intelligent, and context-aware \ac{UAV} systems.

\item[(f)] \textbf{Standardized benchmarks and evaluation:} As Transformer architectures and \ac{LLM}s continue to reshape the capabilities of \ac{UAV} systems in perception, decision‑making, and human interaction, there is a growing need for standardized benchmarks and evaluation protocols tailored to this emerging paradigm. Currently, research in this domain is fragmented, often relying on custom datasets, task‑specific metrics, and ad‑hoc evaluation setups that hinder reproducibility, comparability, and holistic performance assessment. This lack of standardization limits researchers’ ability to rigorously quantify progress, identify bottlenecks, and generalize findings across \ac{UAV} platforms and applications. Future work should therefore prioritise the creation of comprehensive, multimodal \ac{UAV} datasets that include synchronised visual, textual, sensor, and mission‑level data suitable for evaluating Transformer and \ac{LLM} models across a wide range of tasks—such as instruction following, target detection, route generation, and collaborative decision‑making. In addition to traditional performance indicators, there is a pressing need to introduce new metrics that are tailored to the multimodal and generative nature of Transformer and \ac{LLM}-based UAV systems. For instance, a cross-modal alignment score (CMAS) could quantify how well generated textual commands or plans align with the UAV's visual context or mission objectives \cite{zhou2023improving}, using retrieval-based metrics like Recall@k. Grounding consistency score (GCS) could evaluate the percentage of correctly referenced visual or geospatial entities within generated language, assessing the model's multimodal grounding capabilities. To address safety and trustworthiness, metrics such as hallucination rate (HR) and toxicity risk score (TRS) can help capture the frequency of factually unsupported or potentially harmful content generated by LLMs; tools like LLMDet~\cite{wu2023llmdet} may be integrated into evaluation pipelines for automated detection of adversarial or manipulated language. Furthermore, an Attention faithfulness index (AFI), based on the divergence between attention heatmaps and external saliency methods (e.g., Grad-CAM), could assess the interpretability and internal reasoning transparency of Transformer models. \ac{DRL}-based UAV models could be assessed with trajectory success reward (TSR), which measures cumulative mission rewards, while normalizing with energy consumption per reward (E/R) provides insights into operational efficiency. Lastly, hardware-aware metrics such as edge latency per token (ELPT) \cite{eliopoulos2025pruning} can evaluate the real-time feasibility of LLM-based systems deployed on embedded platforms, by measuring the average inference delay per output token. By incorporating these metrics, future research can move beyond narrow accuracy benchmarks and embrace a broader evaluation spectrum that captures safety, alignment, interpretability, and system efficiency. This should be supported by scenario-based simulations, digital twin environments, and human-in-the-loop experiments to ensure the real-world relevance and robustness of Transformer- and \ac{LLM}-based \ac{UAV} systems. Such a comprehensive evaluation framework is essential for advancing trustworthy and mission-ready aerial autonomy.

\item [(g)] \textbf{Enhancing UAV reasoning via Transformers: } One promising direction for future research lies in enhancing high-level reasoning capabilities in \acp{UAV} using Transformer architectures. While current Transformer-based UAV models excel at tasks such as tracking, classification, and semantic segmentation, their potential for complex decision-making and inference remains underexplored. Future efforts should aim to integrate symbolic reasoning, planning, and contextual understanding within Transformer frameworks to support autonomous missions that require situational awareness and adaptive behavior. This could involve coupling Transformers with knowledge graphs for image or point-cloud \cite{kheddar20image}, neurosymbolic models \cite{cai2025neusis}, or \acp{LLM} fine-tuned for domain-specific UAV logic. Additionally, designing lightweight, interpretable Transformer variants capable of on-board reasoning under computational constraints will be essential for real-world deployment. Such advancements would empower \acp{UAV} not only to perceive their environments but also to reason, predict, and act with a higher degree of autonomy in dynamic and uncertain scenarios.

\item [(g)] \textbf{Exploring GAN- and AE-based Transformers:}  Another promising research direction involves the integration of \acp{GAN} and \acp{AE} with Transformer architectures to address data scarcity, uncertainty, and complexity in \ac{UAV} applications \cite{li2024zero,li2021md}. \ac{GAN}-Transformer hybrids can be leveraged for high-fidelity data generation \cite{liu2025transformer}, sim-to-real domain adaptation, and synthetic augmentation of aerial imagery or data, which is especially valuable in training perception models under diverse or rare conditions. In parallel, \ac{AE}-Transformer models offer a compact and robust framework for feature compression, anomaly detection, and latent representation learning from multimodal UAV sensor streams. For instance, VQ-VAE Transformers  can enable efficient encoding of visual scenes prior to decision-making \cite{yan2021videogpt}, while sequence-to-sequence \ac{AE}-Transformers can support trajectory prediction or event summarization in real time. These combined architectures present an opportunity to enhance the adaptability and generalization capabilities of Transformer-based \ac{UAV} systems, particularly under limited supervision or constrained computing environments. Further research should investigate their optimization for edge deployment, adversarial robustness, and alignment with regulatory constraints in safety-critical missions.

    
\end{itemize}

\section{Conclusion} 
\label{sec8}

This review has presented a comprehensive analysis of recent advancements in Transformer- and \ac{LLM}-based models for \ac{UAV} applications. By categorizing various Transformer architectures—such as \ac{CNN}–Transformers, \acp{ViT}, Swin Transformers, Spatio–Temporal Transformers, and \ac{DRL}-based models—the study underscored their impact on enhancing perception, decision-making, and reasoning in aerial systems. Hybrid models like \ac{CNN}–Transformers effectively balance local and global feature extraction, while \acp{ViT} and Swin Transformers offer improved classification and scalability for high-resolution tasks. \acp{STT} enable dynamic modeling for tasks like trajectory prediction and swarm coordination, and \ac{DRL}-based Transformers enhance adaptive decision-making. The integration of \acp{LLM} introduces capabilities such as mission planning and natural language interaction, despite current deployment limitations. Key challenges remain, including computational demands, limited datasets, and the lack of standardized benchmarks. Future research should focus on lightweight and efficient architectures, robust multimodal fusion, domain adaptation, and real-time edge deployment. Transformer-based models hold substantial promise for next-generation \ac{UAV} autonomy, offering a path toward more intelligent, responsive, and collaborative aerial systems.

\printcredits

\section*{Declaration of competing interest}
The authors declare that they have no known competing financial interests or personal relationships that could have appeared to influence the work reported in this paper.

\section*{Data availability}
No data was used for the research described in the article.

\section*{Acknowledgement}

\bibliographystyle{elsarticle-num}
\bibliography{references}

\end{document}